\documentclass[journal]{IEEEtran}
\usepackage{times}
\usepackage{epsfig}
\usepackage{graphicx}
\usepackage{amsmath}
\usepackage{amssymb}

\usepackage{comment}
\usepackage{color}
\usepackage{colortbl}
\usepackage{booktabs}
\usepackage{subfig}
\usepackage[export]{adjustbox}
\usepackage{enumitem,kantlipsum}

\usepackage{xcolor}

\usepackage{floatrow}
\newfloatcommand{capbtabbox}{table}[][\FBwidth]
\usepackage{blindtext}

\usepackage[pagebackref=true,breaklinks=true,letterpaper=true,colorlinks,bookmarks=false]{hyperref}

\DeclareMathOperator*{\argmax}{argmax}   

\usepackage[pagebackref=true,breaklinks=true,colorlinks,bookmarks=false]{hyperref}

\usepackage{fancyhdr}
\usepackage{cite}

\ifCLASSINFOpdf
\else
\fi
\begin{document}

%
\title{iToF2dToF: A Robust and Flexible Representation for Data-Driven Time-of-Flight Imaging}
%
%
%

\author{Felipe Gutierrez-Barragan*$^{1,2}$
, Huaijin Chen$^{1}$
, Mohit Gupta$^{2}$
, Andreas Velten$^{2}$ 
, and Jinwei Gu$^{1}$ 
\\ $^1$SenseBrain Technology \ \ \ \ \ \ $^2$University of Wisconsin-Madison
\thanks{*Work done during internship at SenseBrain Technology.}
\thanks{FGB, AV, and MG ({\{fgutierrez3, velten\}@wisc.edu, mohitg@cs.wisc.edu}) are with the University of Wisconsin-Madison, Madison, WI 53706. HC and JG (\{chenhuaijin, gujinwei\}@sensebrain.site) are with SenseBrain Technology, San Jose, CA 95131.}
\thanks{Project page: \url{http://pages.cs.wisc.edu/\~felipe/project-pages/2021-itof2dtof/}}
}

\maketitle

\thispagestyle{fancy}

\begin{abstract}
Indirect Time-of-Flight (iToF) cameras are a promising depth sensing technology.
However, they are prone to errors caused by multi-path interference (MPI) and low signal-to-noise ratio (SNR).
Traditional methods, after denoising, mitigate MPI by estimating a transient image that encodes depths. 
Recently, data-driven methods that jointly denoise and mitigate MPI have become state-of-the-art without using the intermediate transient representation.  
In this paper, we propose to revisit the transient representation. 
Using data-driven priors, we interpolate/extrapolate iToF frequencies and use them to estimate the transient image.
Given direct ToF (dToF) sensors capture transient images, we name our method iToF2dToF.
The transient representation is flexible.
It can be integrated with different rule-based depth sensing algorithms that are robust to low SNR and can deal with ambiguous scenarios that arise in practice (e.g., specular MPI, optical cross-talk).
We demonstrate the benefits of iToF2dToF over previous methods in real depth sensing scenarios.

\end{abstract}

\vspace{-0.15in}
\begin{IEEEkeywords}
depth sensing, time-of-flight.
\end{IEEEkeywords}

%
\IEEEpeerreviewmaketitle

\vspace{-0.1in}
\section{Introduction}
\label{sec:1_intro}
\begin{figure*}[t]
    \centering
    \includegraphics[width=.99\textwidth]{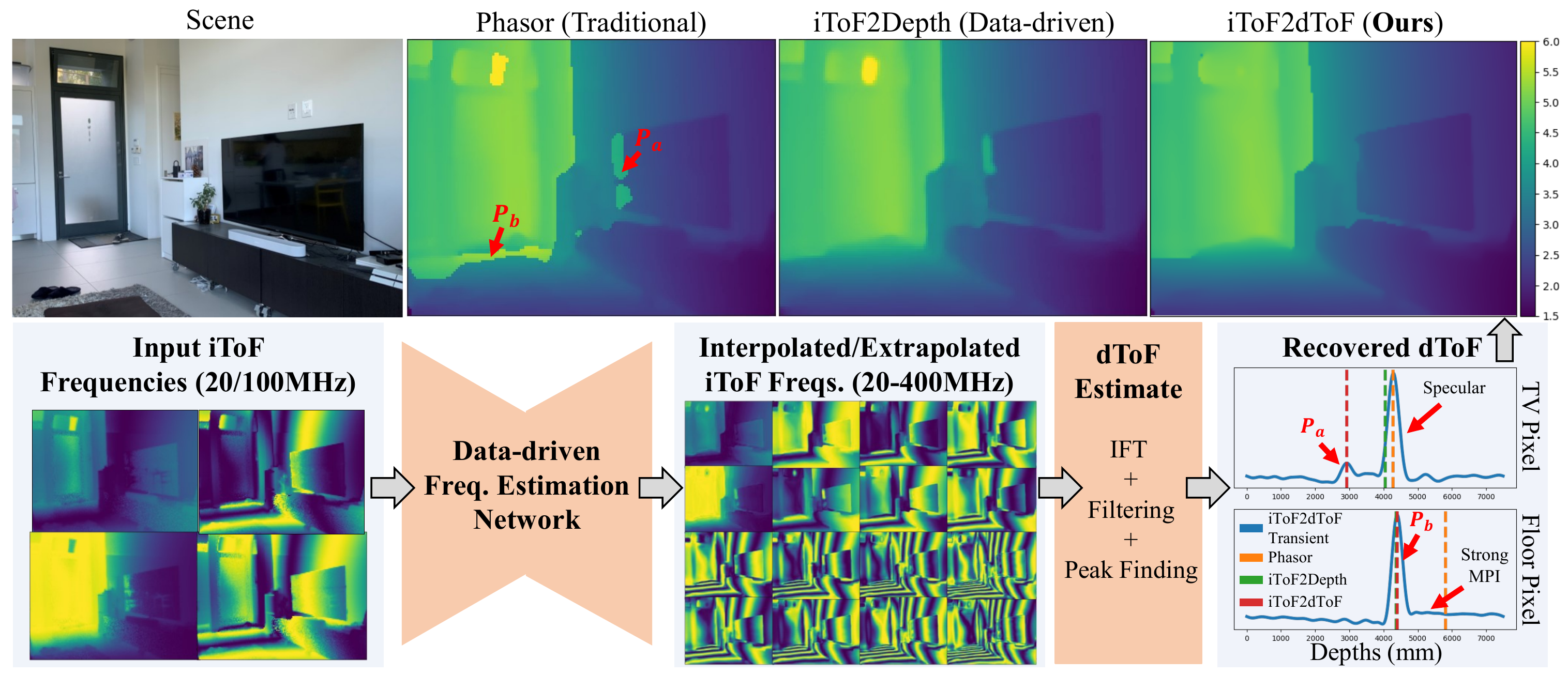}
    \vspace{-0.1in}
    \caption{\textbf{iToF2dToF.}  We compare the iToF-based depth estimation results of previous traditional (Phasor \cite{gupta2015phasor}) and data-driven (iToF2Depth  \cite{su2018deep}) approaches with our proposed data-driven method, iToF2dToF. 
    Previous data-driven approaches map iToF data directly to depths making it hard to adapt to new scenarios such as specular MPI (e.g., TV Pixels). 
    In iToF2dToF we interpolate/extrapolate the captured iToF frequencies and use them to estimate the transient image of the scene. Using rule-based peak finding algorithms, iToF2dToF, produces higher quality depth maps than previous solutions.
    }
    \label{fig:itof2dtof}
\end{figure*}

Indirect Time-of-Flight (iToF) imaging systems measure depths by illuminating the scene with a time-varying periodic signal, and computing the time-shift of the reflected signal~\cite{lange2001solid}.
iToF cameras can achieve high-resolution, while maintaining low-cost, small size and low power consumption, making them a popular choice, especially in resource constrained applications (e.g., Kinect for AR/VR).

Despite these strengths, iToF sensors face several practical challenges; namely, multi-path interference (MPI), and low signal-to-noise ratio (SNR).
MPI arises from the indirect light paths captured by each pixel, which may cause strong depth errors.
For example, points in a concave object may appear farther away, and specular objects or optical cross-talk can cause erroneous depth measurements.
Low SNR, on the other hand, occurs when operated with low exposure times to reduce motion blur and power consumption \cite{chen2020very}.
Overall, practical solutions to MPI and low SNR are critical for emerging iToF-based 3D applications.

Recently, data-driven methods that jointly denoise and correct MPI errors, have been proposed \cite{su2018deep, marco2017deeptof, guo2018tackling, agresti2019unsupervised, qiu2019deep}. 
These solutions do not rely on additional hardware or long acquisition times, making them compact and low-power. 
Furthermore, it is possible to implement them in real-time using light-weight neural networks \cite{marco2017deeptof}, or through model compression and inference acceleration \cite{cheng2018model, howard2017mobilenets}. 

Current data-driven models for iToF cameras perform supervision on the final depthmap representation during training.
This approach works well when the training dataset provides \textit{explicit supervision} for each depth sensing scenario that may arise.
Explicit supervision is sometimes not available in scenarios that are challenging to simulate or capture ground truth (e.g., optical cross-talk), or in ambiguous scenarios, like sparse MPI, where the iToF data can encode multiple distances (e.g., specular or transparent surfaces, and depth discontinuities).
The lack of such explicit supervision for each scenario can impact the generalization of the model.


In this paper, we propose an alternative output representation for data-driven iToF models. 
Similar to previous work, we input sparse iToF frequency measurements to the data-driven model. 
Different from previous approaches, our model outputs interpolated/extrapolated frequencies, on which supervision is performed.
The estimated frequency data is used to estimate dToF/transient\footnote{We use the terms dToF image and transient image interchangeably.} images; therefore, the proposed method is called iToF2dToF.
The dToF representation benefits from the data-driven model's denoising capabilities, and continues to mitigate MPI by separating the direct and indirect illumination \cite{wu2014decomposing}.
Furthermore, the data-driven model does not require explicit supervision on each depth sensing scenario because it only reconstructs the per-pixel dToF waveforms that encode distances in their peaks.
To estimate depths, iToF2dToF is integrated with simple rule-based peak finding algorithms. 
Specifically,


\begin{enumerate}[leftmargin=*, topsep=1pt, partopsep=1pt, itemsep=0pt]

    \item Integrating iToF2dToF with a max peak finding algorithm results in \textit{robust} depth sensing even at low SNR. 
    This design is motivated by its relation to a recent depth decoding method for iToF \cite{gutierrez2019practical} and structured light \cite{mirdehghan2018optimal}.  
    
    \item The \textit{flexible} dToF representation allows us to design simple rule-based algorithms for challenging scenarios, like specular MPI and optical cross-talk, without re-training networks or increasing dataset size.
    
\end{enumerate}

\noindent To summarize, our contributions are:

\begin{itemize}[leftmargin=*, topsep=1pt, partopsep=1pt, itemsep=0pt]

    \item A robust and flexible representation for data-driven iToF that improves depth sensing in different scenarios. 
    
    \item An extensive validation of our data-driven models on real-world data with ground truth. 
    
    \item A synthetic iToF dataset with realistic geometry, textures, and noise.
    
    
    

    
\end{itemize}

\section{Related Work}
\label{sec:2_related}

Mitigating MPI and recovering reliable depths from low SNR measurements are problems that have been extensively studied in the iToF literature \cite{achar2017epipolar, kadambi2016macroscopic, naik2015light, o2014temporal, agresti2018combination, whyte2015resolving, gupta2018optimal, gutierrez2019practical, adam2017bayesian}. 
In this paper, we focus on solutions that could be applied to commercial iToF cameras without hardware modifications. 

\subsection{Transient Imaging}

A \textit{transient image} visualizes a light pulse propagating through a scene \cite{velten2013femto}, where each pixel stores an intensity time profile, referred as a \textit{transient pixel}. 
Usually, a transient pixel encodes depth as the first or max peak \cite{wu2014decomposing}. 
dToF systems capture transient images \cite{o2017reconstructing,raghuram2019storm, warburton2017observation}, enabling accurate depth reconstruction in many scenarios \cite{gupta2019asynchronous, heide2018sub, lyons2019computational, chan2019long}. 


\smallskip

\noindent \textbf{iToF Transient Imaging:} 
Typical iToF systems sample temporal frequencies from the Fourier Transform of the scene's transient image \cite{lin2014fourier}. 
A large number of frequencies are required to recover a transient image.
Therefore, to greatly reduce the number of measurements, iToF image formation models make strong assumptions on the transient structure. 
Previous work can be divided in two categories:

\begin{itemize}[leftmargin=*, topsep=2pt, partopsep=2pt, itemsep=1pt]
    \item \textbf{Sparse Transient Imaging:} 
    One common approach assumes a sparse $K$-path model where each transient pixel is the sum of $K$ delta functions, one for each light path.
    Both analytical \cite{godbaz2012closed,kirmani2013spumic, peters2015solving} and numerical \cite{dorrington2011separating, kadambi2013coded, bhandari2014resolving, freedman2014sra} solutions have been proposed. However, when $K \geq 3$, these methods require many measurements resulting in long acquisition times and high-power consumption.
    
    \item \textbf{Non-Sparse Transient Imaging:} 
    Another family of methods relaxes the sparsity assumption on the transient image, but requires dense frequency sampling.
    Heide et al. \cite{heide2013low} fitted a parametric model to each transient pixel.
    Through Fourier analysis, Lin et al. \cite{lin2014fourier,lin2016frequency}, recovered a transient image.
    More recently, \cite{wang2020model} used a wavelet-based framework that achieved state-of-the-art accuracy, albeit at a higher computational cost. 
    Peters et al. \cite{peters2015solving} proposed a real-time transient reconstruction algorithm that requires as few as 3 frequencies. 
    However, it requires frame averaging due to its sensitivity to noise \cite{wang2020model}. 
\end{itemize}

In this paper we introduce a data-driven non-sparse iToF transient imaging approach for the extreme case where only 2 frequencies are measured.


\subsection{Data-driven ToF Imaging}

Transient rendering \cite{jarabo2014framework,jarabo2017recent,pediredla2019ellipsoidal} has enabled the simulation of ToF datasets, which has been essential for data-driven iToF due to the unavailability of real-world datasets with ground truth.
Broadly, previous data-driven iToF models have used two types of ``Input2Output'' representations: Depth2Depth \cite{marco2017deeptof, agresti2018deep, agresti2019unsupervised, qiu2019deep} and iToF2Depth \cite{su2018deep, guo2018tackling}. 
Depth2Depth models take as input noisy and MPI-corrupted iToF depth images obtained from one \cite{marco2017deeptof} or multiple frequencies \cite{agresti2018deep, agresti2019unsupervised, qiu2019deep}.
iToF2Depth models take as input raw multi-frequency iToF measurements with minimal pre-processing.
Although, a variety of network architectures have been explored for these ``Input2Output'' representations (U-net \cite{marco2017deeptof, su2018deep}, coarse-to-fine CNN \cite{agresti2018deep, agresti2019unsupervised}, and KPN \cite{guo2018tackling, qiu2019deep}, supervision has only been done on the output depth images. 
In this paper, we propose a different ``Input2Output'' representation (i.e., iToF2dToF) where the output, on which supervision is performed, are frequency-domain dToF images.

\section{Image Formation Model}
\label{sec:3_background}

\begin{figure}
\centering
\centerline{
	\includegraphics[width=\linewidth]{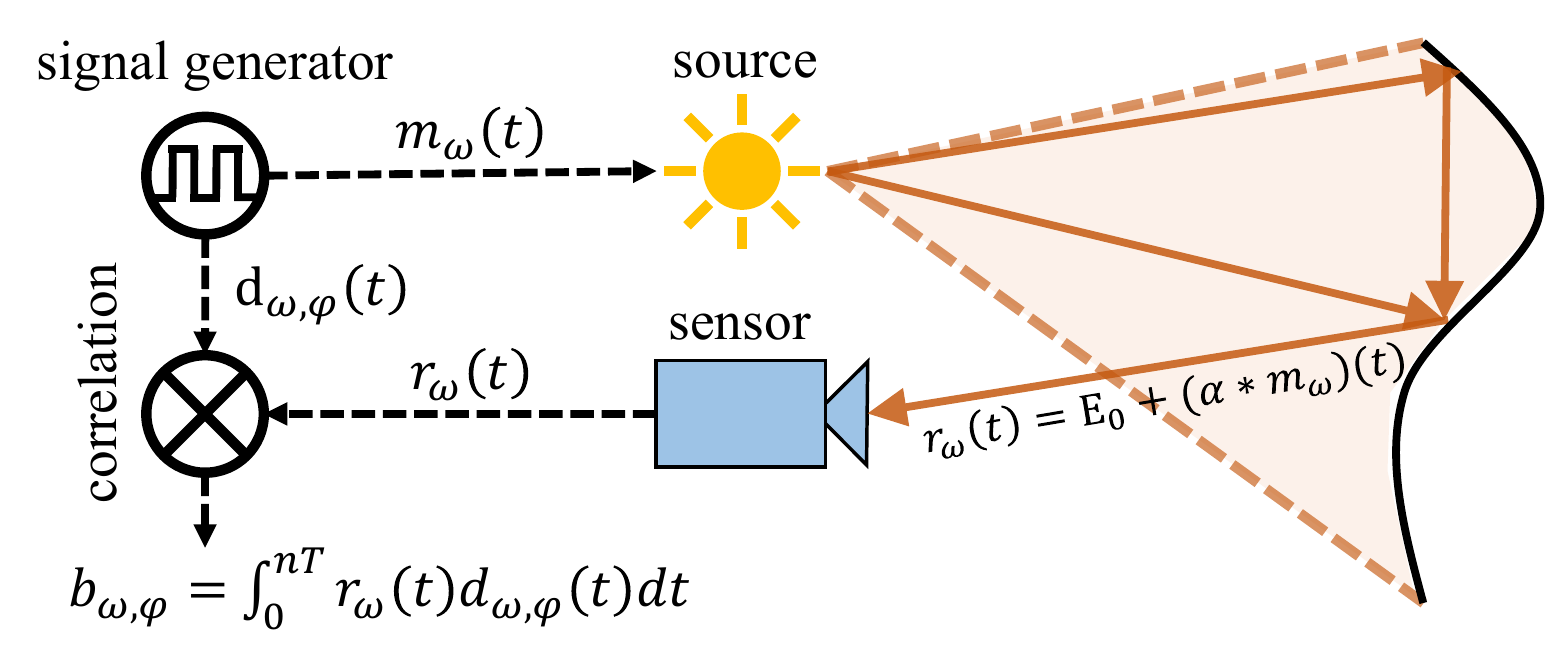}
}
\vspace{-0.1in}
\caption{\textbf{iToF Imaging.} The scene is flash illuminated with an amplitude modulated signal $m_{\omega}(t)$. The radiance, $r_{\omega}(t)$, arriving at each pixel is the sum of $m_{\omega}(t)$ over different light paths. At each pixel, $r_{\omega}(t)$ is correlated with a demodulation function, $d_{\omega, \varphi}(t)$. The product is integrated for an exposure time $nT$ and the resulting intensity, $b_{\omega, \varphi}$, is recorded.}
\label{fig:iToF}
\end{figure}

Let $\alpha(x,y,t)$ be the transient image of a flash illuminated scene (Figure \ref{fig:iToF}), where $\alpha(x,y,t)$ is the intensity of pixel ($x$, $y$) at time $t$. 
We will consider each transient pixel independently so we drop $x$ and $y$. 
The domain of $\alpha(t)$ is on the nanosecond/picosecond scale, and $\alpha(t) \geq 0$ for all $t$. 




ToF systems consist of a light source whose intensity is  modulated with a periodic function $m_{\omega}(t)$ with period $T$.
Here $\omega = 2\pi f$ where $f$ is the repetition frequency.
We assume that $T$ is chosen such that all interesting light paths in $\alpha(t)$ happen at $t < T$, i.e., $\alpha(t) = 0$ for $t \geq T$.
This assumption is necessary because  we cannot recover the general aperiodic form of $\alpha(t)$ \cite{peters2015solving}. 

As illustrated in Figure \ref{fig:iToF}, light propagates to the scene and returns to the sensor along different paths.
The radiance incident on the sensor will be the convolution of $m_{\omega}(t)$ and the transient response of the scene, with a constant offset $E_{0}$ due to ambient light:

\begin{equation}
\small
	r_{\omega}(t) = E_{0} + \int_{0}^{\infty}\alpha(\tau)m_{\omega}(t-\tau)d\tau = E_{0} + (\alpha \ast m_{\omega})(t) 
\label{eq:pixel_radiance}
\end{equation}

A dToF system, measures $\alpha(t)$ using a delta-train modulation signal and sampling $r_{\omega}(t)$ with a high-speed sensor. 

In iToF cameras, $r_{\omega}(t)$ is correlated with the sensor's periodic demodulation function, $d_{\omega, \varphi}(t)$, where $\varphi$ is a controllable phase shift. 
Generally, the correlation is done by modulating the pixel gain according to $d_{\omega, \varphi}(t)$. 
In practice, one way to achieve this, is with dual-tap demodulation pixels \cite{lange2001solid, schwarte1997new} that re-direct the photo-current between two buckets according to $d_{\omega, \varphi}(t)$. 
We model $d_{\omega, \varphi}(t)$ as zero-mean, which is achieved by taking the difference between the two buckets.
The measured brightness for an exposure time, $nT$, would be,

\begin{equation}
    b_{\omega, \varphi} = \int_{0}^{nT} r_{\omega}(t) d_{\omega, \varphi}(t) dt 
    \label{eq:pixel_intensity}
\end{equation}

\noindent Constants in Equation \ref{eq:pixel_intensity} cancel since $d_{\omega, \varphi}(t)$ is zero-mean.


\subsection{Frequency Sampling with iToF Cameras}

In most iToF systems $m_{\omega}(t)$ and $d_{\omega, \varphi}(t)$ are band-limited square functions, modelled as sinusoids.  
If we write $\alpha(t)$ as its Fourier Series between [$0$, $T$], then $r_{\omega}(t)$ becomes:


\begin{align*}
    r_{\omega}(t) &= E_{0} + \alpha(t) \ast (P\cos(\omega t) + P) \\
         &= E_{0} + \sum_{k=0}^{K} A_{\omega_{k}}e^{(i(\omega_{k}t - \phi_{\omega_{k}}))}
         \ast (P\cos(\omega t) + P) \\
         &= A_{\omega}\cos(\omega t - \phi_{\omega}) + L
\end{align*}

\noindent where $\omega_{k} = \omega k$, $k$ is an integer, $P$ is the average power, $L = E_{0} + P\int_{0}^{T} \alpha(t)dt$ is a constant, and $A_{\omega}$, $\phi_{\omega}$ depend on the superposition of the returning sinusoids over all optical paths. In theory, $K$ can be infinite, but in practice it depends on the time discretization. If we take two measurements with $d_{\omega, 0}(t) = \cos(\omega t)$ and $d_{\omega, \frac{\pi}{2}}(t) = \sin(\omega t)$, we get the following brightness measurements:



\begin{equation}
    b_{\omega, 0} = \frac{A_{\omega}}{2}\cos(\phi_{\omega}), \quad b_{\omega, \frac{\pi}{2}} = \frac{A_{\omega}}{2}\sin(\phi_{\omega})
    \label{eq:b_cos_sin}
\end{equation}

In the simple case where the sensor only receives direct illumination, i.e., $\alpha(t) = \delta(t - \frac{2d}{c})$, $\phi_{\omega}$ encodes depth as:

\begin{equation}
    \phi_{\omega} = \frac{2\omega d}{c}
    \label{eq:directonly_depth}
\end{equation}

\noindent However, in the presence of MPI Equation \ref{eq:directonly_depth} will give the incorrect depth. 

\smallskip

\noindent \textbf{Multi-frequency Sampling:} By repeating the measurements in Equation \ref{eq:b_cos_sin} at different $\omega_{k}$ for $k = [1...K]$, we can recover ($A_{\omega_{k}}$, $\phi_{\omega_{k}}$). In other words, $\alpha(t)$ is reconstructed by measuring its Fourier coefficients.  
Since $\alpha(t)$ is computed for $t < T$ the maximum recoverable depth is $d_{\text{max}} = \frac{cT}{2}$. 







\section{The iToF2dToF Framework}
\label{sec:4_itof2dtof}

Current iToF cameras are limited to sampling 2-3 frequencies, where the largest frequency is around $100$MHz. 
This is due to constraints such as power consumption, real-time operation, and bandwidth.
Therefore, reconstruction of $\alpha(t)$ is not possible through dense frequency sampling, unless strong assumptions about $\alpha(t)$ are made. 
In this section we introduce a framework that uses data-driven priors to estimate $\alpha(t)$ from sparse frequency measurements.

\subsection{Data-driven Frequency Estimation}
\label{sec:4_1_freq_superres}

The proposed framework is shown in Figure~\ref{fig:itof2dtof}. First, we concatenate the measured brightness pairs ($b_{\omega_k, 0}$, $b_{\omega_k, \frac{\pi}{2}}$). Brightness images with $n_r$ rows, $n_c$ columns, and $K$ frequencies, will result in a $n_r$x$n_c$x$2K$ tensor $\boldsymbol{B}$. We define a frequency estimation network, $g_\theta$, that interpolates/extrapolates the frequencies up to $\omega_{S}$, where $\omega_{S} \geq \omega_k$:

\begin{equation}
    g_\theta(\boldsymbol{B}) =
    \{ b_{\omega_1, 0}, b_{\omega_1, \frac{\pi}{2}}, 
     \cdots, 
     b_{\omega_{S}, 0}, b_{\omega_{S}, \frac{\pi}{2}}\}  
    \label{eq:network}
\end{equation}

\noindent To avoid phase wrapping ambiguities $\boldsymbol{B}$ includes $\omega_1$.

\begin{figure}
\centering
\includegraphics[width=\linewidth]{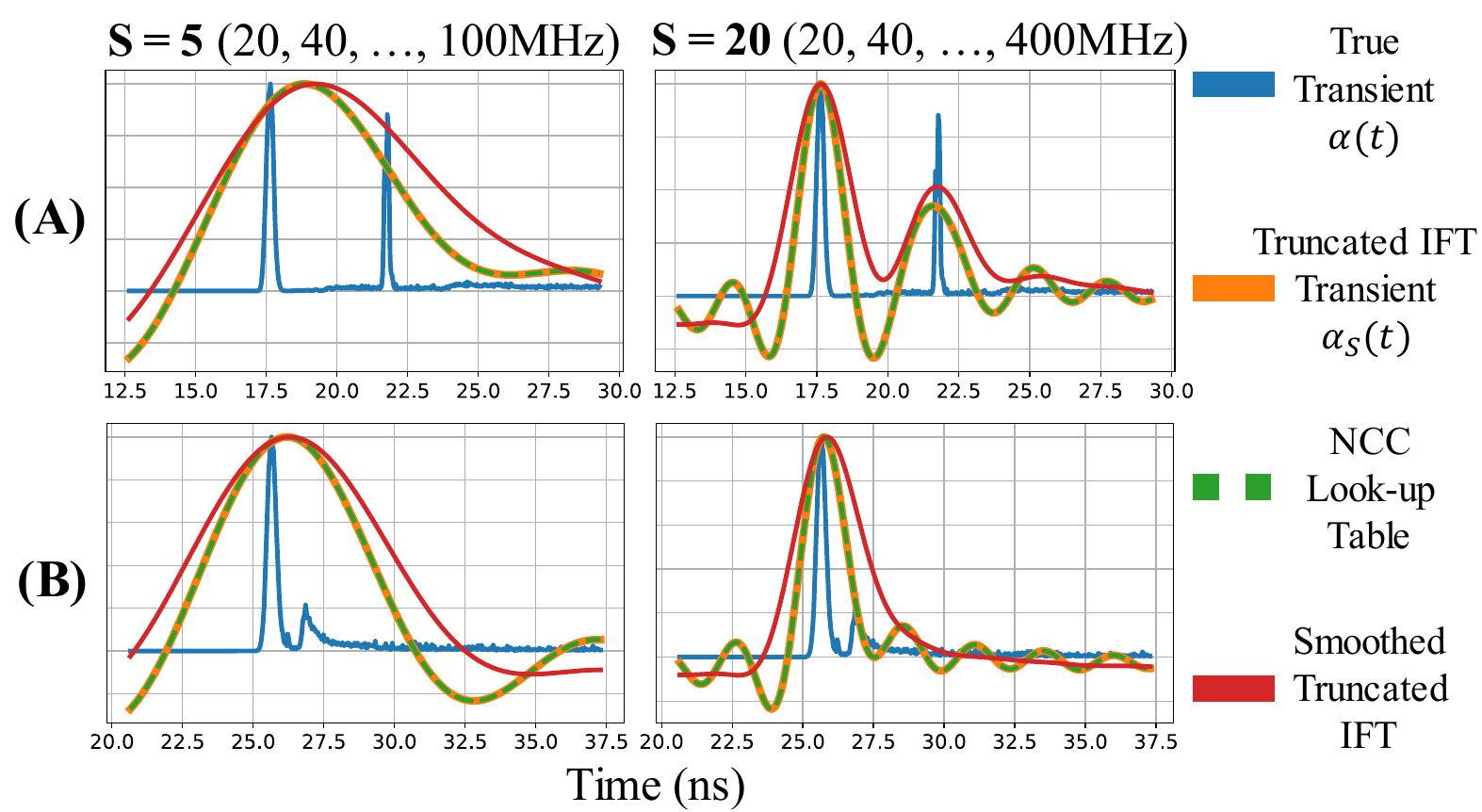}
\caption{\textbf{Truncated IFT}. Examples of transient pixels reconstructed using a truncated IFT. 
A truncated IFT transient, $\alpha_{S}(t)$, with frequencies up to 100MHz ($1^{\text{st}}$ column) is not able to separate sparse MPI peaks (Row A), and diffuse MPI (Row B) introduces a bias in the peak. 
Introducing higher frequencies in the truncated IFT ($2^{\text{nd}}$ column) allows $\alpha_{S}(t)$ to separate the sparse MPI, and the diffuse MPI peak bias becomes smaller. 
} 
\vspace{-0.1in}
\label{fig:truncated_transient_example}
\end{figure}

Why should $g_{\theta}$ be able to learn this form of frequency super-resolution? 
We observe that, although, transient pixel waveforms can have arbitrary shapes, most waveforms have a low-dimensional underlying structure. 
For example, in scenarios where direct illumination is dominant, one frequency is sufficient to estimate the transient pixel, and hence, all of its Fourier coefficients. 
For sparse MPI where there are two peaks (Row (A) in Figure \ref{fig:truncated_transient_example}), two frequencies are sufficient \cite{peters2015solving}. 
Finally, for diffuse MPI (Row (B) in Figure \ref{fig:truncated_transient_example}) the transient pixel can be approximated by a low-dimensional parametric model~\cite{heide2013low}, which reduces the solution's degrees of freedom. 
Based on this observation, we hypothesize and empirically demonstrate that the priors learned by $g_{\theta}$ enable it to interpolate/extrapolate frequencies from a few input frequencies.


\subsection{The dToF Representation}
\label{sec:4_2_dtof_representation}

One way to estimate $\alpha(t)$ from the output Fourier coefficients at $\omega_1, \hdots, \omega_{S}$ is to compute its truncated Fourier Series with the first $S$ terms:

\begin{equation}
    \alpha_{S}(t) = 
    \sum_{s=1}^{S} A_{\omega_{s}}e^{(i(\omega_{s}t - \phi_{\omega_{s}}))}
    \label{eq:truncated_series}
\end{equation}

\noindent Equation \ref{eq:truncated_series} can be implemented by taking the Inverse Fourier Transform (IFT) of the estimated brightness pairs in Equation \ref{eq:network}, with ($b_{\omega_s, 0}$, $b_{\omega_s, \frac{\pi}{2}}$) as the real and imaginary parts of $\omega_s$. 
A truncated IFT will contain Gibbs ringing artifacts as shown in Figure \ref{fig:truncated_transient_example}. 
These artifacts are not an issue if we only need the maximum peak.
However, to find other peaks, it is useful to apply a smoothing function, like a Hamming window \cite{harris1978use}, that reduces the ringing but does not completely remove higher frequencies (red line in Figure \ref{fig:truncated_transient_example}). 
Moreover, besides a truncated IFT, other transient estimation methods like the ones discussed in Section \ref{sec:2_related} are applicable.
We limit our analysis to the truncated IFT due to its simplicity, minimal assumptions, and relation to a robust depth decoding method discussed next. 

\smallskip

\noindent \textbf{Robust Depth Estimation:} 
Given $\alpha_{S}(t)$, assuming diffuse reflections, depths are encoded by the maximum peak location, i.e., $\argmax \alpha_{S}(t)$.
In other words, $\alpha_{S}(t)$ is a depth lookup table. 
It turns out that $\alpha_{S}(t)$ has an interesting connection to the lookup table used in the normalized cross-correlation (NCC) depth decoding algorithm, originally proposed in the structured light context \cite{mirdehghan2018optimal, chen2020auto}, and recently applied to iToF \cite{gutierrez2019practical}. 
In our case, the projected/reference signals are sinusoids at multiple frequencies, and the observed signals are the brightness measurements estimated in Equation \ref{eq:network}. 
The NCC between the reference and observed signals results in the NCC depth lookup table (green line in Figure \ref{fig:truncated_transient_example}).
We observe that the NCC lookup table is equivalent (up to a scale) to $\alpha_{S}(t)$ (orange line in Figure \ref{fig:truncated_transient_example}). 
Assuming a direct-only illumination model, NCC decoding is near-optimal in the presence of additive white Gaussian noise \cite{mirdehghan2018optimal}. 
Therefore, given Equation \ref{eq:network} $2S$ brightness measurements, we can use the max peak of the NCC lookup table or $\alpha_{S}(t)$ for noise-robust depth estimation.
Interestingly, despite the direct-only assumption in NCC decoding, the lookup table will account for direct and indirect light paths, given enough frequencies.
This means that, although, NCC's optimality may not hold for pixels with MPI, it will still mitigate MPI.


\smallskip

\noindent \textbf{Flexible Depth Estimation:} 
In certain challenging depth sensing scenarios, like sparse MPI, the maximum peak may not encode depth.
Examples of these scenarios include: specular MPI, optical cross-talk, and depth discontinuities.
Some of these scenarios may not be modeled in the training set (e.g., optical cross-talk), or in the case of depth discontinuities the correct peak may be application-dependent (see Supplementary Section 5.C for an example), making them challenging to resolve using a data-driven model.
Fortunately, the dToF response that arises in all of these situations follows a similar sparse 2-path model, which will be well-represented in the dataset at glossy surfaces (e.g., wood floor in Figure \ref{fig:dataset_overview}) and at edge pixels. 
Therefore, the data-driven model in iToF2dToF should be able to generalize and predict the correct sparse dToF response. 
Finally, we can design rule-based peak finding algorithms around the flexible dToF representation to recover the correct depths in these situations.


\section{Datasets and Implementation}
\label{sec:5_datasets_and_implementation}

In this section, we introduce our simulator and new synthetic Multi-Frequency ToF (MF-ToF) dataset. 
We also describe the acquired real datasets. 
Lastly, we discuss the architecture used for iToF2dToF and the baseline models.

\subsection{Simulator and Synthetic Datasets}
\label{sec:5_simulator_and_synthetic_datasets}

We implemented a data generation pipeline and created the MF-ToF dataset. 
Table \ref{tab:synthetic_dataset_comparison} outlines the aspects that our dataset improves over previous datasets \cite{su2018deep, marco2017deeptof, guo2018tackling, agresti2019unsupervised}.

\smallskip

\noindent \textbf{iToF Simulator:} We render transient images using MitsubaToF \cite{pediredla2019ellipsoidal, jakob2010mitsuba}. 
The iToF simulator takes as input the transient images, sensor parameters, and frequency, to produce 4-phase sinusoid measurements. 
Assuming sinusoid functions works well in practice because commercial iToF modules come calibrated. 
Finally, read and photon noise are added, and we check for saturation. 
To validate the simulator's signal and noise levels we use real iToF data to tune the simulator parameters (see Supplementary Section 1).

\smallskip

\begin{figure}
\vspace{-0.1in}
\centering
\includegraphics[width=\linewidth]{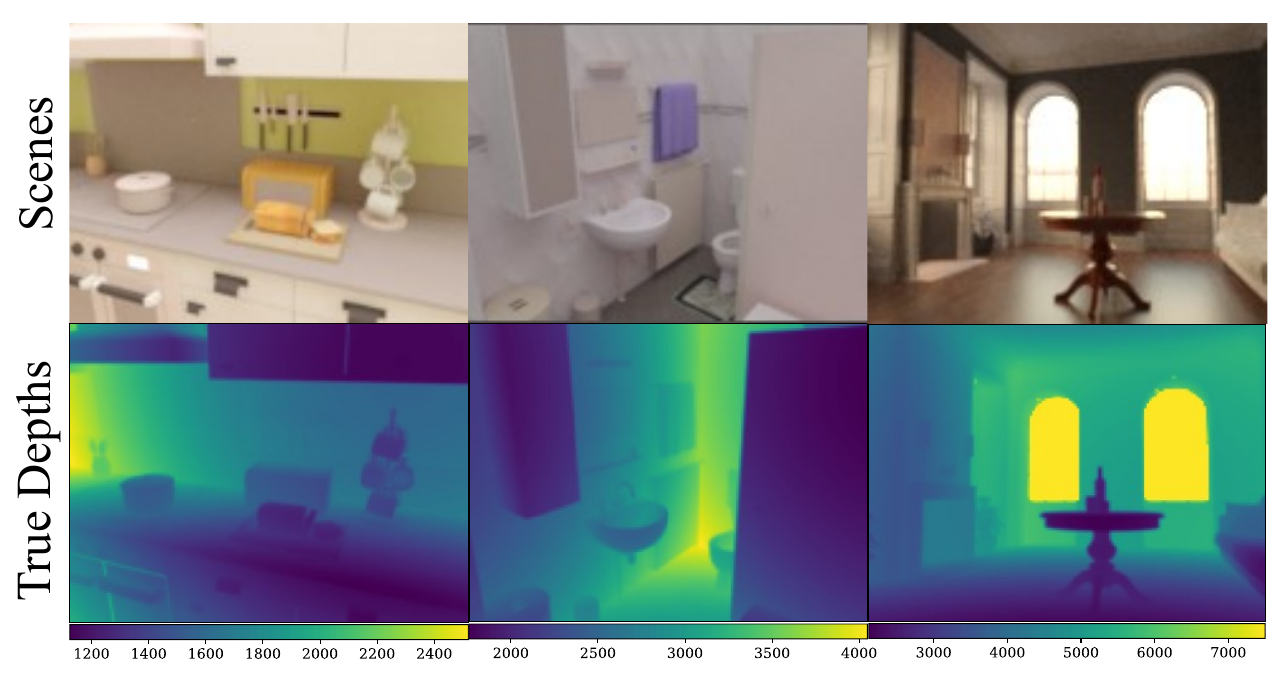}
\caption{\textbf{Synthetic Scenes} from the MF-ToF dataset.}
\label{fig:dataset_overview}
\vspace{-0.1in}
\end{figure}

\noindent \textbf{3D Scenes and View Generation:} 
We gathered 14 Mitsuba scene models from \cite{bitterli2016rendering}, and 11 Blender scene models from \cite{blendswap2016} which we convert to Mitsuba. 
The scenes have realistic layouts, textures/albedos, and diffuse and glossy materials as seen in Figure \ref{fig:dataset_overview}. 
For each scene, we semi-automatically generate 200 views used to render 5000 transient images. 

\medskip

\noindent \textbf{MF-ToF Dataset:} 
We simulate 5000 iToF images containing frequencies from 20MHz to 600MHz in steps of 20. 
To cover a wide range of SNR levels we simulate and average multiple frames. 
According to our simulator validation, averaging 2 frames has similar noise levels to an exposure time of 0.3ms in the real sensor. 
Therefore, for each image the number of averaged frames is randomized between 1 and 12 to emulate exposure times ranging from 0.15-1.8ms.
We use this dataset for training.
We simulate 4 more datasets with varying number of averaged frames to evaluate the performance of each model at different SNR levels (Figure \ref{fig:simulation_noise}).
The dataset has a 120x160 spatial resolution.

For more details about the simulator and the synthetic datasets, please refer to Supplementary Section 2.

\begin{table}
\caption{\textbf{Synthetic iToF Dataset Comparison}.}
\label{tab:synthetic_dataset_comparison}
\resizebox{\linewidth}{!}
{
\begin{tabular}{cccccc}
\hline
                                                                                    & \textbf{\begin{tabular}[c]{@{}c@{}}FLAT\\ \cite{guo2018tackling}\end{tabular}} & \textbf{\begin{tabular}[c]{@{}c@{}}DeepToF\\ \cite{marco2017deeptof}\end{tabular}} & \textbf{\begin{tabular}[c]{@{}c@{}}E2E ToF\\ \cite{su2018deep} \end{tabular}} & \textbf{\begin{tabular}[c]{@{}c@{}}DA CNN\\ \cite{agresti2019unsupervised} \end{tabular}} & \textbf{\begin{tabular}[c]{@{}c@{}}MF-ToF\\ (\textbf{Ours})\end{tabular}}                                \\ \hline
\textbf{\begin{tabular}[c]{@{}c@{}}Number of Unique\\ 3D Scene Models\end{tabular}} & \cellcolor[HTML]{C1DFC1}70                                                & \cellcolor[HTML]{FFFACF}25                                                      & \cellcolor[HTML]{F0BEBF}5                                                        & \cellcolor[HTML]{FFFACF}35                                                           & \cellcolor[HTML]{FFFACF}25                                                                         \\ \hline
\textbf{\begin{tabular}[c]{@{}c@{}}Scene Layout \\ Realism\end{tabular}}            & \cellcolor[HTML]{F0BEBF}Low                                               & \cellcolor[HTML]{C1DFC1}High                                                    & \cellcolor[HTML]{C1DFC1}High                                                     & \cellcolor[HTML]{C1DFC1}High                                                         & \cellcolor[HTML]{C1DFC1}High                                                                       \\ \hline
\textbf{\begin{tabular}[c]{@{}c@{}}Scene Textures \\ Realism\end{tabular}}          & \cellcolor[HTML]{F0BEBF}Low                                               & \cellcolor[HTML]{F0BEBF}Low                                                     & \cellcolor[HTML]{F0BEBF}Low                                                      & \cellcolor[HTML]{C1DFC1}High                                                         & \cellcolor[HTML]{C1DFC1}High                                                                       \\ \hline
\textbf{\begin{tabular}[c]{@{}c@{}}Non-Diffuse \\ Materials\end{tabular}}           & \cellcolor[HTML]{F0BEBF}None                                              & \cellcolor[HTML]{F0BEBF}None                                                    & \cellcolor[HTML]{F0BEBF}None                                                     & \cellcolor[HTML]{F0BEBF}None                                                         & \cellcolor[HTML]{FFFACF}\begin{tabular}[c]{@{}c@{}}Plastic - described \\ in Mitsuba \cite{jakob2010mitsuba} \end{tabular} \\ \hline
\textbf{\begin{tabular}[c]{@{}c@{}}Unique Camera \\ Views per Scene\end{tabular}}   & \cellcolor[HTML]{FFFACF}30                                                & \cellcolor[HTML]{FFFACF}7                                                       & \cellcolor[HTML]{C1DFC1}250                                                      & \cellcolor[HTML]{F0BEBF}2                                                            & \cellcolor[HTML]{C1DFC1}200                                                                        \\ \hline
\textbf{\begin{tabular}[c]{@{}c@{}}Total Images w/out \\ Augmentation\end{tabular}} & \cellcolor[HTML]{C1DFC1}2100                                              & \cellcolor[HTML]{F0BEBF}175                                                     & \cellcolor[HTML]{FFFACF}1250                                                     & \cellcolor[HTML]{F0BEBF}70                                                           & \cellcolor[HTML]{C1DFC1}5,000                                                                      \\ \hline
\textbf{\begin{tabular}[c]{@{}c@{}}Simulator \\ Validation\end{tabular}} & \cellcolor[HTML]{C1DFC1}Yes                                              & \cellcolor[HTML]{F0BEBF}No                                                     & \cellcolor[HTML]{F0BEBF}No                                                     & \cellcolor[HTML]{F0BEBF}No                                                           & \cellcolor[HTML]{C1DFC1}Yes                                                                      \\ \hline
\textbf{\begin{tabular}[c]{@{}c@{}}Number of \\ Frequencies \end{tabular}} & \cellcolor[HTML]{F0BEBF}3                                              & \cellcolor[HTML]{F0BEBF}1                                                     & \cellcolor[HTML]{F0BEBF}2                                                     & \cellcolor[HTML]{F0BEBF}3                                                           & \cellcolor[HTML]{C1DFC1}30                                                                      \\ \hline
\end{tabular}
}
\vspace{-0.1in}
\end{table}

\subsection{Real Datasets}
\label{sec:5_real_datasets}

We use an iToF module with 480x640 resolution for data collection. 
All images are resampled to 120x160 to match the MF-ToF dataset resolution.
The sensor is configured and calibrated to take measurements at 20MHz and 100MHz, and raw data access is available at these frequencies. 
In addition to real-world scenes like the ones in Figures \ref{fig:itof2dtof} and \ref{fig:xtalk}, we acquire scenes in controlled scenarios with ground truth depths. Namely,

\begin{itemize}[leftmargin=*, topsep=2pt, partopsep=2pt, itemsep=1pt]
    \item \textbf{Wall scenes:} We imaged a flat wall at 8 depths (0.25-2m in steps of 0.25m), at 3 exposures (0.15, 0.3, 0.5ms), and with 3 surface reflectances (5, 20, 50\%). To verify the true depth, we fixate a cross-shaped laser distance sensor next to the iToF module. For alignment, we attached a mirror to the wall and made sure that the reflected laser pointer returned to the emitting point, ensuring that the sensor and wall plane were parallel.
    
    \item \textbf{MPI scenes w/ ground truth:} We captured 7 real-world scenes with partial ground truth. Figure \ref{fig:real_mpi_errors} shows 2 of them. To obtain ground truth depths we average 100 frames of an empty scene with negligible MPI (e.g., a wall or an empty table as in Figure \ref{fig:real_mpi_errors}). We use the noise-free data to compute ground truth depths with the Phasor method described in Section \ref{sec:6_1_ablation}. Afterwards, we capture the same scene a second time with objects placed in it. We manually create a mask indicating the regions where ground truth depths can be obtained from the first capture (see Figure \ref{fig:real_mpi_errors}). Each scene was captured at 4 exposures (1, 0.5, 0.2, 0.1ms), to cover a wide range of SNR levels.

\end{itemize}

\subsection{Training and Implementation}
\label{sec:5_training_and_implementation}

The inputs to the network are frequency measurements at 20 and 100MHz, where each frequency has 2 input channels ($b_{\omega, 0}$, $b_{\omega, \frac{\pi}{2}}$). 
In the iToF2dToF models the network outputs the interpolated and extrapolated frequencies up to a given maximum frequency. 
We normalize each frequency by dividing by its amplitude \cite{su2018deep}. 
Although, we find that using amplitude information improves the performance of all models, we also observe that models trained with synthetic amplitude information did not generalize well to real data.
Extensive calibration \cite{guo2018tackling} or more complex architectures \cite{agresti2019unsupervised} can help. 
Nonetheless, the benefits of iToF2dToF hold with or without amplitude information (see Supplementary Section 6.B). 



The network architecture used for all iToF2dToF and baseline models was a U-net \cite{ronneberger2015u} with skip connections, ReLU activations, and learned upsampling, implemented in PyTorch \cite{paszke2019pytorch}. 
The number of model parameters ranges between 1.873-1.877M depending on the number of output channels.
We minimize the L1 loss between the U-net's output and the target output.
For a detailed description of the U-net architecture, please refer to Supplementary Section 2.

For training we split the dataset into train, valid, and test sets. The training set is composed of all views generated from 21 scenes (4200 instances). The validation and testing sets are each composed of views generated from 2 scenes. 
Unless stated otherwise, we train all models using the ADAM optimizer \cite{kingma2014adam} with a constant learning rate of 0.0001 for 300 epochs, followed by 700 epochs with a linearly decaying learning rate. 
We use a batch size of 32. 
During training, we mask pixels with depths greater than 7.5m (max depth for 20MHz) and with ``infinite'' depths corresponding to pixels looking at empty space (e.g., open windows or doors). 
During testing we remove images that contain these invalid pixels. 
Furthermore, during training, we apply random flips for data augmentation. 
At each epoch we evaluate the model's loss on the validation set and we keep track of the model with the lowest validation loss.

\section{Experiments and Results}
\label{sec:6_results}

In this section, we perform an ablation study to find the maximum frequency iToF2dToF can effectively extrapolate to, and compare it to multiple traditional and data-driven iToF models. 
The synthetic test set has 182 instances. 
We evaluate the robustness to noise on real and synthetic data. 
Finally, we demonstrate the flexibility of iToF2dToF on two real examples of specular MPI and optical cross-talk.

\smallskip

\noindent \textbf{Performance Metric:} To quantify performance we use percentile mean absolute errors (MAE), similar to \cite{adam2017bayesian, su2018deep}. For each image, the depth errors are sorted from lowest to highest, divided into 4 percentile groups (0-75\%, 75-85\%, 85-95\%, 95-99\%), and the MAE is calculated within each group. This grouping allows to understand the performance in high SNR/low MPI (i.e., 0-75\% percentile) and in low SNR/high MPI (i.e., 85-99\% percentile) regions. When calculating the percentile MAE on synthetic data, we mask edge pixels with depth discontinuities (i.e., flying pixels) because the ground truth depth is not reliable. We do not consider pixels with the largest 1\% error to avoid invalid pixels that arise from unmasked flying pixels, or infinite ray pixels (may occur at the intersection of two meshes).
\subsection{Baseline Comparison and Ablation Study}
\label{sec:6_1_ablation}

\begin{table}
\centering
\setlength\tabcolsep{3.25pt}
\resizebox{\linewidth}{!}
{
\begin{tabular}{lcccc}
\hline
\multicolumn{5}{c}{\textbf{Synthetic Test Set Percentile MAE (mm)}}                                                                                 \\ \hline
\textbf{Model}                                             & \textbf{0-75\%} & \textbf{75-85\%} & \textbf{85-95\%} & \textbf{95-99\%} \\ \hline
Phasor (No Noise) \cite{gupta2015phasor}                                                  & 9.53           & 29.58            & 46.37            & 94.79            \\ \hline
SRA (No Noise) \cite{freedman2014sra}                                                  & 13.97           & 41.75            & 62.70            & 126.51            \\ \hline
Depth2Depth Baseline                                                & 13.40           & 42.21   & 69.07   & 136.96            \\ \hline
iToF2iToF Baseline                                                 & 10.56           & 31.93            & 49.71            & 101.82            \\ \hline
iToF2Depth Baseline                                                & 7.49           & 21.86   & 34.99   & 88.03            \\ \hline
iToF2dToF @140MHz                                                  & 9.23   & 28.08            & 44.31            & 91.53   \\ \hline
iToF2dToF @200MHz                                                  & 7.85   & 23.26            & 36.69            & 78.60   \\ \hline
iToF2dToF @300MHz                                                  & 7.41   & 21.23            & 33.33            & 74.68   \\ \hline
iToF2dToF @400MHz                                                  & \textbf{7.19}   & 20.42            & 32.18            & \textbf{71.56}   \\ \hline
iToF2dToF @500MHz                                                  & 7.22   & \textbf{20.40}            & \textbf{32.17}            & 72.12   \\ \hline
iToF2dToF @600MHz                                                  & 7.33   & 20.66           & 32.76            & 76.13   \\ \hline
\end{tabular}
}
\caption{Percentile MAE calculated on the simulated test set containing a wide range of SNR levels. 
Training iToF2dToF to extrapolate to higher frequencies improves performance up to around 400MHz. 
}
\label{tab:synthetic_test_results}
\vspace{-0.1in}
\end{table}

We evaluate iToF2dToF with different maximum frequencies and compare it against the following approaches:

\begin{itemize}[leftmargin=*, topsep=2pt, partopsep=2pt, itemsep=1pt]
    \item \textbf{Phasor: } Depths are computed using a lookup table approach often used for multi-frequency iToF \cite{gupta2015phasor, gutierrez2019practical}.
    \item \textbf{SRA: } Depths are computed using the sparse reflections analysis technique introduced in \cite{freedman2014sra}.
    \item \textbf{Depth2Depth: } A network trained to remove noise and MPI errors from the depths produced by Phasor. This is a similar model to \cite{marco2017deeptof}, without the pre-training stage.
    \item \textbf{iToF2Depth: } A network trained to map noisy iToF data to denoised and MPI-free depths. 
    This is a similar model to \cite{su2018deep}, but with a smaller network. 
    This model required more training to converge, so instead of 700 epochs of linearly decaying learning rate, we use 1700.
    Throughout this paper the terms, ``iToF2Depth'' and ``iToF2Depth Baseline'', refer to the same model. 
    In Supplementary Section 6.C, we compare with additional iToF2Depth models that use a larger network and additional regularization, as in \cite{su2018deep}.
    
    \item \textbf{iToF2iToF Baseline: } A network trained to denoise the input iToF measurements. 
\end{itemize}

Table \ref{tab:synthetic_test_results} compares the performance of iToF2dToF and the aforementioned baselines. 
All the data-driven models use the U-net architecture described in Section \ref{sec:5_training_and_implementation}. 
Performance of iToF2dToF improves as we increase the maximum frequency it extrapolates to. 
Around 400MHz performance plateaus, and further extrapolation does not lead to better results. 
For the remainder of the paper the iToF2dToF model extrapolates up to 400MHz. 
Please refer to Supplementary Section 6.A for additional results showcasing the effect of frequency on depth errors and also the recovered transient signals.

Most data-driven approaches are able to correct for MPI to some degree, as seen in Table \ref{tab:synthetic_test_results}. 
This is evident by comparing their metrics with Phasor, whose depth errors are solely due to MPI because we do not add noise to it.
Additionally, we observe poor performance by the SRA model. 
This is likely due to SRA's K-path modeling assumption not being true for scenes without reflective objects, and also only providing 2 frequencies. 
Similar to \cite{su2018deep}, we find that Depth2Depth with a U-net architecture performs worse than a U-net that uses the raw measurements as input (i.e., iToF2Depth). 
Overall, iToF2dToF outperforms all the evaluated models. 
Other MPI correction methods \cite{kadambi2013coded, peters2015solving, heide2013low, lin2014fourier, achar2017epipolar, bhandari2014resolving} require special acquisition strategies or more frequencies, making them difficult to compare with. 

\begin{figure}[t!]
\centering
\centerline{
	\includegraphics[width=\linewidth]{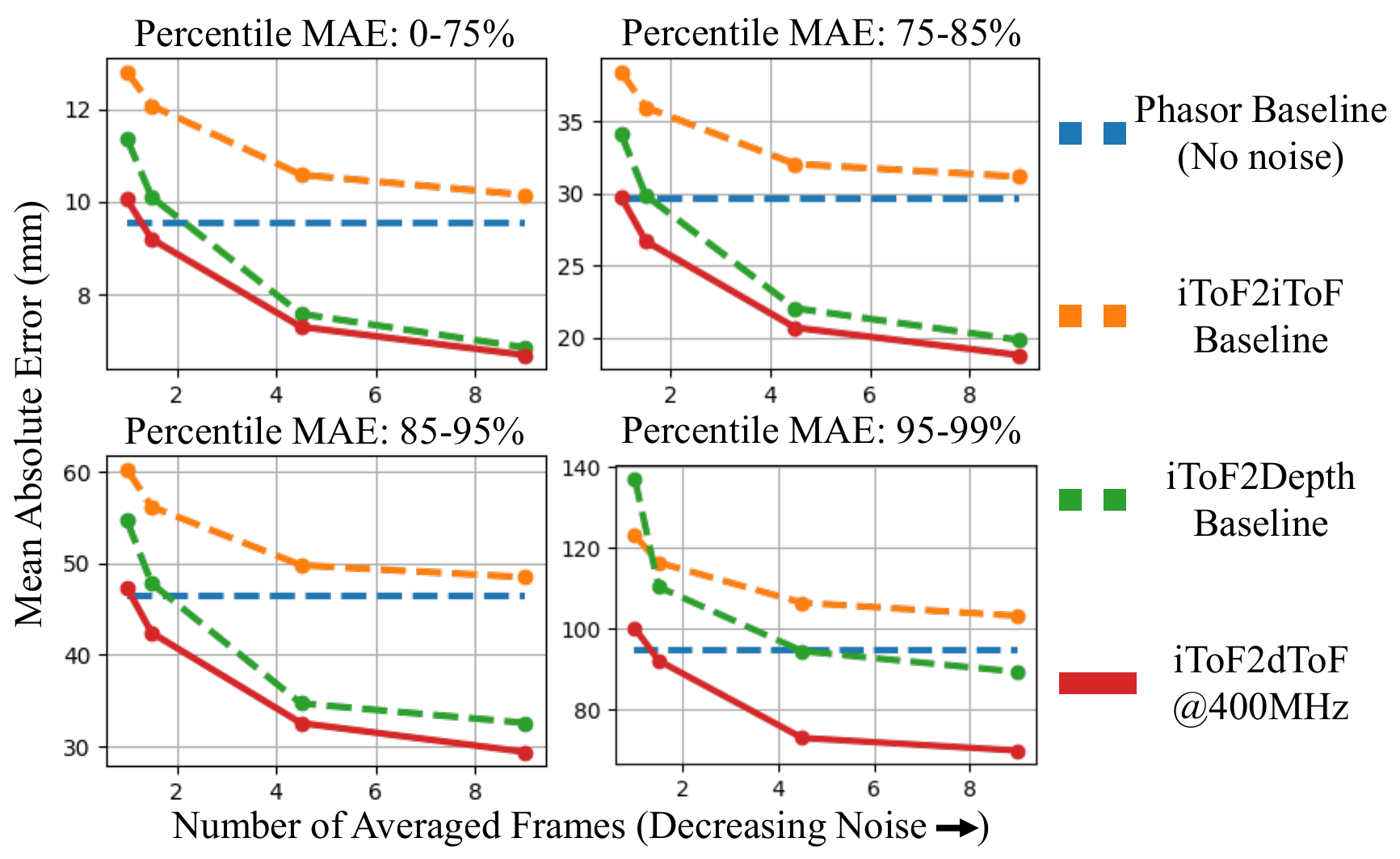}
}
\caption{\textbf{Noise vs. Errors}. We simulate the same test set at varying SNR levels as described in Section \ref{sec:5_simulator_and_synthetic_datasets}. 
Each model was trained with images containing the full range of SNR levels. 
The Phasor baseline does not contain noise and is only shown to confirm the MPI correction capabilities of the data-driven models. 
}
\label{fig:simulation_noise}
\vspace{-0.1in}
\end{figure}

\begin{figure}
\vspace{-0.1in}
\centering
\includegraphics[width=\linewidth]{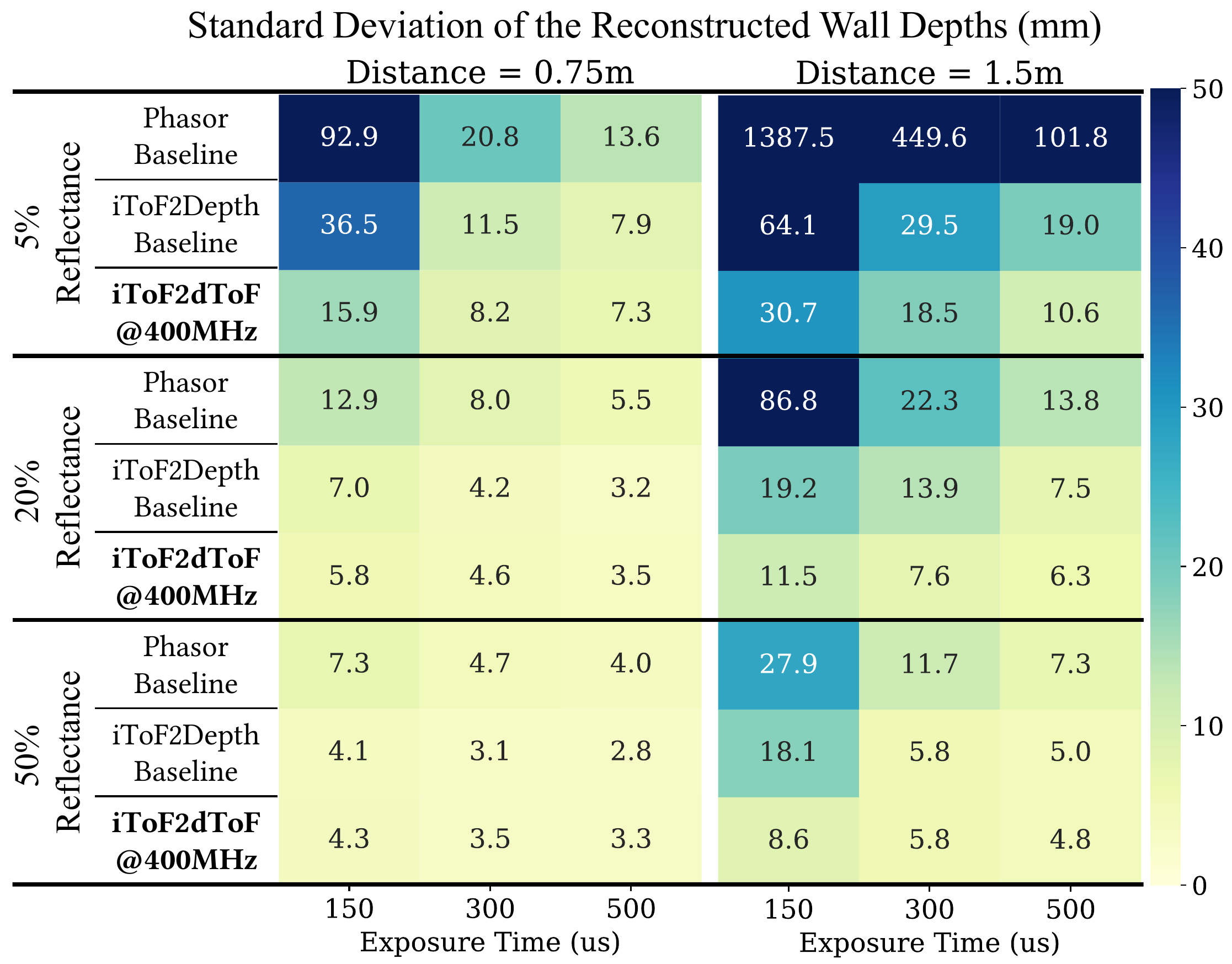}
\caption{\textbf{Wall Depths Standard Deviation}. 
As described in Section \ref{sec:5_real_datasets}, we image a wall at different distances, under 9 combinations of exposure and reflectance settings.
The reconstructed depths by iToF2dToF exhibit the lowest standard deviation at most SNR levels, particularly at low SNR.
}
\label{fig:wall_depth_std}
\vspace{-0.1in}
\end{figure}

\begin{figure*}
\centering
\includegraphics[width=\textwidth]{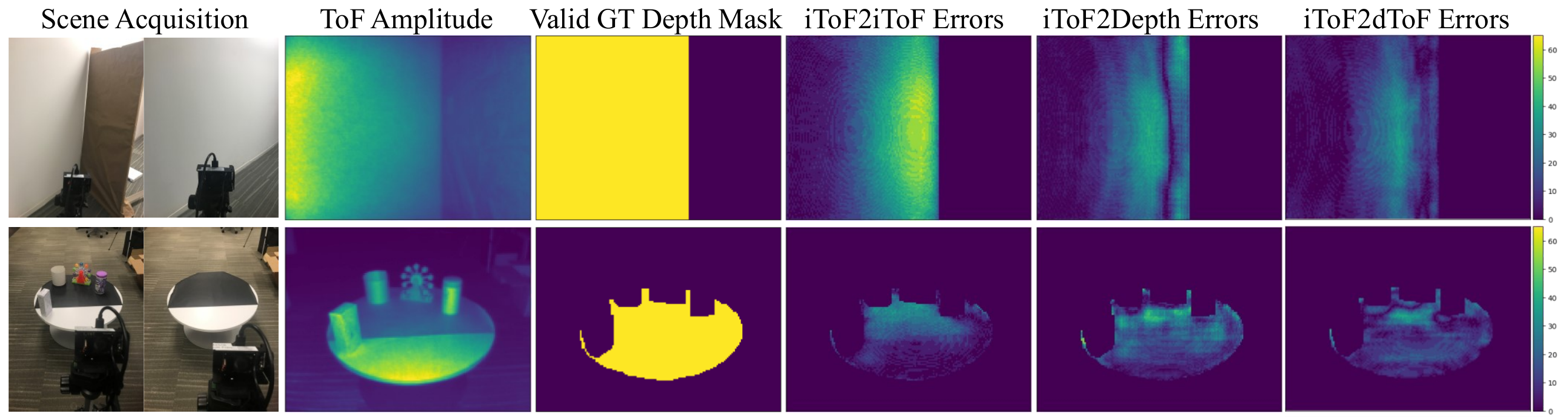}
\caption{\textbf{Real-world MPI Correction.} 
To quantitatively evaluate the MPI correction capabilities of each method, we acquire scene pairs as shown in the left-most images. 
We capture the scene with and without objects in it, and use the depths from the scene without objects as the ground truth depths. 
We manually draw masks that indicate regions where ground truth depths are known. 
In both scenes, captured with a 1ms exposure time, iToF2dToF achieves the lowest depth errors.  
}
\label{fig:real_mpi_errors}
\vspace{-0.1in}
\end{figure*}

\subsection{Robustness of iToF2dToF}
\label{sec:6_2_robustness}

In this section, we demonstrate the performance of iToF2dToF in scenarios with a wide range of SNR levels. 

\smallskip

\noindent \textbf{Robustness on Synthetic Data: } Figure \ref{fig:simulation_noise} shows performance at different SNR levels. 
As we decrease the test set SNR, the performance gap of iToF2dToF and the end-to-end model (iToF2Depth) increases. 
Surprisingly, in the lowest SNR setting iToF2Depth performs worse or comparably to a denoising network (iToF2iToF). 
We find that using a larger network for iToF2Depth, helps improve its robustness to noise. Nonetheless, as shown in Supplementary Section 6.C, the larger network continues to perform worse than iToF2dToF at low SNR. 



\smallskip

\noindent \textbf{Robustness in Real Wall Dataset: }
Figure \ref{fig:wall_depth_std} shows standard deviation of the reconstructed depths for a planar wall at different SNR levels. 
We apply light gaussian smoothing on the raw data for Phasor. 
At high SNR levels (i.e., high exposure, small distance, and high reflectance), iToF2dToF performs comparably to other learning-based baselines and achieve a low standard deviation ($<5$mm) in the reconstructed depths. 
This is expected because all reasonable models
should converge to the correct solution as we increase SNR, especially for scenes with no MPI like the wall reconstructions.
At low SNR settings, however, iToF2dToF consistently achieves $\sim$2x lower standard deviation. 
In Supplementary Section 3 we show results with a similar trend at more distances.

\smallskip

\noindent \textbf{Robustness in Real-world Data: }
Table \ref{tab:real_test_results} summarizes the percentile MAE obtained from 28 scenes (7 scenes, 4 exposures per scene) with partial ground truth, as described in Section \ref{sec:5_real_datasets}. Figure \ref{fig:real_mpi_errors} shows the depth errors for 2 of the them. 
Quantitatively, iToF2dToF achieves lower percentile MAE than all baselines, and outperforms iToF2Depth in 6 out of 7 scenes. 
Qualitatively, Figure \ref{fig:real_qualitative_depths} shows that iToF2dToF recovers depth images with fewer artifacts.
In Supplementary Sections 4 and 5, we present extensions of Figures \ref{fig:real_mpi_errors} and \ref{fig:real_qualitative_depths} for all acquired scenes and additional qualitative results.

\begin{table}
\centering
\setlength\tabcolsep{3.25pt}
\resizebox{\linewidth}{!}
{
\begin{tabular}{lcccc}
\hline
\multicolumn{5}{c}{\textbf{Real-world Test Set Percentile MAE (mm)}}                                                                                 \\ \hline
\textbf{Model}                                             & \textbf{0-75\%} & \textbf{75-85\%} & \textbf{85-95\%} & \textbf{95-100\%} \\ \hline
Phasor \cite{gupta2015phasor}                                                  & 42.51           & 234.93            & 440.18            & 1150.66            \\ \hline
iToF2iToF Baseline                                                 & 10.12           & 32.58            & 43.21            & 68.31            \\ \hline
iToF2Depth Baseline                                                & 7.70           & 21.82   & 30.46   & 65.05            \\ \hline
iToF2dToF @400MHz                                                  & \textbf{6.64}   & \textbf{19.46}            & \textbf{27.57}            & \textbf{60.29}   \\ \hline
\end{tabular}
}
\vspace{-0.1in}
\caption{Percentile MAE for real-world scenes with ground truth captured as in Figure \ref{fig:real_mpi_errors}. 
}
\label{tab:real_test_results}
\end{table}

\begin{figure}
\vspace{-0.2in}
\centering
\includegraphics[width=\textwidth]{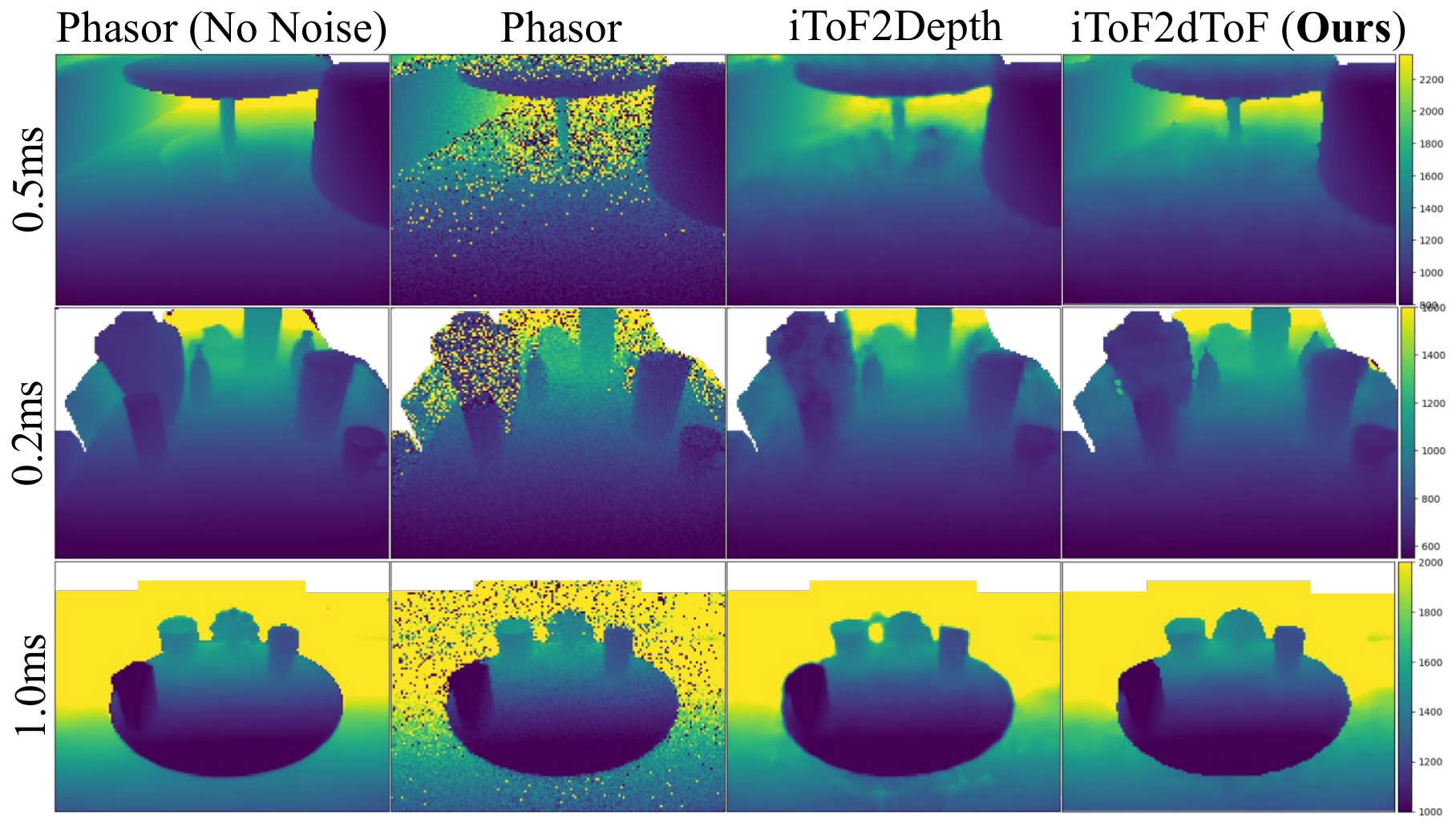}
\caption{\textbf{Real-world Depth Images.} 
Recovered depths for multiple scenes at various exposure times. 
The Phasor (No Noise) images provide an approximate view of how the correct depth image should look like. 
For visualization purposes, we mask pixels (white regions) that exhibit phase wrapping or that were still noisy in the ``noiseless'' Phasor image. 
We find that iToF2dToF generates higher-quality depth maps, particularly, in low SNR regions. 
}
\label{fig:real_qualitative_depths}
\vspace{-0.1in}
\end{figure}

\begin{figure}[ht]
\vspace{-0.2in}
\centering
    \includegraphics[width=\textwidth]{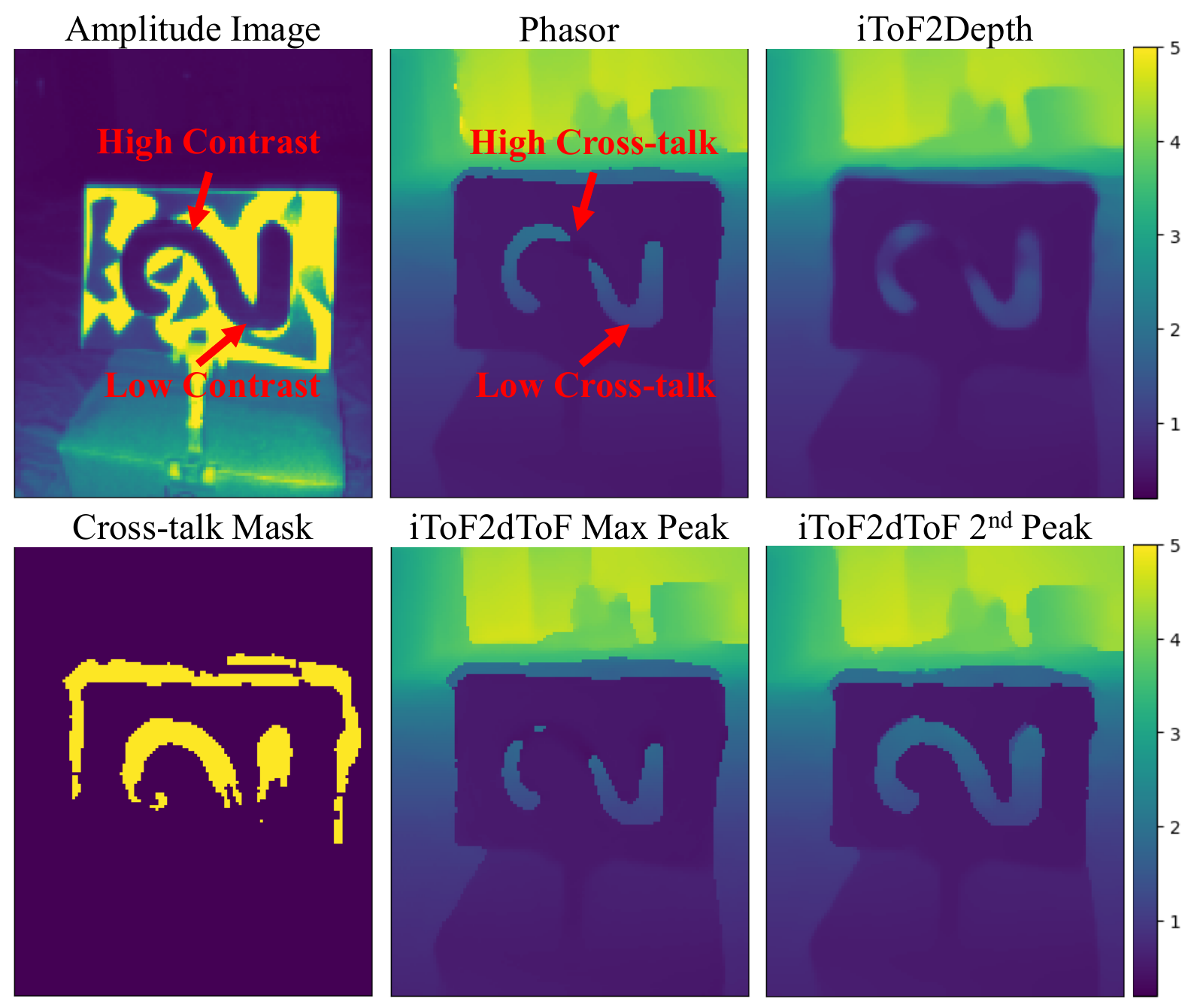}
\caption{\textbf{Optical Cross-Talk Correction.} 
Cross-talk arises in ToF images with \textit{high contrast} regions. 
In these regions all methods incorrectly assign the foreground depth to background pixels. 
Fortunately, the dToF representation and the amplitude image allow us to identify high contrast regions and correct those pixels by selecting the depth of the second peak (iToF2dToF $2^{\text{nd}}$ Peak). 
}
\label{fig:xtalk}
\vspace{-0.1in}
\end{figure}

\subsection{Flexibility of iToF2dToF}
\label{sec:6_3_flexibility}

We integrate iToF2dToF with two rule-based algorithms to resolve specular MPI and optical cross-talk. 
In Supplementary Section 5 we show two more examples that use the flexible dToF representation for specular MPI and depth refinement. 

\smallskip

\noindent \textbf{Specular MPI:} 
The maximum peak of specular pixels may not encode the true depth because the direct reflection ($1^{\text{st}}$ peak) is not the maximum, e.g., TV pixel in Figure \ref{fig:itof2dtof}. 
In Figure \ref{fig:itof2dtof}, we adapt iToF2dToF to use a $1^{\text{st}}$ peak finding algorithm. 
First, we find candidate peaks with Scipy's  \cite{2020scipy} built-in \textit{find\_peaks} function. 
The largest 2 peaks, whose heights are at least 2x the median of the transient pixel, are selected.
Finally, the $1^{\text{st}}$ peak is selected and a 3x3 median filter is applied to remove outliers caused by noisy peaks.
%

\smallskip

\noindent \textbf{Optical Cross-Talk:} 
When a scene contains high contrast regions, such as a bright foreground and dark background, the signal from the bright foreground pixels leaks into the dark regions. 
This ToF artifact is a type of MPI attributed to lens inter-reflections \cite{mure2007real} and often called optical cross-talk. Fundamentally, cross-talk is a kind of sparse MPI where the correct depth corresponds to the second peak in the transient response. 
Despite not explicitly modeling optical cross-talk in the training set, in Figure \ref{fig:xtalk} we show how iToF2dToF can be integrated with the following peak finding algorithm to resolve cross-talk. 

We begin by finding high contrast regions in the amplitude image.
To this end, we take the absolute difference between the amplitude image and a blurred amplitude image, and divide by the amplitude image.
We threshold this ratio to obtain a mask of high contrast regions. 
This high contrast mask is then multiplied with a second mask indicating all pixels with a $2^{\text{nd}}$ peak. 
To find the $2^{\text{nd}}$ peak we use the same algorithm as for specular MPI. 
Finally, we use the mask to select the pixels that should use the depth of the second peak. 
This algorithm leads to a few outliers in the $2^{\text{nd}}$ peak depth image, so a median filter is applied.



\section{Discussion and Limitations}
\label{sec:7_discussion}
\noindent \textbf{Challenges of Depthmap Supervision:} 
Data-driven iToF models that perform supervision on the final depthmap representation can achieve good performance as long as the training dataset provides \textit{explicit supervision} for each scenario. 
Explicit supervision, however, is hard to provide in ambiguous depth sensing scenarios (e.g., specular MPI, cross-talk, and depth discontinuities) where depths are encoded differently in the data (i.e., max peak vs. 1st peak vs. second peak). Specular MPI correction requires a well-balanced dataset with diffuse and specular materials. Optical cross-talk would require an accurate optics simulator. Finally, correctly assigning depths values at depth discontinuities can be application dependent (see Supplementary Section 5.C), which would require end-to-end models to be re-trained for each application.
In practice, a large, diverse, and accurate dataset that provides this level of supervision is challenging to collect or even simulate.
To alleviate this depthmap supervision dependency, we propose to learn an intermediate representation that encodes depths, but decouples the depth estimation step from training. 
At test time, depths are estimated from the intermediate representation using rule-based algorithms that embed domain knowledge, enabling high-performance in multiple challenging scenarios. 

More broadly, end-to-end learning-based models that perform supervision on the final target representation (e.g., depthmaps in our case) have achieved state-of-the-art in many applications. 
Our work suggests that it is sometimes beneficial to perform supervision on a sensible intermediate representation, from which the final representation can be extracted using rule-based algorithms. 
This approach was particularly useful in corner cases and ambiguous scenarios where it was easier to perform supervision on the intermediate representation. 
Finally, in the more common scenarios, the proposed method also achieved good performance, in particular, at low SNR.


\smallskip

\noindent 
\textbf{Transient vs. Depth Supervision: }One limitation of most data-driven iToF models is their reliance on ground truth transient images. This is particularly true for iToF2dToF where the supervision is done on frequency-domain transient images. Although, synthetic transient image datasets have become the standard for training data-driven iToF imaging models \cite{su2018deep,guo2018tackling,qiu2019deep,marco2017deeptof, agresti2019unsupervised}, this is not always necessary \cite{son2016learning, chen2020very}.
In Supplementary Section 6.D we analyzed an iToF2dToF model that was supervised on the final depth images estimated from the intermediate transient representation. Although, we do observe some benefits of this training approach, this model is not able to match the performance of iToF2dToF in the more challenging cases.

\smallskip

\noindent \textbf{Scattering Media: }In Supplementary Section 6.E we evaluate iToF2dToF on scattering media. Although, iToF2dToF does not completely break like other learning-based models, our transient analysis suggests poor generalization to this scenario. 

\smallskip

\noindent \textbf{Other Architectures:} To isolate the benefits of iToF2dToF's ``Input2Output'' representation comparisons were done with networks of similar size, same architecture, and same training set. Specifically, we chose a fixed U-net architecture, and trained it on different representations. 
An extended analysis on how other architectures (e.g., KPN \cite{guo2018tackling, qiu2019deep}) work with all representations is an interesting avenue for future work.

\smallskip

\noindent \textbf{Bridging the iToF and dToF Gap:} dToF sensors overcome the challenges iToF faces by directly capturing transient images. 
However, their low-resolution, high-cost, moving parts, and high power consumption, prevent their use in certain applications. 
To some degree, we are closing the gap between iToF and dToF by exploring the extreme transient imaging case where only \emph{two} frequencies are available. 
We can indeed reconstruct the transient from the limited input, albeit, at a lower time resolution than current dToF sensors.
An interesting direction for future work could explore the limits of data-driven iToF-based transient imaging with additional and higher frequencies \cite{bamji2018impixel} and its applications \cite{su2016material}.


\smallskip
\noindent \textbf{Acknowledgements:} The authors would like to thank Jiaojiao Tian, Yu Yao, and Yin Zhang for their help with data collection.





\ifCLASSOPTIONcaptionsoff
  \newpage
\fi

\clearpage

\onecolumn

\renewcommand{\figurename}{Supplementary Figure}
\renewcommand{\thesection}{S. \arabic{section}}
\setcounter{figure}{0}
\setcounter{section}{0}
\setcounter{page}{1}
\begin{center}
  {\Large \bf Supplementary Document for ``iToF2dToF: A Robust and Flexible Representation for Data-Driven Time-of-Flight Imaging"}
  \smallskip
  
    {Felipe Gutierrez-Barragan, Huaijin Chen, Mohit Gupta, Andreas Velten, Jinwei Gu}
    
    {Project page: \url{http://pages.cs.wisc.edu/\~felipe/project-pages/2021-itof2dtof/}}
  
  

\end{center}

\section{Simulator and Signal Level Validation}
\label{sec:supplement-1_simulator}

To simulate realistic signal and noise levels, we tune the exposure and power parameters of our simulator by comparing the simulated images with similar real captured images. The real and synthetic ToF images have the following parameters:

\begin{itemize}
\item \textbf{Real Data Parameters:} We captured dual-frequency ToF data of a white flat plain 500mm, 1000mm, and 2000mm away from the camera. The images are captured using a 0.3ms exposure and a light source average power of 1W. For each frequency, dual-tap 4-phase measurements are made, resulting in 8 measurements per frequency.
\item \textbf{Simulated Data Parameters:} We simulated dual-frequency ToF data of a white flat wall 500mm, 1000mm, and 2000mm away from the camera. The exposure time is also set to 0.3ms, and 4-phase measurements are simulated per frequency. To match real data acquisition 2 frames are averaged for each phase. To approximately match the signal levels (i.e., amplitude) of the real ToF data we set the average source power in the simulation to 32W. 
\end{itemize}

\noindent To obtain ($b_{\omega, 0}$, $b_{\omega, \frac{\pi}{2}}$) from 4-phase data we simply take the difference between the 0-180 and 90-270 measurements.

\medskip

\begin{figure}[h]
\centering
    \includegraphics[width=0.85\textwidth]{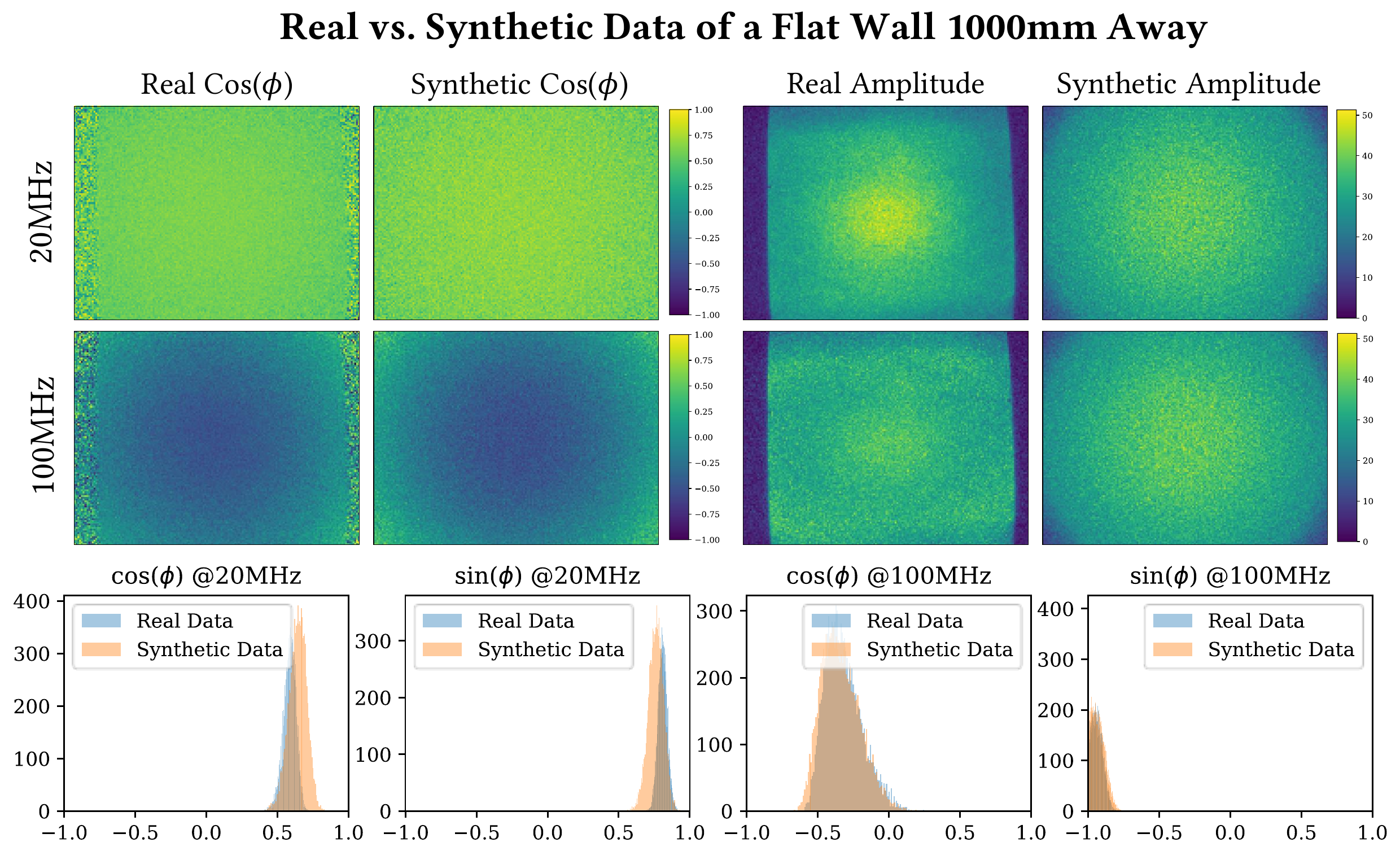}
\caption{\textbf{Synthetic vs. Real ToF Images} of a flat wall 1000mm away from the camera. The first row shows the real and synthetic phase images (columns 1 and 2), and the real and synthetic amplitude images (columns 3 and 4), for 20MHz. The second row shows the same images for 100MHz. 
The flat plain that was captured did not cover the full field of view of the camera, as seen in the real phase and amplitude images. 
Therefore, when calculating the distribution of the recovered phases (third row), we cropped the real and synthetic images such that we only included the valid pixels.    }
\vspace{-0.1in}
\label{fig:supplement_synthetic_vs_real_1000mm}
\end{figure}

\noindent \textbf{Amplitude Images:} Figures \ref{fig:supplement_synthetic_vs_real_1000mm} and \ref{fig:supplement_synthetic_vs_real} compare the synthetic and real data with the above parameters. Due to the limited bandwidth of the real ToF camera, the 100MHz amplitude image is lower than the 20MHz. 
To compensate for bandwidth in our simulation, we tuned the simulation power parameter such that the amplitude images at 20MHz and 100MHz fall in between the real amplitude images. 
This leads to similar noise levels, as shown in the histograms of the recovered phases.

\medskip

\noindent \textbf{Phase Images:} Qualitatively, the recovered phase images match for the captured and simulated scenes, as shown in Figures \ref{fig:supplement_synthetic_vs_real_1000mm} and \ref{fig:supplement_synthetic_vs_real}. Quantitatively, the histogram of phases for the captured and simulated data matched for 100MHz. For 20MHz, synthetic data displays higher variance in the histogram, which is expected since the simulated signal levels are lower. 
Furthermore, due to imperfect calibration at 20MHz, we observe a small phase offset between the real an synthetic data. 
Unfortunately, due to the lack of low-level control of the ToF module we are not able to correct for this.
Nonetheless, our results show that this small phase offset does not have a significant effect on the performance of the trained models.


\clearpage

\begin{figure}[h]
\centering
\centerline{
	\includegraphics[width=\linewidth]{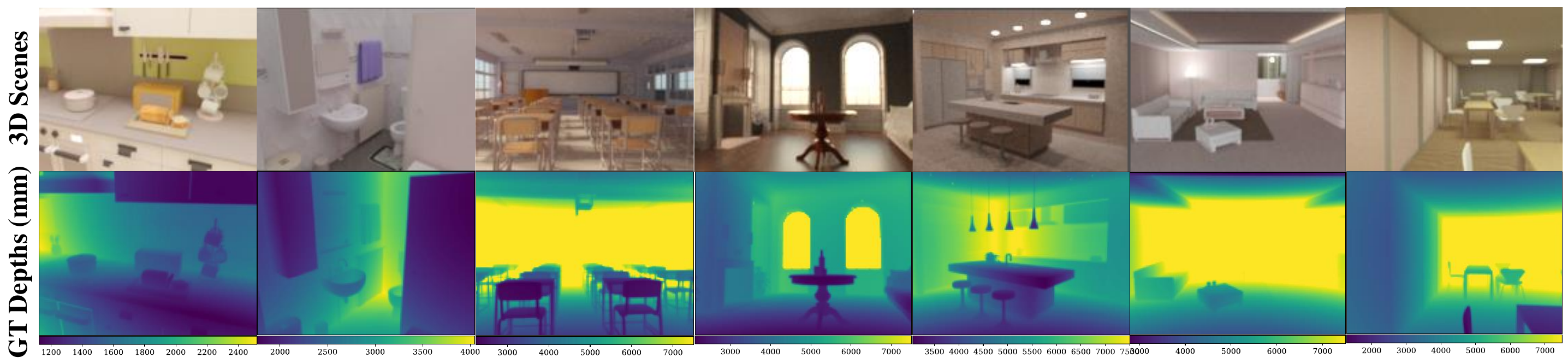}
}
\caption{\textbf{Dataset Overview.} Example synthetic 3D scenes from the dataset.}
\label{fig:supplement_dataset_overview}
\vspace{-0.1in}
\end{figure}

\vspace{-0.25in}

\section{Implementation Details}
\label{sec:supplement-3_implementation}

\subsection{Transient Data Generation and iToF Simulator}

In this section we describe the synthetic data generation pipeline we used to create the MF-ToF dataset used for training and the ablation studies. Our pipeline shares some similarities to the data generation pipelines developed for previous data-driven iToF works \cite{guo2018tackling, marco2017deeptof, su2018deep, agresti2019unsupervised, qiu2019deep}, and some differences which we will point out. 
\medskip

\noindent \textbf{Transient Data: }For transient rendering we use MitsubaToF \cite{pediredla2019ellipsoidal}, an extension to the widely used Mitsuba renderer \cite{jakob2010mitsuba}. Previous works have used a modified pbrt-v3 \cite{pharr2016physically} for time-resolved rendering \cite{su2018deep, qiu2019deep}, proprietary software \cite{agresti2019unsupervised}, or recent transient rendering software \cite{marco2017deeptof, jarabo2014framework, guo2018tackling}. We chose Mitsuba because it is faster than pbrt-v3 \cite{glanz2017pbrcomparison}, and it provided a plugin to convert Blender 3D scene models to Mitsuba \cite{mitsubaplugins}. 
Using Mitsuba's notation, we use a pulsed \textit{spot} light emitter, with a 30 degree \textit{beamwidth}, that flash-illuminates the scene for transient rendering. 
These parameters match the simulated and real sensor's FoV. 
All transient images in this paper have a 50ps time resolution and 2000 time bins resulting in a time domain of 0-1000ns. This time domain contains all light paths that are 0-20 meters in length, and provides a depth resolution of 5mm. 

\medskip

\noindent \textbf{3D Scene Models:} We gathered 14 Mitsuba scene models from this repository \cite{bitterli2016rendering}. Additionally, we downloaded 11 Blender scenes from \cite{blendswap2016} and converted them to Mitsuba using the exporter plugin \cite{mitsubaplugins} along with our post-processing scripts to correct for some errors made by the plugin. For all scenes we change completely transparent and completely specular (mirror-like) materials to diffuse materials with a constant randomly chosen albedo. This avoids over-fitting to these corner cases, for which even an ideal dToF system cannot recover depths because fully transparent and specular materials will not have a direct illumination signal. Our 3D scene models, as shown in Figure \ref{fig:supplement_dataset_overview}, contain realistic texture/albedos and also diffuse and glossy materials (smooth plastic in Mitsuba terminology). For simulating textures/albedos, previous works have relied on randomizing constant albedos over the full scene model \cite{su2018deep,marco2017deeptof} or multiplying the resulting rendered images by a texture image \cite{guo2018tackling}, which leads to less realistic transient images.

\medskip

\noindent \textbf{Camera View Generation: }For a given 3D scene model we semi-automatically generate 200 camera views. To guarantee a diverse set of camera views, we manually pick 4-10 view points for each scene. We use the manually picked view points as the starting point for our automatic camera trajectory generation. When automatically generating a new camera view we take a small step forward in the current view direction, and then apply a small random rotation to the current view direction. We check for collisions by generating the depth map of the current view and checking if any points are closer than 200mm. If a collision is detected the camera view direction is flipped. If we detect the camera moving towards empty space we slowly rotate the camera until it is looking toward scene content again. Our camera view generation program relied heavily on Mitsuba's Python API. To create large synthetic ToF datasets, previous works, have also generated camera views automatically \cite{su2018deep} and manually \cite{agresti2019unsupervised,marco2017deeptof}.

\medskip

\noindent \textbf{iToF Simulation: }The input to our iToF simulator is a transient image. The transient image is convolved with a sinusoidal light source modulation function, and the resulting function is the signal arriving at the sensor. Similar to previous works \cite{su2018deep, marco2017deeptof, qiu2019deep}, we find that we obtain good generalization assuming sinusoidal signals because the real iToF module functions are calibrated. In other words, the iToF module's output phase measurements are roughly the same as an ideal sinusoid phase measurements. The arriving signal is multiplied with the sensor demodulation function, and integrated. Similar to previous works, we do not consider ambient illumination \footnote{In practice, iToF modules operate indoors where ambient illumination is low, and are also equipped with spectral filters which largely mitigate the ambient illumination incident on the sensor.} \cite{su2018deep, guo2018tackling}. Next, the integrated signal is scaled according to different sensor parameters (quantum efficiency, exposure time, electron to voltage factor) obtained from the iToF module used in experiments. Finally, read and photon noise are added to the iToF measurement. To achieve realistic signal and noise levels we calibrate the average light source power parameter used in simulation with real iToF data (details in Section \ref{sec:supplement-1_simulator}).

\subsection{Synthetic Multi-Frequency ToF (MF-ToF) Dataset}
\label{sec:supplement-2_mftof_dataset}

Using the simulator and the 5000 transient images, we generate the synthetic MF-ToF dataset. 
Different from previous works, we simulate multiple frequencies from 20MHz to 600MHz in steps of 20. 
A 20MHz repetition frequency leads to an unambiguous depth range of 7.5m, which is sufficient for indoor applications. 
To cover a wide range of SNR levels we simulate multiple frames and average them. 
According to our simulator validation in Section \ref{sec:supplement-1_simulator}, averaging 2 frames leads to similar noise levels as using an exposure time of 0.3ms in the real sensor. 
Therefore, for each image the number of averaged frames is randomly chosen between 1 and 12 to emulate exposure times ranging between 0.15-1.8ms, which are reasonable in commercial iToF cameras \cite{chen2020very}. We use this dataset for training.
Furthermore, we simulate 4 additional datasets where the range of averaged frames is 1-1, 1-2, 3-6, 6-12 (instead of 1-12), and we use these datasets for evaluating the performance of each model at different SNR levels as shown in Section 6.2 in the main document. 

Limited by the transient rendering time ($\sim$7 days on a 32 core machine), the dataset has a 120x160 spatial resolution.
Although, the dataset has a lower spatial resolution than current commercial iToF sensors, a higher resolution dataset may not be essential to learn to correct MPI. 
This is because MPI is often similar for large regions in the image, therefore, it is an artifact that can be resolved at a lower resolution. 
Nonetheless, the denoising component of the model will benefit from a high-resolution synthetic dataset.
Therefore, future iterations of the MF-ToF dataset will be at the full resolution of commercial iToF sensors.

\clearpage

\begin{table}[h]

\resizebox{\linewidth}{!}
{
\begin{tabular}{@{}ccccccccc@{}}
\toprule
\multicolumn{9}{c}{\textbf{Network Architectures}} \\ \midrule
\textbf{Name} & \textbf{InConv} & \textbf{D1} & \textbf{D2} & \textbf{U1} & \textbf{Conv1} & \textbf{U2} & \textbf{Conv2} & \textbf{OutConv} \\ \midrule
\textbf{Layer} & \begin{tabular}[c]{@{}c@{}}Conv2D \\ + ReLU\\ + Conv2D \\ + ReLU\end{tabular} & \begin{tabular}[c]{@{}c@{}}Conv2D \\ + ReLU\\ + Conv2D \\ + ReLU\end{tabular} & \begin{tabular}[c]{@{}c@{}}Conv2D \\ + ReLU\\ + Conv2D \\ + ReLU\end{tabular} & ConvTranspose2D & \begin{tabular}[c]{@{}c@{}}Conv2D \\ + ReLU\\ + Conv2D \\ + ReLU\end{tabular} & ConvTranspose2D & \begin{tabular}[c]{@{}c@{}}Conv2D \\ + ReLU\\ + Conv2D \\ + ReLU\end{tabular} & Conv2D \\ \midrule
\textbf{Kernel} & 7x7 / 3x3 & 3x3 / 3x3 & 3x3 / 3x3 & 2x2 & 3x3 / 3x3 & 2x2 & 3x3 / 3x3 & 1x1 \\ \midrule
\textbf{Stride} & 1 / 1 & 2 / 1 & 2 / 1 & 2 & 1 / 1 & 2 & 1 / 1 & 1 / 1 \\ \midrule
\textbf{\begin{tabular}[c]{@{}c@{}}Skip\\ Connection\end{tabular}} & --- & --- & --- & --- & From D1 & --- & From InConv & --- \\ \midrule
\textbf{Input} & Raw ToF & InConv & D1 & D2 & [U1, D1] & Conv1 & [U2, InConv] & Conv2 \\ \midrule
\textbf{Channels I/O} & 4 / 64 & 64 / 128 & 128 / 256 & 256 / 128 & 256 / 128 & 128 / 64 & 128 / 64 & 64 / $N_{\text{out}}$ \\ \midrule
\textbf{\begin{tabular}[c]{@{}c@{}}iToF2Depth\\ Output - $N_{\text{out}} = 1$\end{tabular}} & 64 x $H$ x $W$ & 128 x $\frac{H}{2}$ x $\frac{W}{2}$ & 256x $\frac{H}{4}$ x $\frac{W}{4}$ & 128 x $\frac{H}{2}$ x $\frac{W}{2}$ & 128 x $\frac{H}{2}$ x $\frac{W}{2}$ & 64 x $H$ x $W$ & 64 x $H$ x $W$ & 1 x $H$ x $W$ \\ \midrule
\textbf{\begin{tabular}[c]{@{}c@{}}iToF2iToF\\ Output - $N_{\text{out}} = 4$\end{tabular}} & 64 x $H$ x $W$ & 128 x $\frac{H}{2}$ x $\frac{W}{2}$ & 256x $\frac{H}{4}$ x $\frac{W}{4}$ & 128 x $\frac{H}{2}$ x $\frac{W}{2}$ & 128 x $\frac{H}{2}$ x $\frac{W}{2}$ & 64 x $H$ x $W$ & 64 x $H$ x $W$ & 4 x $H$ x $W$ \\ \midrule
\textbf{\begin{tabular}[c]{@{}c@{}}iToF2dToF\\ Output - $N_{\text{out}} = 2S$\end{tabular}} & 64 x $H$ x $W$ & 128 x $\frac{H}{2}$ x $\frac{W}{2}$ & 256x $\frac{H}{4}$ x $\frac{W}{4}$ & 128 x $\frac{H}{2}$ x $\frac{W}{2}$ & 128 x $\frac{H}{2}$ x $\frac{W}{2}$ & 64 x $H$ x $W$ & 64 x $H$ x $W$ & $2S$ x $H$ x $W$ \\ \bottomrule
\end{tabular}
\caption{Detailed U-net network architectures. All models used the same backbone U-net. The only difference across models was the number of output channels in the \textit{OutConv} layer. For iToF2Depth, the number of output channels was 1 ($N_{\text{out}} = 1$) because we directly output the depth image. For iToF2iToF, $N_{\text{out}} = 4$ because we output the denoised input ToF images. For iToF2dToF, the $N_{\text{out}}$ depends on the maximum frequency we extrapolate, specified by the parameter $S$.
In our main results we used a repetition frequency of 20MHz, and the maximum frequency we extrapolated to was 400MHz, making $S = 20$. }
\label{tab:supplementary_network_arch}
}
\vspace{-0.1in}
\end{table}
\vspace{-0.1in}

\subsection{Network Architectures and Training}

\noindent \textbf{Input and Output: }The inputs to the network are the raw dual-frequency iToF images at 20 and 100MHz. For each frequency, a raw measurement corresponds to 2 input channels ($b_{\omega, 0}$, $b_{\omega, \frac{\pi}{2}}$). In the iToF2dToF models the network outputs the interpolated and extrapolated frequencies up to a given maximum frequency. 
We normalize each frequency by dividing by its amplitude \cite{su2018deep}, i.e., $\hat{b}_{\omega, 0} = b_{\omega, 0}/\sqrt{ b_{\omega, 0}^2 + b_{\omega, \frac{\pi}{2}}^2}$.
As discussed in Section \ref{sec:supplement-5_synthetic_data_results_with_amplitude}, we find that using amplitude information is useful, however, we also observe that models trained with amplitude information did not generalize well to real data.

\medskip

\noindent \textbf{Architecture: }As outlined in Table \ref{tab:supplementary_network_arch} we use a simple U-net \cite{ronneberger2015u} for all data-driven models in this paper.
The U-net used skip connection (concatenated), ReLU activations, and learned upsampling.
For downsampling and upsampling we used PyTorch's \cite{paszke2019pytorch} built-in \textit{Conv2d} and \textit{ConvTranspose2d} with strides of 2. 
Depending on the number of output channels the total number of parameters in the models ranged between 1.873-1.877M. 

\medskip

\noindent \textbf{Loss: }We minimize the mean absolute error (i.e. L1 loss) between the U-net's output ($g_{\theta}(\mathbf{B})$) and the target output $\mathbf{T}$:

\vspace{-0.1in}

\begin{equation}
    \mathcal{L}_{L1} = \frac{1}{N} \sum_{i} | g_{\theta}(\mathbf{B})_i -  \mathbf{T}_i |
\label{eq:pixel_radiance}
\end{equation}

\vspace{-0.05in}

\noindent For iToF2Depth $\mathbf{T}$ is the ground truth depth image, and for iToF2dToF $\mathbf{T}$ will be the ground truth brightness images for $\omega_{1}, \omega_{2}, \hdots, \omega_{S}$.
We experimented with a more end-to-end training strategy where the loss function was the KL-Divergence of the recovered transient and the target transient.
However, this approach performed comparably to a simple L1 loss on the output frequencies.
Nonetheless, loss functions designed for data-driven transient imaging is a promising avenue for future work.
While we acknowledge that advanced network design and training strategies explored in previous works \cite{su2018deep,qiu2019deep,marco2017deeptof}, such as adversarial loss and pre-training, can further improve the performance, we focus on the benefits of the dToF representation because they are orthogonal to the benefits from such methods.

\medskip

\noindent \textbf{Training: }
For training we split the dataset into train, valid, and test sets. The training set is composed of all views generated from 21 scenes (4200 instances). The validation and testing sets are each composed of views generated from 2 scenes. 
We train all models in this paper using the ADAM optimizer \cite{kingma2014adam} with a constant learning rate of 0.0001 for 300 epochs, followed by 700 epochs with a linearly decaying learning rate. IToF2Depth required additional training for convergence so we had to use 1700 epochs of linearly decaying learning rate. We use a batch size of 32. 
During training, we mask pixels with depths greater than 7.5m (max depth for 20MHz) and with ``infinite'' depths corresponding to pixels looking at empty space (e.g., open windows or doors). 
Furthermore, during training, we apply random flips for data augmentation. 
At each epoch we evaluate the model's loss on the validation set and we keep track of the model with the lowest validation loss. 
\clearpage
\section{Real Wall Depth Reconstruction Statistics}
\label{sec:supplement-4_wall_stats}

In this section we present additional real data results illustrating the robustness of iToF2dToF to noise. To this end, we compared the standard deviation and the histograms of the reconstructed depths of planar wall images at combinations of 8 distances (250mm-2000mm in steps of 250mm), 3 exposure times (0.5ms, 0.3ms, 0.15ms), and 3 reflectivities (50\%, 20\%, 5\%). Figures \ref{fig:supplement_synthetic_vs_real_1000mm} and \ref{fig:supplement_synthetic_vs_real} show the amplitude and phase images of the planar target we imaged at 3 of the 8 distances. 

\smallskip

\noindent \textbf{Data Acquisition Procedure: } To verify the true depth for each planar wall target, we fixate a cross-shaped laser distance sensor next to the iToF module. For alignment, we attached a mirror to the target and made sure that the reflected laser pointer returned to the emitting point. In this way, we made sure the iToF sensor plane was parallel to the wall. 

\smallskip

\begin{figure}
\centering
    \includegraphics[width=0.99\textwidth]{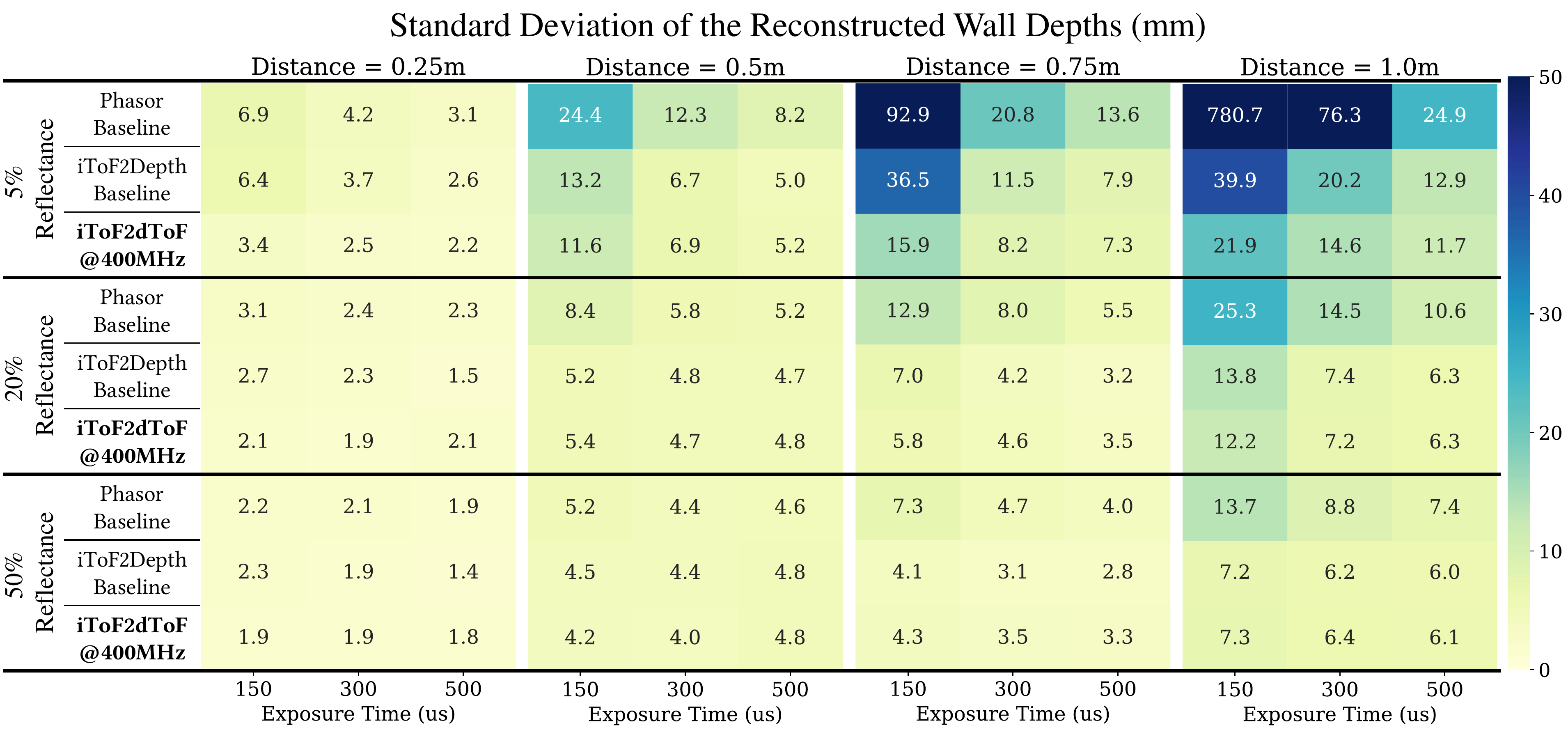}
    \includegraphics[width=0.99\textwidth]{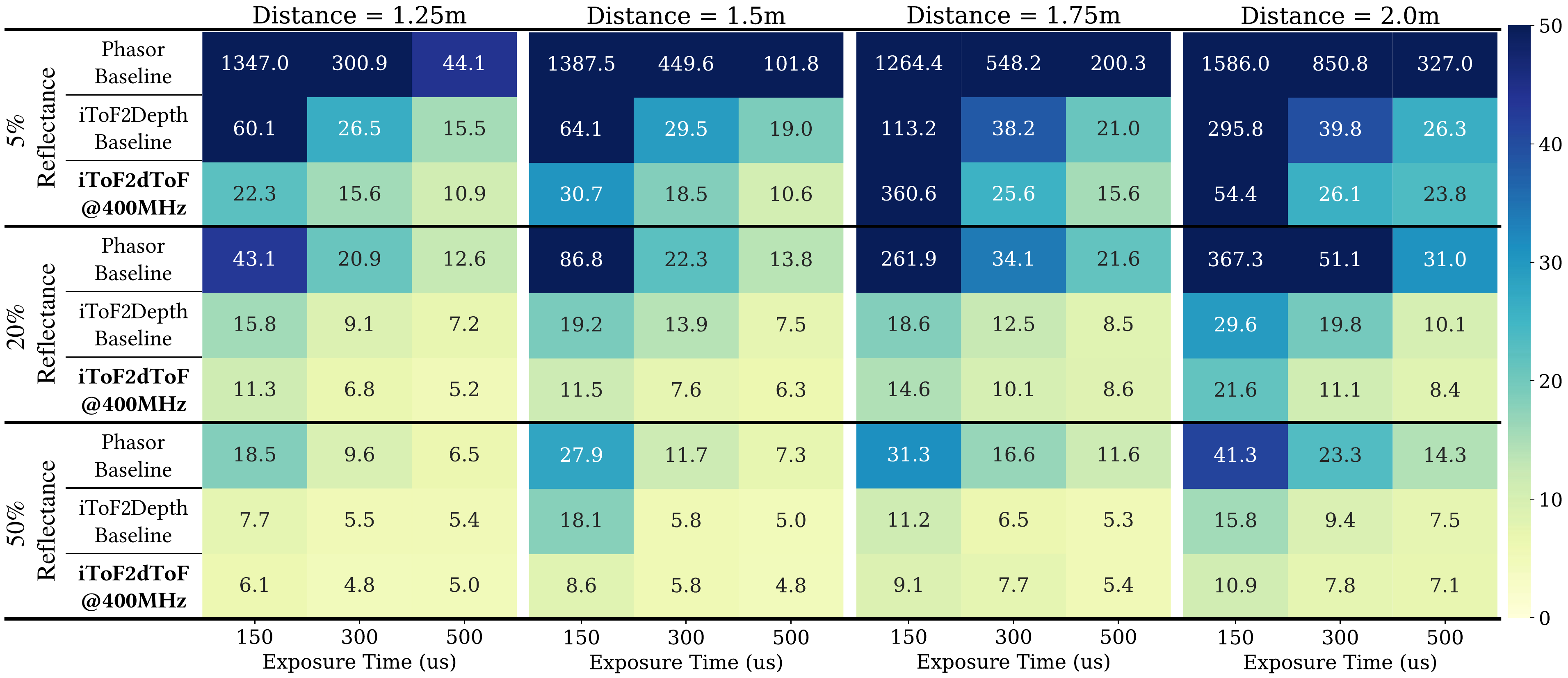}
\vspace{-0.1in}
\caption{\textbf{Wall Depths Standard Deviation}. We collect a iToF data of a wall at different distances, under 9 SNR settings obtained from the combination of different exposure times and reflectance. For most SNR levels, iToF2dToF achieves lower standard deviation in the recovered depths. The benefits of iToF2dToF are most significant at low SNR.}

\label{fig:supplement_wall_stddev}
\end{figure}

\begin{figure}
\centering
    \includegraphics[width=0.45\textwidth]{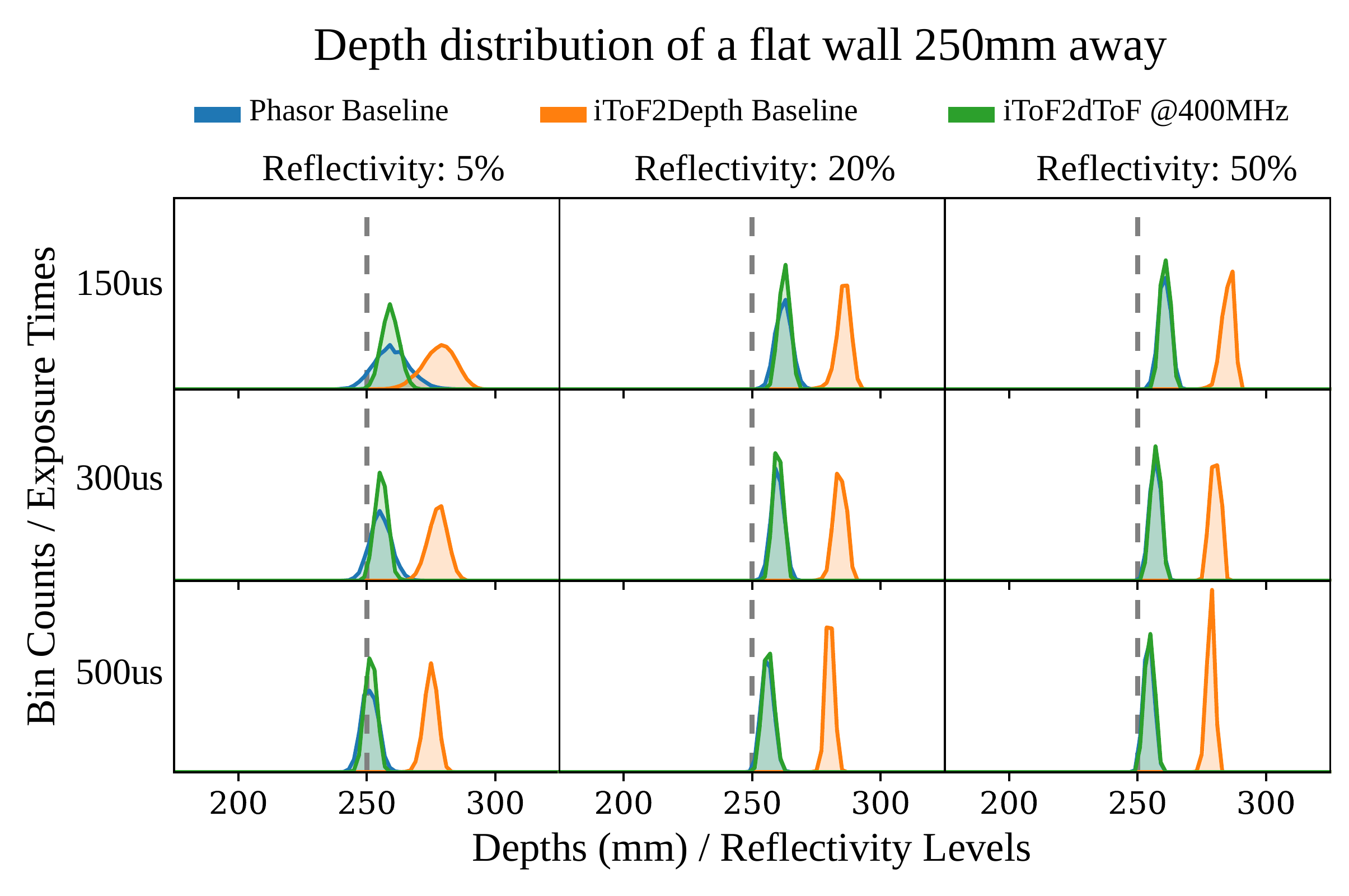}
    \includegraphics[width=0.45\textwidth]{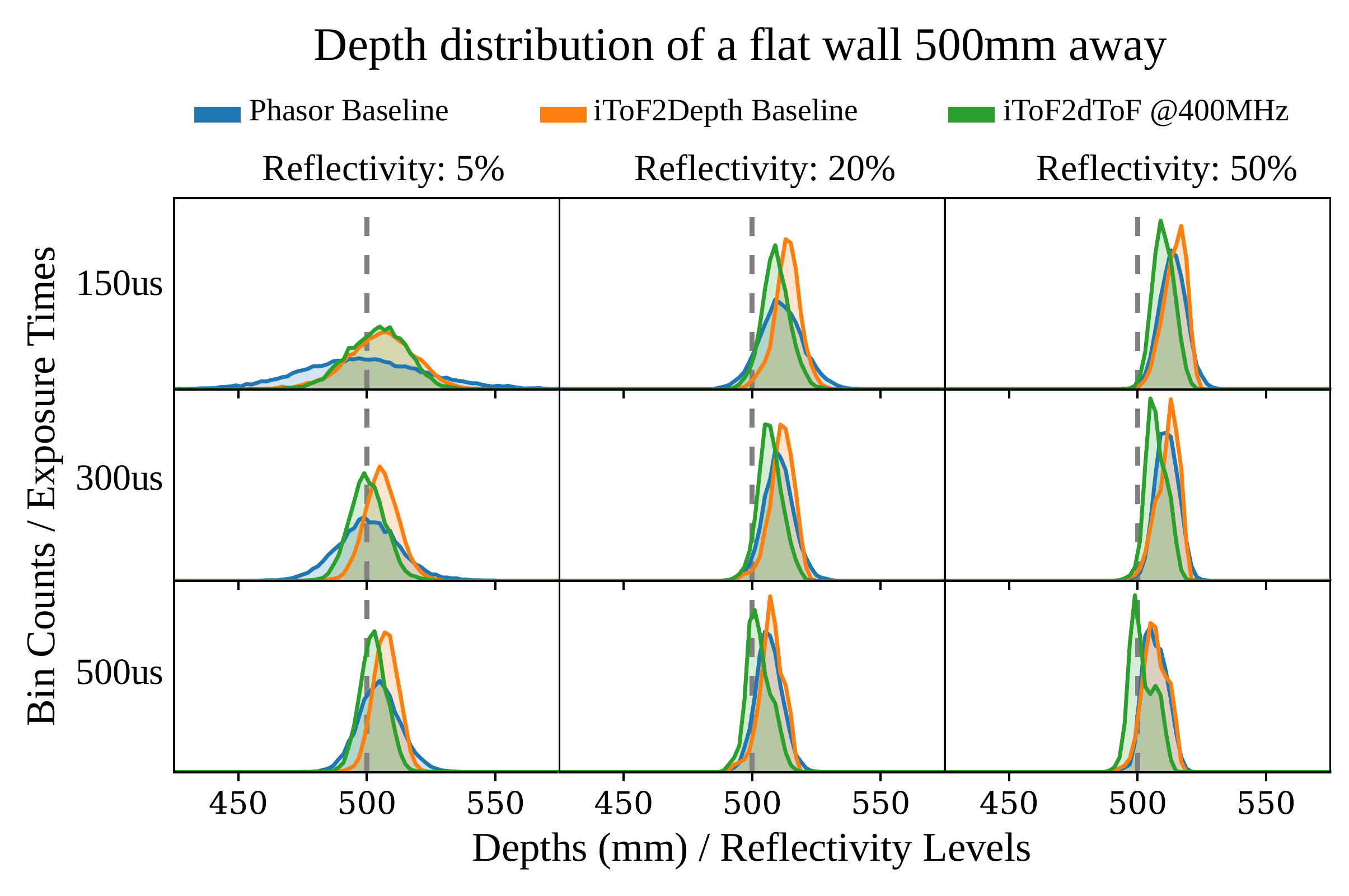}
    \includegraphics[width=0.45\textwidth]{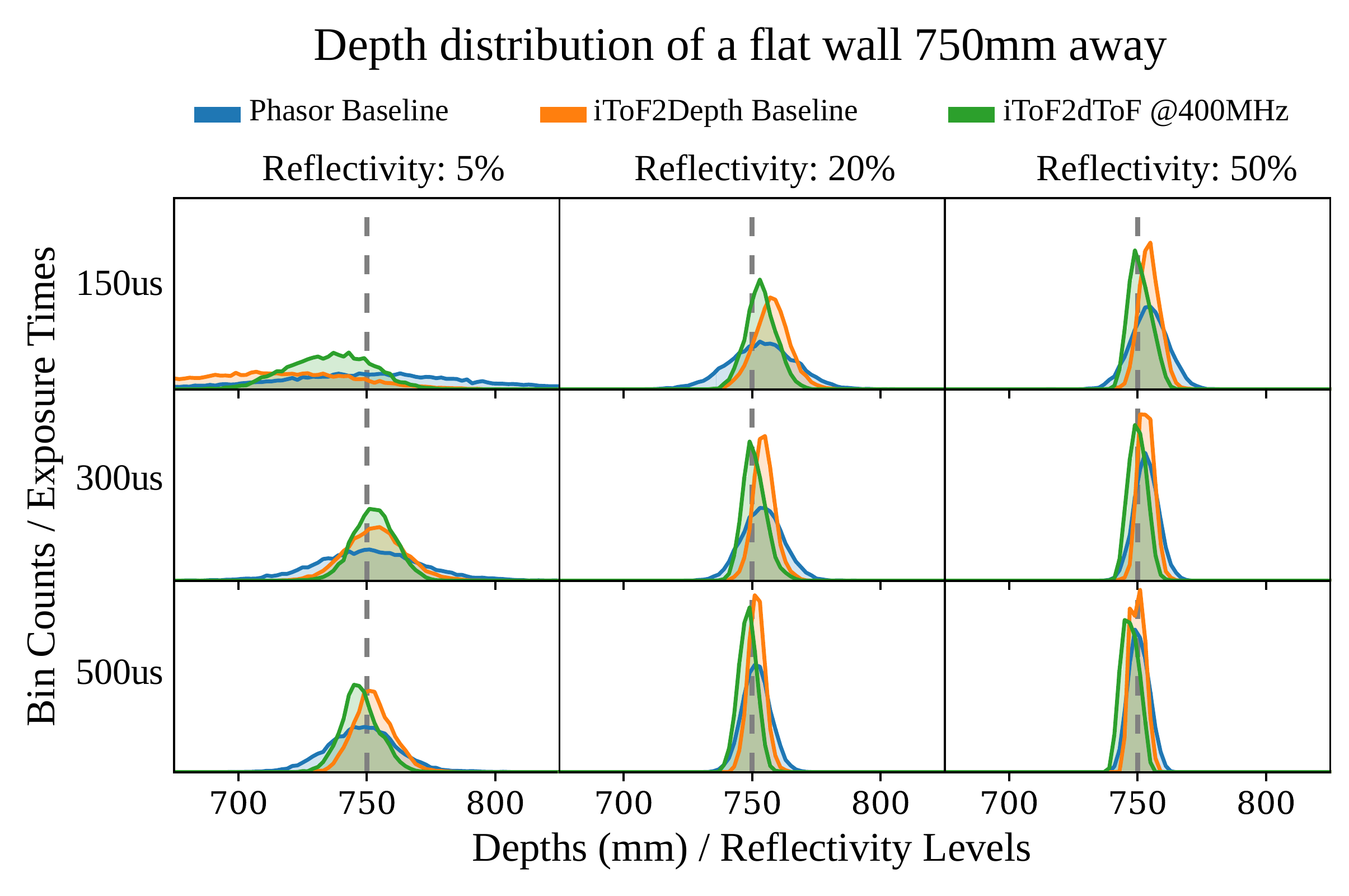}
    \includegraphics[width=0.45\textwidth]{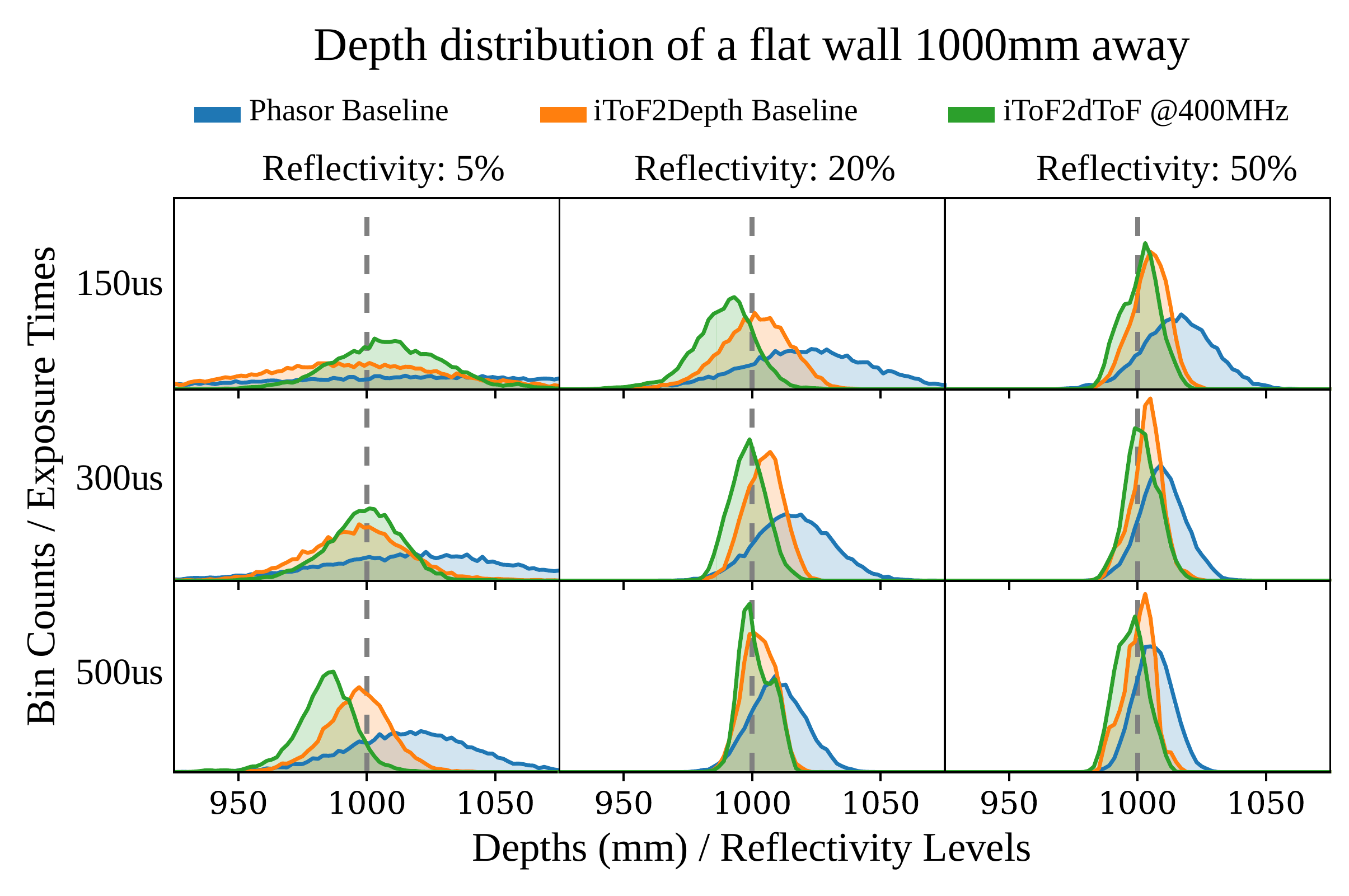}
    \includegraphics[width=0.45\textwidth]{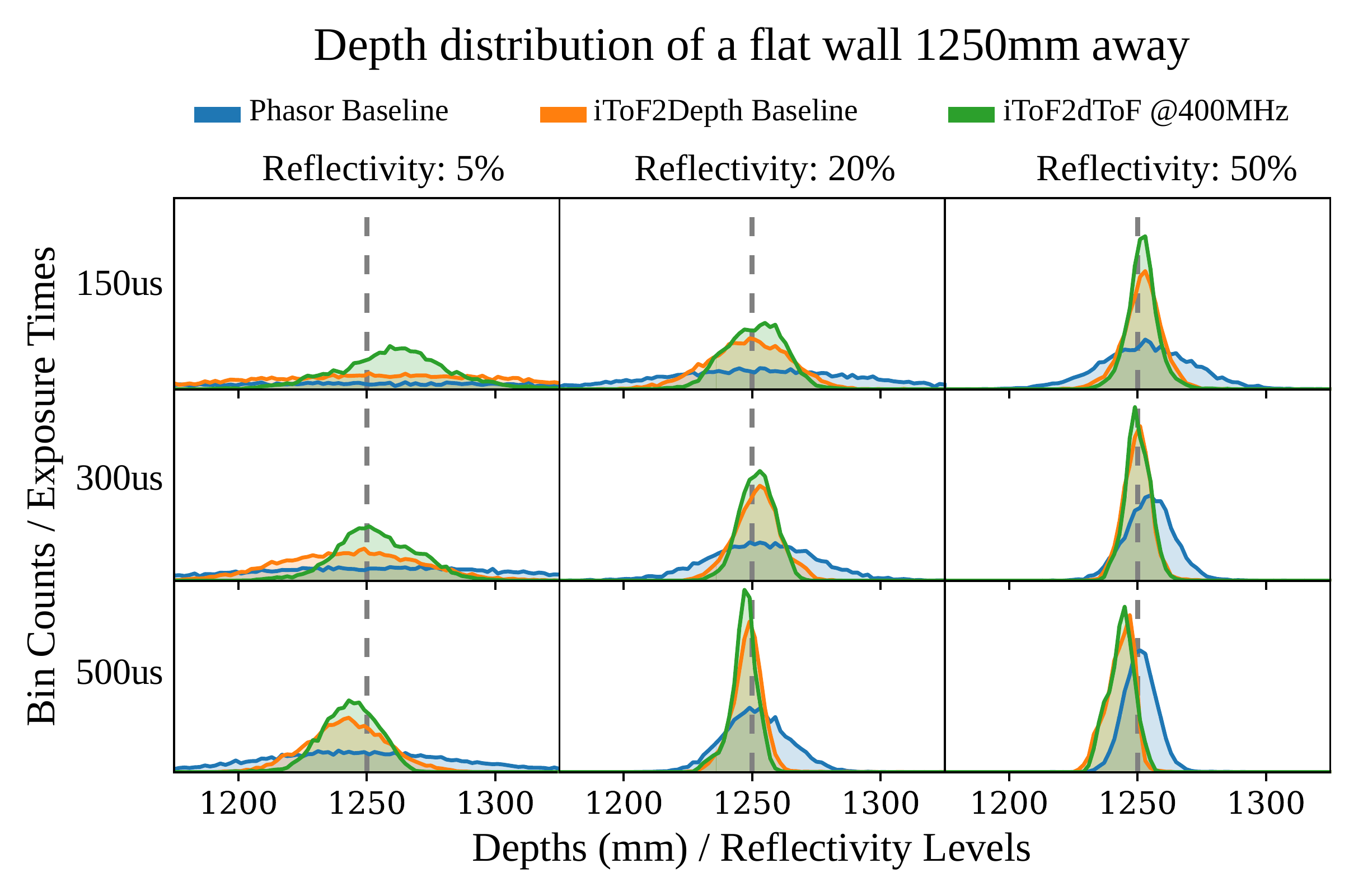}
    \includegraphics[width=0.45\textwidth]{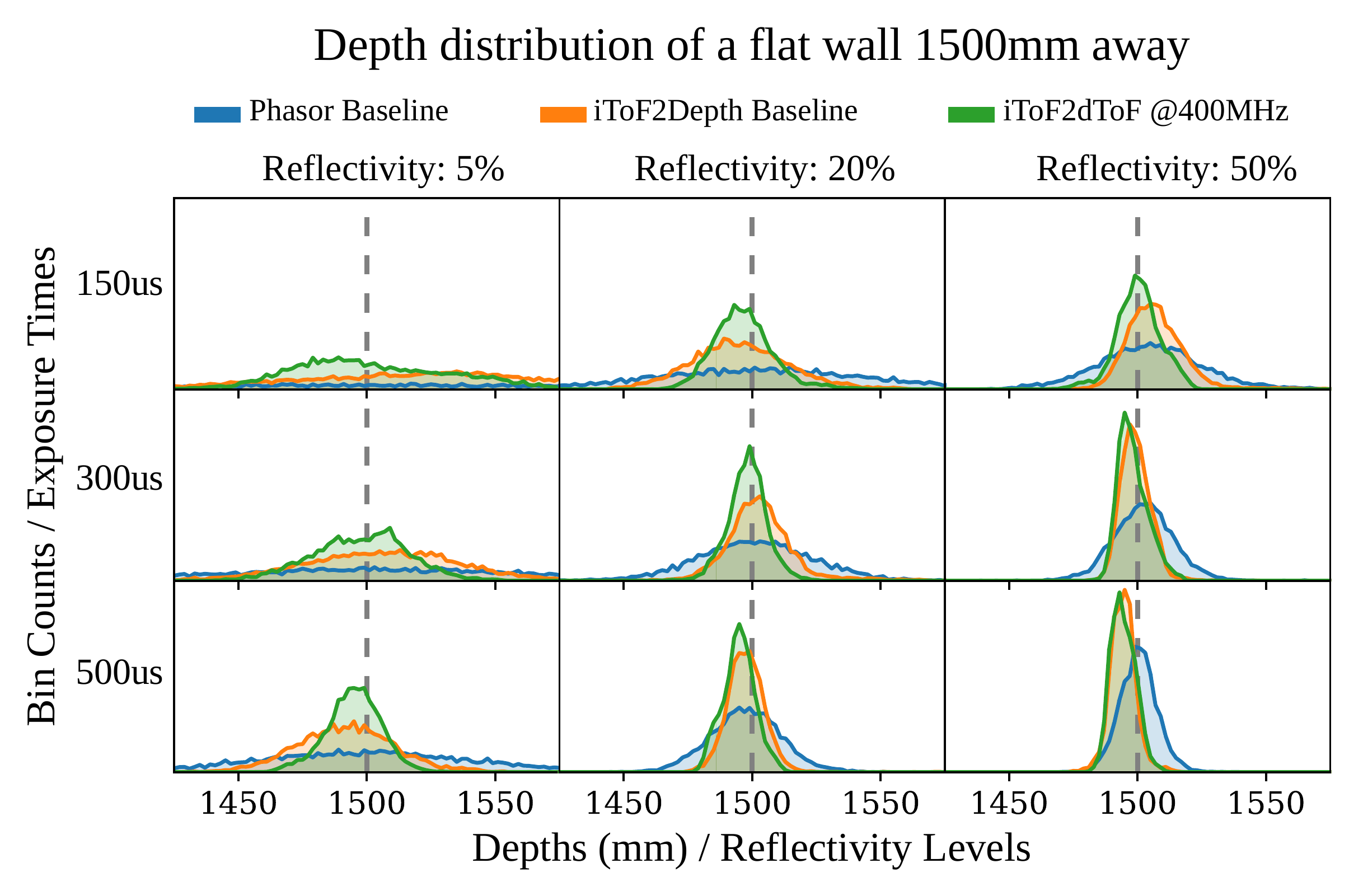}
    \includegraphics[width=0.45\textwidth]{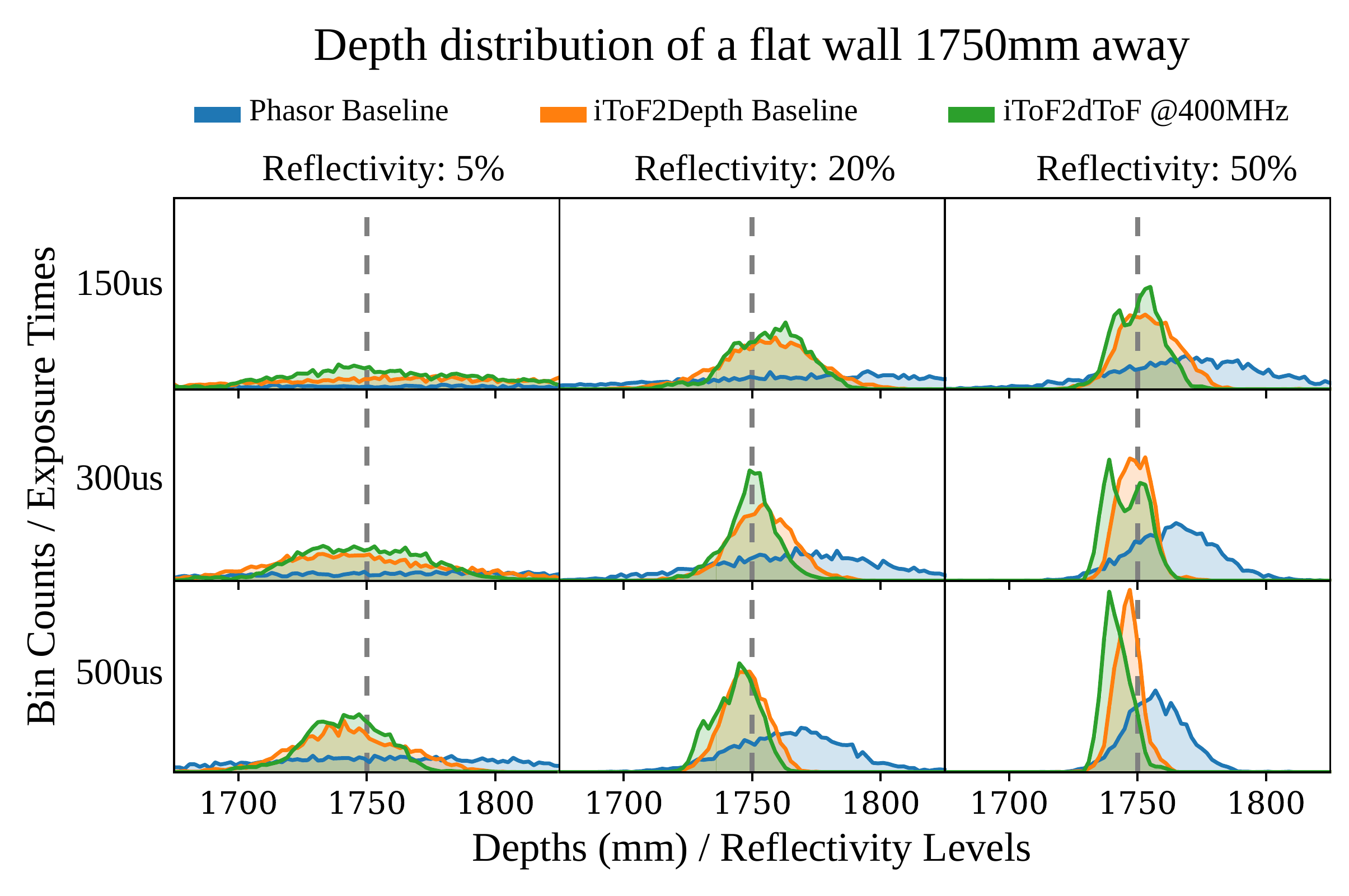}
    \includegraphics[width=0.45\textwidth]{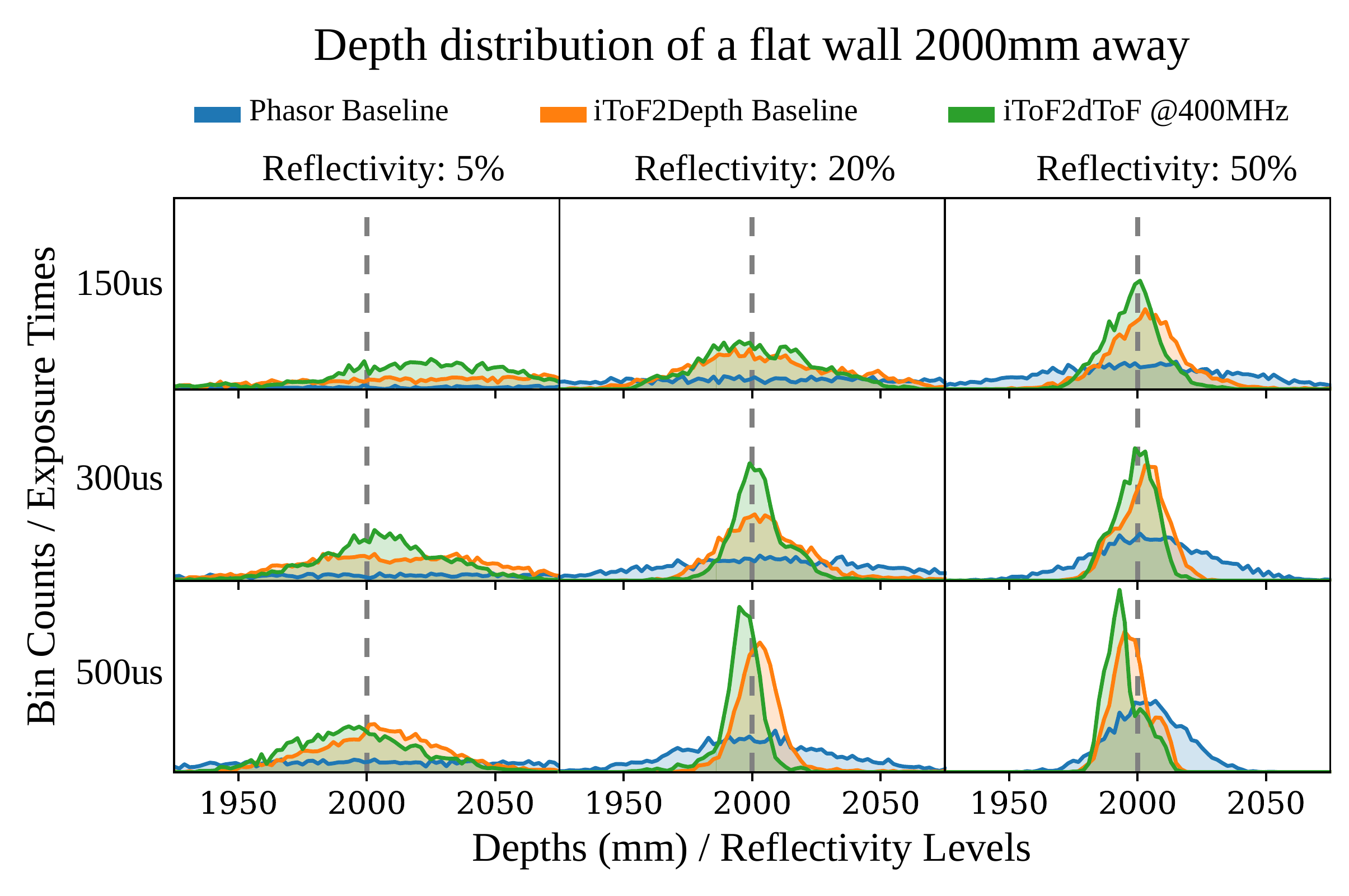}
\caption{\textbf{Wall Depths Histograms}. The variance of the estimated iToF2dToF depths is consistently lower than the baselines.
Due to factors such as: sensor non-linearities, small ground truth errors, and small calibration errors; all methods exhibit a small depth bias that oscillates roughly +/- 2cm around the ground truth depth, depending on the exposure and reflectivity settings. 
}
\vspace{-0.1in}

\label{fig:supplement_wall_histograms}
\end{figure}

\noindent \textbf{Main Observations: } Figure \ref{fig:supplement_wall_stddev} shows a quantitative comparison of the standard deviation of the reconstructed wall depths. For most SNR settings, iToF2dToF recovers depths with lower standard deviation than all methods. The benefits of iToF2dToF are most significant at low SNR settings (i.e., low reflectance and exposure). At the distance of 1.75m, for the lowest SNR setting, iToF2dToF has a higher standard deviation due to outliers in the recovered depths. Figure \ref{fig:supplement_wall_histograms} shows the histograms of the recovered depths for 8 distances. We can observe that the distribution of recovered depths by iToF2dToF has lower variance at low SNR. 
Note that, depending on exposure and reflectivity settings, all methods exhibit a small depth bias that oscillates +/-2cm around the ground truth depth, due to factors such as: sensor non-linearities, small ground truth errors, and small calibration errors.
Finally, it is important to note that the histograms in Figure \ref{fig:supplement_wall_histograms} do not include outliers, while the computation of standard deviations for Figure \ref{fig:supplement_wall_stddev} does include outliers.

\begin{itemize}
    \item \textbf{Artifacts on Depth Histograms: }The depth histograms for farther away depths (1750mm and 2000mm) are noisy. This is due to two reasons. As observed in Figures \ref{fig:supplement_synthetic_vs_real_1000mm} and \ref{fig:supplement_synthetic_vs_real} the planar target we imaged covered a smaller portion of the sensors field of view as it was moved farther away. This means that fewer pixels were used when calculating the standard deviations and depth histograms. Furthermore, more outliers appear at the lowest SNR settings leaving even less pixels available to construct the histograms in Figure \ref{fig:supplement_wall_histograms}.
    For some combination of depth, exposure time, and reflectivity iToF2dToF has a distribution with 2 peaks centered at the ground truth depth. This might be due to a combination of the limited number of samples in the noisier histograms, and also the depth discretization of 0.5cm of the recovered transient pixel used for depth estimation. 
\end{itemize}

\noindent \textbf{Summary: } Overall, in this simplified imaging scenario that only requires denoising to resolve depth errors, iToF2dToF consistently displays better robustness to noise than the evaluated baselines. Furthermore, despite the small depth bias observed in all models, the absolute depth accuracy on real data of the data-driven models (trained on synthetic data) appears to be comparable to a traditional method.

\clearpage

\vspace{-0.25in}


\clearpage

\clearpage
\section{Real-world Experiments with Ground Truth}
\label{sec:supplement-4_mpi_results_with_gt}

In this section we present the results for the real-world test dataset with partial ground truth depths that we collected. The dataset contains 28 instances that come from 7 scenes captured at 4 exposure times ((1, 0.5, 0.2, 0.1ms). 

\smallskip

\begin{figure}[h]
\centering
    \includegraphics[width=\textwidth]{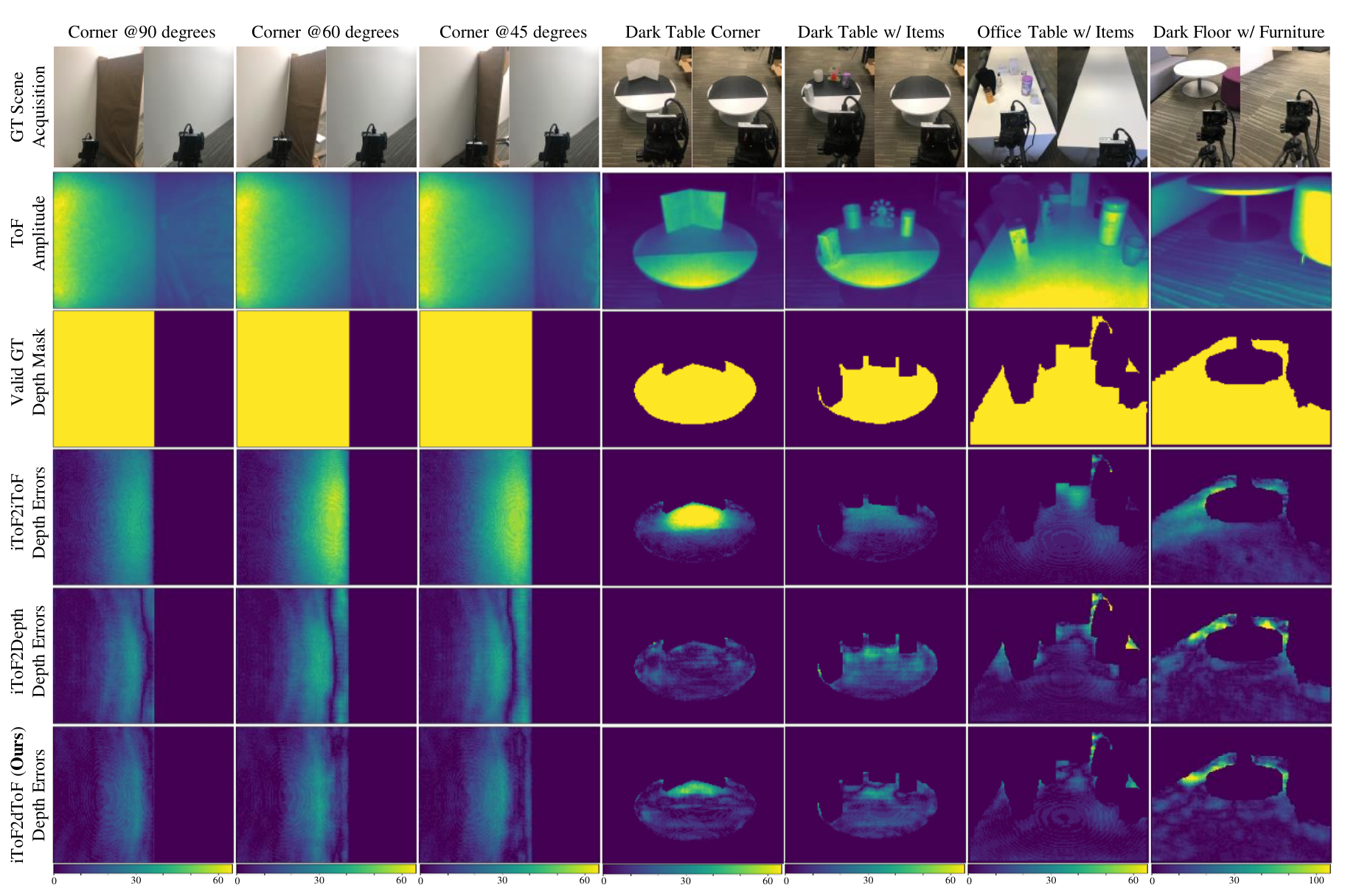}
\vspace{-0.3in}
\caption{\textbf{Real-world Depth Errors at 1ms Exposure Time.} Depth errors for iToF2dToF and two other data-driven baselines. Partial ground truth depths are obtained as described above in Section \ref{sec:supplement-4_mpi_results_with_gt}. The percentile MAE calculated for all models, over these scenes and the scenes in Section \ref{sec:appendix_quantitative_errors}, is reported in Table 3 of the main document.
}

\label{fig:supplement_quantitative_multiSNR_errors_1ms}
\end{figure}

\noindent \textbf{Ground Truth Acquisition Procedure: } To obtain ground truth depths we captured and average 100 frames of an empty scene with negligible MPI (e.g., a wall or an empty table as shown in the first row of Figure \ref{fig:supplement_quantitative_multiSNR_errors_1ms}). We use the noise-free dual-frequency data to compute ground truth depths using the Phasor method described in the main document in Section 6.1. Afterwards, we capture the same scene a second time, but with objects placed in it. Finally, we manually create a mask that indicates the regions where ground truth depths can be obtained from the first capture, as shown in the third row of Figure \ref{fig:supplement_quantitative_multiSNR_errors_1ms}. 

\vspace{-0.15in}

\subsection{Real-world Quantitative Evaluation}

\begin{figure}[h]
\centering
    \includegraphics[width=\textwidth]{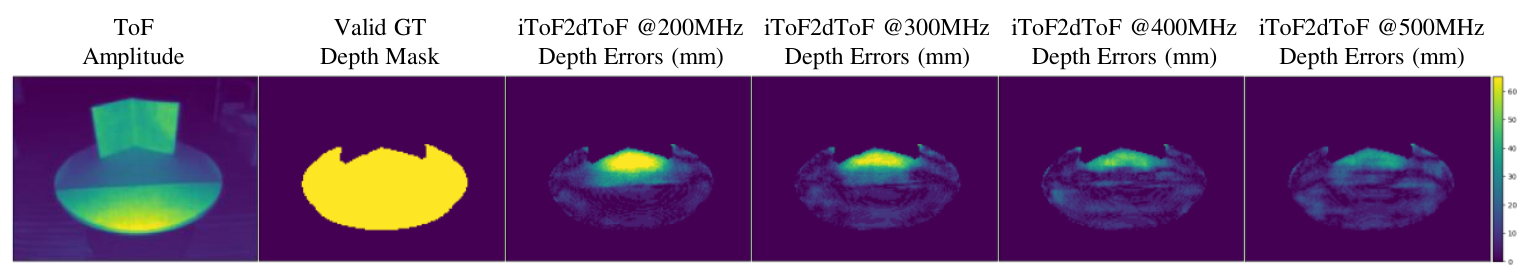}
\vspace{-0.2in}
\caption{\textbf{Effect of Max Frequency in iToF2dToF Depth Errors.} As we increase the maximum frequency used in iToF2dToF, the depth errors due to MPI decrease. 
}
\label{fig:supplement_itof2dtoF_freq_vs_errors}
\end{figure}

\noindent \textbf{Main Observations: } Figure \ref{fig:supplement_quantitative_multiSNR_errors_1ms} shows the depth errors for a denoising network (iToF2iToF), an end-to-end network (iToF2Depth), and for iToF2dToF. iToF2dToF achieves the lowest depth errors in 6 out of 7 of the scenes. 
The only scene that iToF2dToF does not outperform iToF2Depth was in the `Dark Table Corner' scene (4th column). 
There are two explanations for this result. 
First, these corner-like scenes are well-represented in the dataset, which as we have discussed, is the ideal case for end-to-end models like iToF2Depth.
And second, as we observe in Figure \ref{fig:supplement_itof2dtoF_freq_vs_errors}, 400MHz may not be a high enough frequency to completely mitigate MPI in this scenario. 
Nonetheless, iToF2dToF was able to consistently achieve lower depth errors in the more complex scenes (columns 5-7).
In the Appendix \label{sec:appendix_quantitative_errors} we show the results for these scenes using 0.5ms, 0.2ms, and 0.1ms exposure times. In these lower exposure settings, the performance of iToF2dToF degrades gracefully, and the dominant errors start being due to noise and not MPI.

\smallskip

\noindent \textbf{Summary: }Our quantitative evaluation shows that data-driven methods, trained on synthetic data only, have the capability to generalize to real-data. Although, iToF2dToF does achieve the lowest percentile MAE computed over the 28 instance test set we captured, we also observed the strengths of an end-to-end model, like iToF2Depth, in scene configurations that were well-represented in the synthetic dataset. Finally, as we decreased the SNR (exposure time), iToF2dToF's performance degraded gracefully as seen in the Appendix Figures \ref{fig:supplement_quantitative_multiSNR_errors_0.5ms}, \ref{fig:supplement_quantitative_multiSNR_errors_0.2ms}, \ref{fig:supplement_quantitative_multiSNR_errors_0.1ms}.

\subsection{Extremely Low SNR Depth Reconstruction}

\begin{figure}[h]
\centering
    \includegraphics[width=0.9\textwidth]{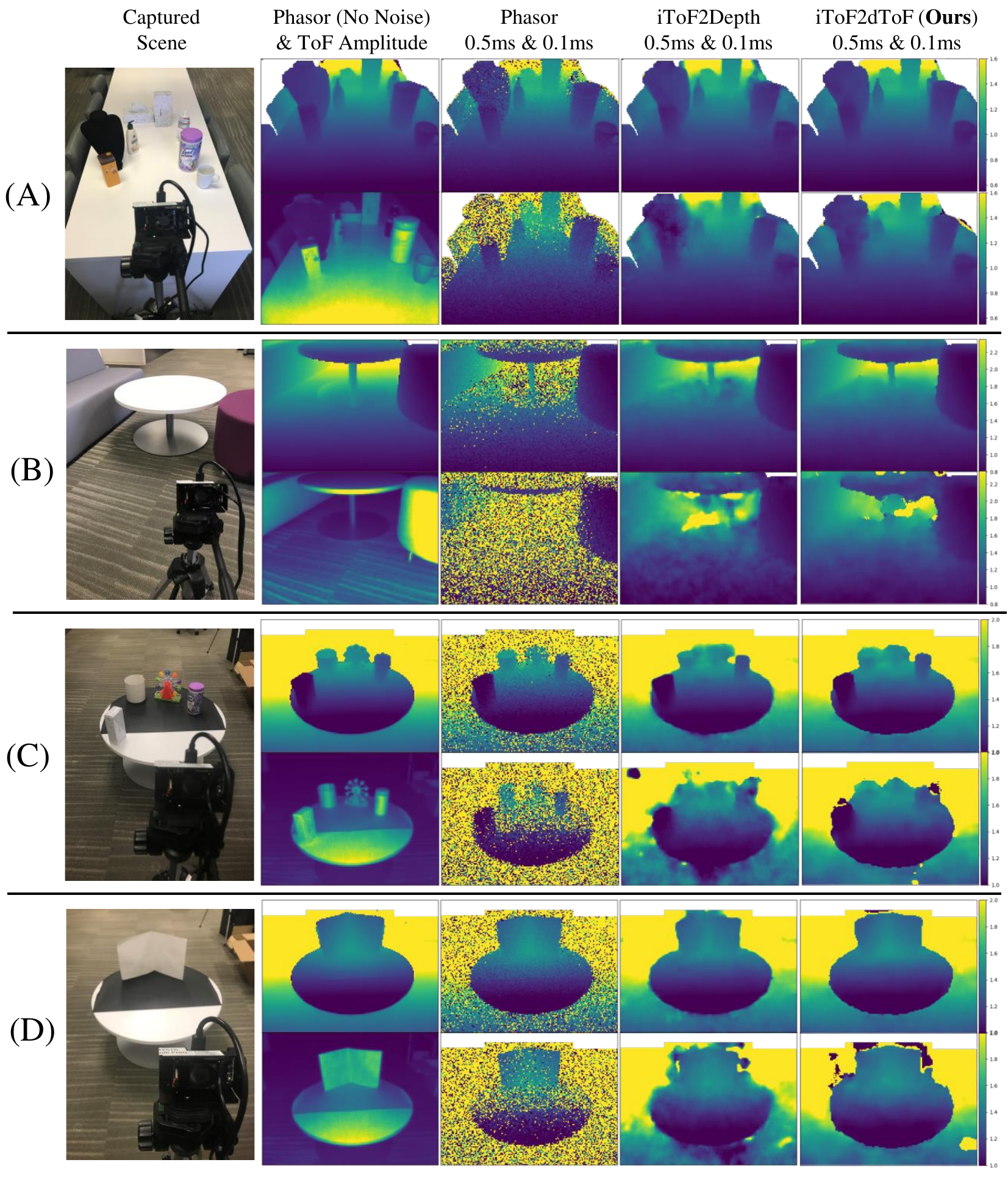}
\vspace{-0.1in}
\caption{\textbf{Extreme Low SNR Depth Reconstructions.} Recovered depths for multiple scenes at low exposure times (0.5ms and 0.1ms) . 
The Phasor (No Noise) images provide an approximate view of how the correct depth image should look like. 
The Phasor depth images (3rd column) does not apply any denoising to the data, and it is a useful image to identify low and high SNR regions. 
For visualization purposes, we mask pixels (white regions) that exhibit phase wrapping or that were still noisy in the ``noiseless'' Phasor image. 
We find that even at extremely low exposure times, iToF2dToF, is still able to recover accurate depths in some regions of the image. }

\label{fig:supplement_qualitative_multiSNR_depths_0.5-0.1ms}
\end{figure}

\noindent \textbf{Main Observations: } Figures \ref{fig:supplement_qualitative_multiSNR_depths_0.5-0.1ms} and \ref{fig:supplement_qualitative_multiSNR_depths_1.0-0.2ms} show the depth images reconstructed by different methods at very low exposure times (0.1ms-1ms). At extremely low SNR, (e.g., 0.1ms or 0.2ms), all depth images contain some artifacts. However, we observe that even at such low exposure times, iToF2dToF, is able to estimate accurate depths in some regions, and overall displays fewer artifacts than other models. 

\noindent \textbf{Summary: } 
In this section, we evaluated iToF2dToF at extremely low exposure times and empirically verified that the quality of the reconstructed depth maps degraded gracefully. 

\clearpage
\section{Additional Real-world Results}
\label{sec:supplement-5_real_data_results}


\subsection{Additional Qualitative Depth Reconstructions}
\label{sec:supplement-5_real_qualitative_depths}

In this section we present additional qualitative depth reconstructions for everyday real-world scenes. Despite the unavailability of ground truth depth in these scenes, we draw some useful observations from the quality of the reconstructions. 

\smallskip

\begin{figure}[h]
\centering
    \includegraphics[width=\textwidth]{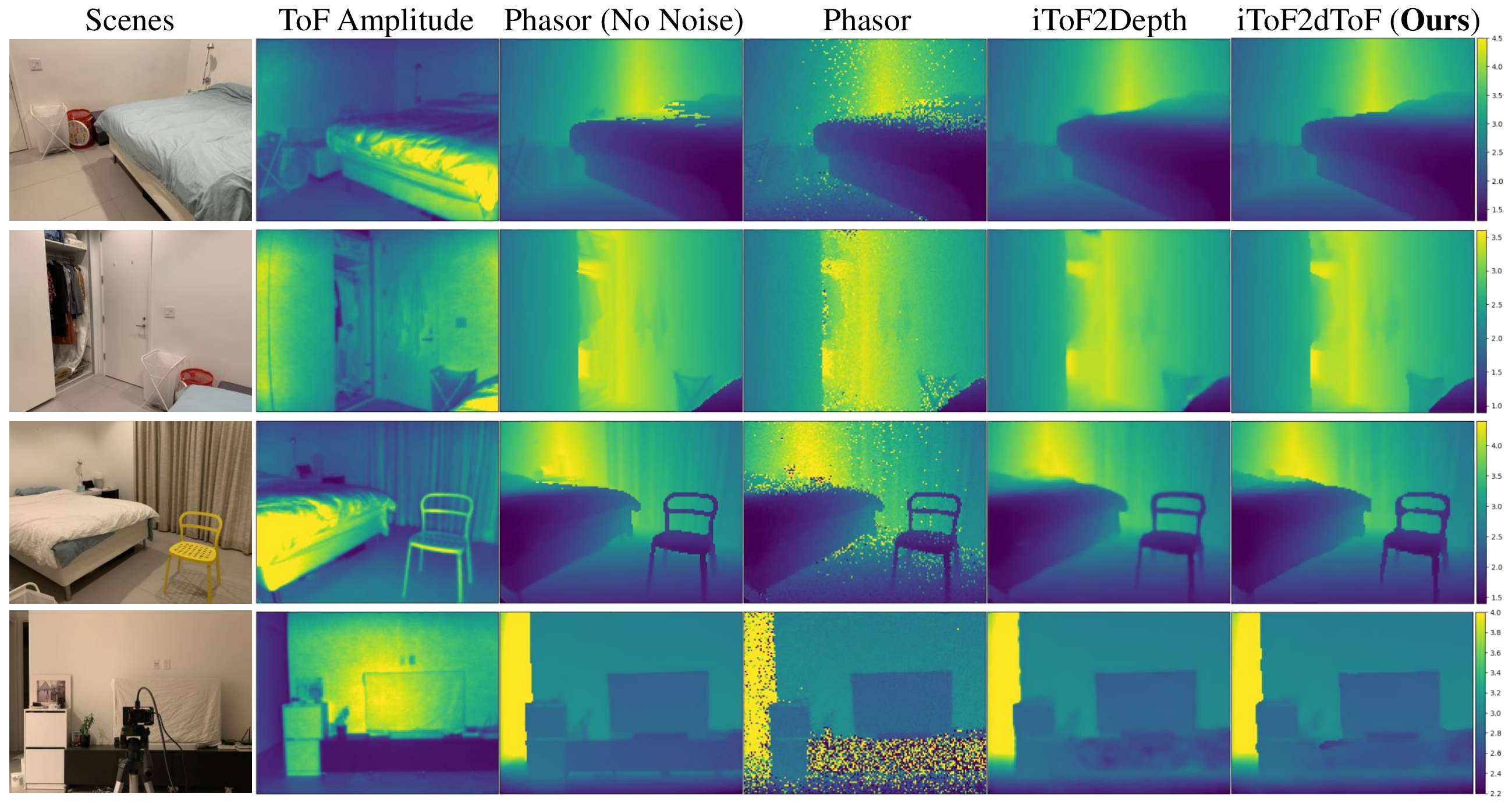}
\vspace{-0.1in}
\caption{\textbf{Real-world Depth Images}. Recovered depths by iToF2dToF and some of the baseline established in the main document. All scenes were acquired using a 1ms exposure time per frame. The Phasor (No Noise) is obtained by averaging 100 iToF frames and then compute depths using the phasor method, so it has minimal noise and can give us an approximate view of the ground truth depths. Both data-driven methods (iToF2Depth and iToF2dToF) exhibit good generalization and recover high-quality depth maps. \textit{Zoom in for better visualization}.}

\label{fig:supplement_qualitative_depths}
\end{figure}

\noindent \textbf{Main Observations: } Figure \ref{fig:supplement_qualitative_depths} shows the estimated depths by iToF2dToF and some of the baselines established in the main paper. The data-driven methods recover depth maps that are comparable to the traditional method Phasor (No Noise) in most regions with high enough SNR, indicating good generalization. We point out two specific points in the depth reconstructions:

\begin{itemize}
    \item \textbf{Low SNR Artifacts:} Low SNR regions can be identified by looking at the 4th column, the depth maps of the Phasor method without any denoising. In the bottom row, in extreme low SNR regions, the performance of data-driven methods degrades gracefully resulting in some artifacts. Nonetheless, in other low SNR regions the data-driven methods and iToF2dToF in particular, produce reliable depth estimates.
    \item \textbf{Edges in Depth Images: }In low-resolution images, depth discontinuities (edges) will look different across methods because they are dealt with differently. In iToF2Depth, the network is trained to predict flying pixels, because that is what the ground truth says. In iToF2dToF, we use a peak finding depth estimation algorithm so the depth will depend on the peak that is chosen between the background and foreground. Similar to iToF2dToF, Phasor, uses a look-up table algorithm. However, due to the low temporal resolution of the look-up table, the peaks of background and foreground blend, resulting again in flying pixels. Ultimately, these differences will be highly mitigated in higher resolution images. Please refer to Section \ref{sec:supplement-5_depth_image_refinement} for a follow-up analysis of this sitation.
\end{itemize}

\noindent \textbf{Summary: }Overall, data-driven methods trained on synthetic data appear to generalize well to real data, despite the modeling assumptions made when simulating the synthetic data. They are able to mitigate strong MPI (top row) and degrade gracefully at low SNR (bottom row). 

\clearpage

\subsection{Additional Specular MPI Correction Result}

In this section we present the depth reconstruction of a glossy table that exhibits semi-specular reflections. We use the same first peak finding algorithm for iToF2dToF that was described in the specular MPI section of the main document.

\medskip

\begin{figure}[h]
\centering
    \includegraphics[width=0.99\textwidth]{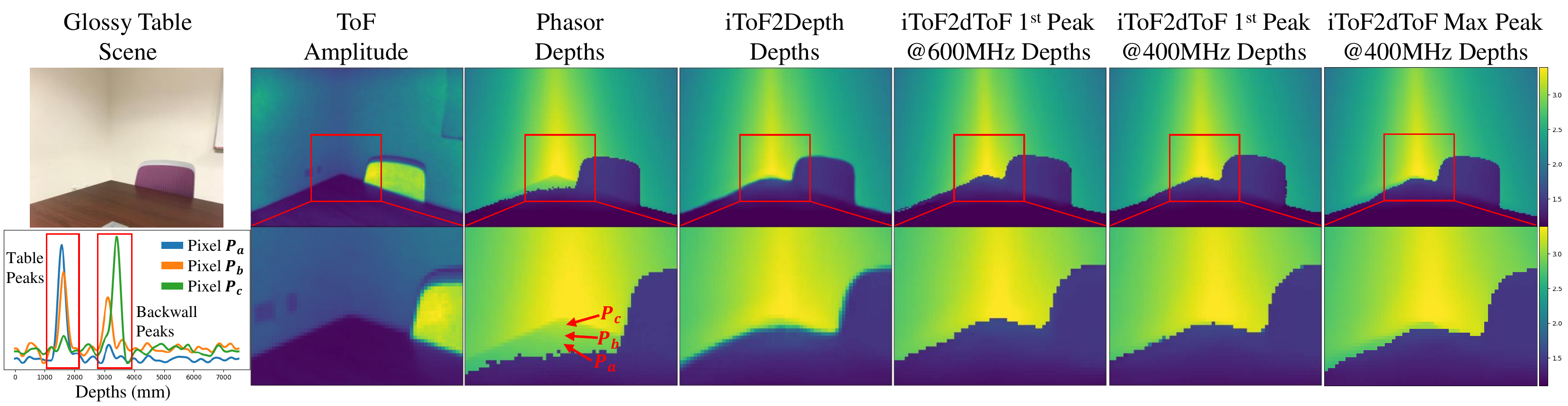}
\vspace{-0.1in}
\caption{\textbf{Glossy Table Depth Reconstruction}. The first row shows the estimated depths of the scene. The second row shows the zoomed in depth maps in the regions that displayed strong specular MPI. In the first column of the first row, we plot 3 transient pixels produced by iToF2dToF. As illustrated in the Phasor depth image, the selected transient pixels correspond to highly glossy points on the table. Pixel $P_a$ has a strong $1^{st}$ peak and weak $2^{nd}$ peak, while pixel $P_c$ has a weak $1^{st}$ peak and strong $2^{nd}$ peak. In these pixels the correct depth corresponds to the $1^{st}$ peak. This results in iToF2dToF $1^{st}$ peak (5/6th columns) producing a more accurate depth map than iToF2dToF Max Peak (7th column). \textit{If possible, zoom in for details like the `glossyness' of the table, the improvements in the depth reconstruction, and the reconstructed transient pixels.}}
\label{fig:supplement_specular_mpi}
\end{figure}

\noindent \textbf{Main Observations: } The data-driven methods (iToF2Depth and iToF2dToF) are less susceptible than the Phasor method to depth errors caused by the specular reflections on the glossy table, as shown in Figure \ref{fig:supplement_specular_mpi}. Although, iToF2Depth recovers accurate depths in the less specular regions of the table, it starts predicting erroneous depths in the more specular regions located towards the corner of the table. 
This is expected because, as observed in transient pixel $P_c$, the direct reflection for these pixels is weak. 
Finally, using iToF2dToF with a $1^{st}$ peak finding algorithm results in the most accurate reconstruction. 
However, the pixels right at the corner of the table are too specular, which results in iToF2dToF choosing the specular peak as the correct peak.
Interestingly, in this example, we found it beneficial to use the iToF2dToF model trained to extrapolate up to 600MHz.

\medskip

\noindent \textbf{Summary: }When imaging complex materials, by analyzing the dToF representation, we can design appropriate peak finding algorithms that can result in higher-quality depth reconstructions. Furthermore, in this example there appears to be a small benefit in extrapolating to a higher frequency. Exploring methods and the benefits of extrapolating to higher frequencies is an interesting direction for future work.

\clearpage
\subsection{Depth Image Refinement}
\label{sec:supplement-5_depth_image_refinement}

In this section we analyze the differences on the depth image edges across different methods. These differences are mainly applicable to low-resolution images, like the ones used in this paper, and as the resolution increases these differences might become negligible.

\smallskip

\begin{figure}[h]
\centering
    \includegraphics[width=0.99\textwidth]{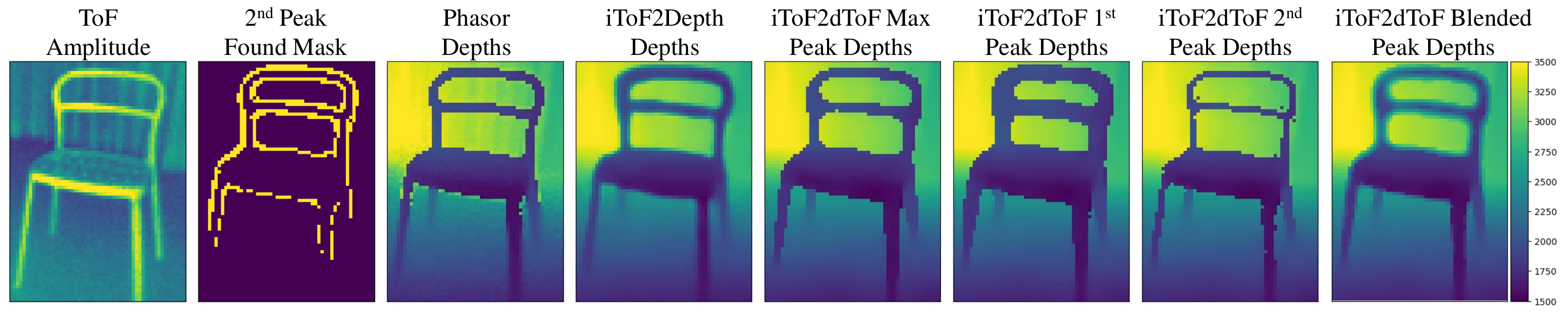}
\vspace{-0.1in}
\caption{\textbf{Depth Image Edge Comparison}. For Phasor, iToF2Depth, and iToF2dToF Max Peak we use the depth estimation algorithms that have been used throughout the paper. For iToF2dToF $1^{st}$ peak, we use the same algorithm that was used for specular MPI correction. For iToF2dToF $2^{nd}$ peak, we compute edges with a canny edge detector on the ToF amplitude image, take the max peak depth image, and replace the depths at edge pixels with the $2^{nd}$ peak depths. To find the $2^{nd}$ peak we use the same peak finding algorithm described for specular MPI. Finally, for iToF2dToF blended peaks, we do a weighted averaging of the $1^{st}$ and $2^{nd}$ peak depth images, where the weights are determined by the peak heights.}
\label{fig:supplement_depth_edges}
\end{figure}

\noindent \textbf{Main Observations: } Figure \ref{fig:supplement_depth_edges} compares the recovered depth maps across different methods. The depth predicted for edge pixels varies across methods. iToF2Depth produces depths that blur the foreground and background (flying pixels). iToF2dToF, depending on the peak finding algorithm, can choose the foreground ($1^{st}$ peak), the background ($2^{nd}$ peak), somewhere in between (blended peaks), or the peak with highest intensity (max peak). Ultimately, the correct depth for an edge pixel is ambiguous and may depend on the application. For instance,

\begin{itemize}
    \item \textbf{Obstacle Avoidance:} Consider a robot vacuum using depth images from an iToF camera to navigate a living room. In this scenario, a depth image where the object edges are dilated, as in iToF2dToF $1^{st}$ peak, might be beneficial to ensure that the vacuum avoids the objects. 
    \item \textbf{Robotic Grasping} Consider a robotic hand picking up objects. In order to ensure a tight grasp of the object it may be beneficial to use a depth image with thinned edges such as iToF2dToF $2^{nd}$ peak. 
\end{itemize}

\noindent \textbf{Summary: }The correct depth for an edge pixel will most likely be determined by the application. In iToF2dToF, we can adjust the depth estimation algorithm to either choose the foreground depth (first peak) or background depth (second peak) depending on the scenario. This is another example of the benefits of estimating a flexible intermediate representation. Nonetheless, as we increase the depth image resolution these edge differences across methods may become negligible and possibly irrelevant.

\clearpage
\section{Additional Synthetic Data Results}
\label{sec:supplement-5_synthetic_data_results}

\subsection{Pixel-wise Frequency Analysis}

In this section we compare the depth errors obtained by iToF2dToF models that extrapolate to 100MHz and 400MHz. Furthermore, we show different transient pixels reconstructed by iToF2dToF and compare them to the ideal dToF case. The ideal dToF reconstructions are obtained using the ground truth frequency data we use to train iToF2dToF. 

\vspace{-0.1in}

\begin{figure}[h]
\centering
    \includegraphics[width=0.99\textwidth]{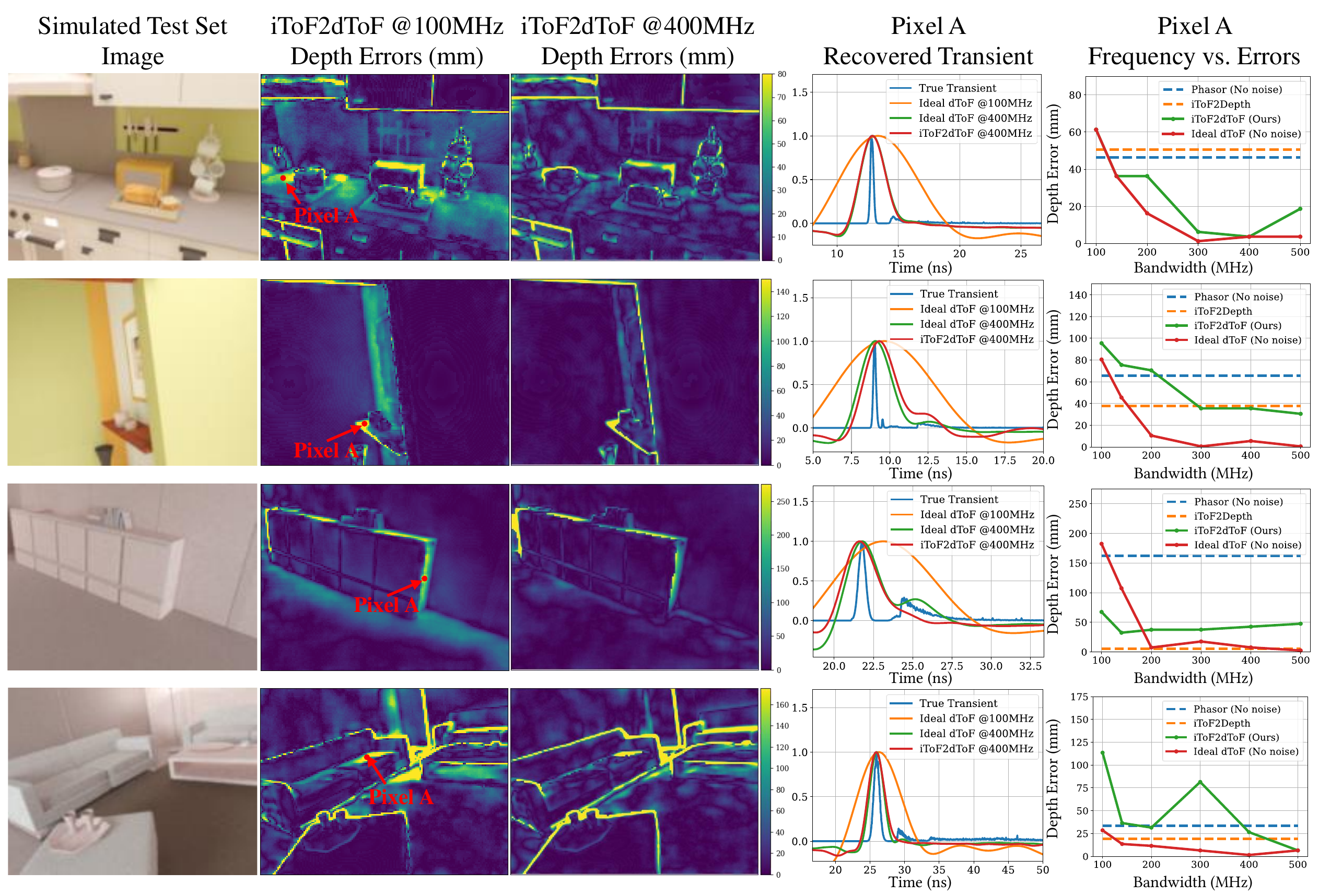}
\vspace{-0.15in}
\caption{\textbf{Pixel-wise Error and Frequency Analysis}. The second and third columns show the absolute depth errors in 4 different test set scenes. For each scene, we also plot one transient pixel (fourth column) reconstructed by iToF2dToF @400MHz. In the transient pixel plots only iToF2dToF contains noise. The true and ideal dToF transient pixels do not have noise. Finally, in the last column, we plot the depth errors of that single pixel for various models as a function of frequency. The performance of the Phasor and iToF2Depth baselines does not depend on frequency. On the other hand, for the iToF2dToF and the ideal dToF models, there is a clear trend of errors decreasing as we increase frequency.  }
\label{fig:supplement_pixel_freq_analysis}
\end{figure}

\begin{figure}[h]
\centering
    \includegraphics[width=0.9\textwidth]{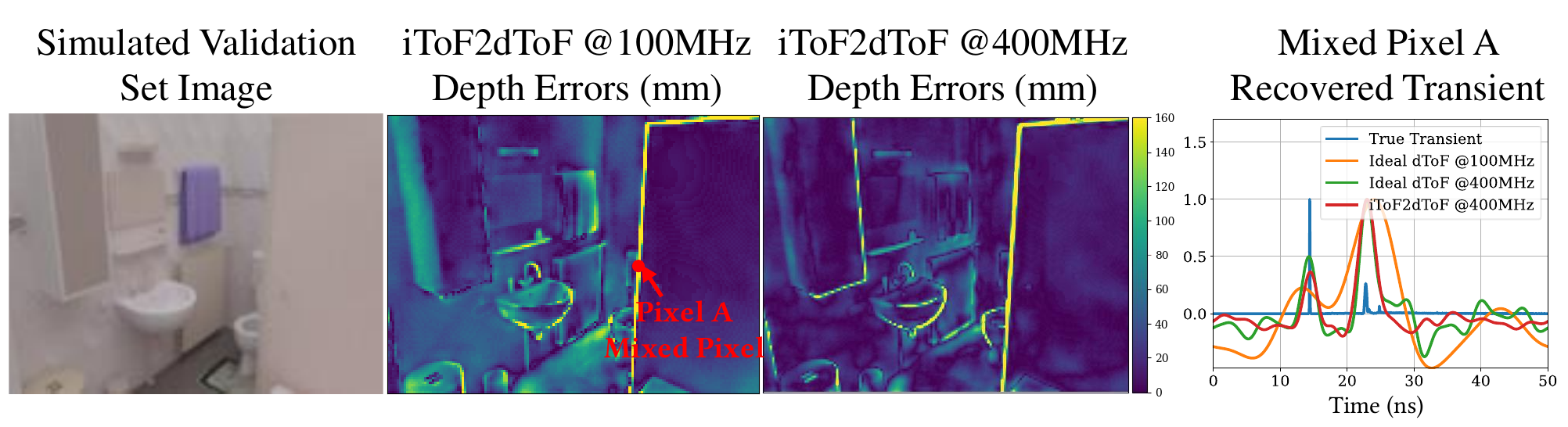}
\vspace{-0.1in}
\caption{\textbf{Sparse Transient Example}. The second and third columns show the absolute depth errors for one validation set scene. The last column plots the sparse transient waveforms for an edge pixel. iToF2dToF is able to accurately estimate the sparse peak locations.  }

\label{fig:supplement_mixed_pixel_transient}
\end{figure}

\noindent \textbf{Main Observations:} Figures \ref{fig:supplement_pixel_freq_analysis} and \ref{fig:supplement_mixed_pixel_transient} compare the depth errors of iToF2dToF at different maximum frequencies. An iToF2dToF model only interpolating frequencies up to 100MHz (second column), continues to exhibit large depth errors due to MPI. Extrapolating up to 400MHz mitigates these errors to a large extent (third column). To understand the effect of frequency on performance we look at iToF2dToF's transient pixel reconstructions (fourth column), in the following cases: 

\begin{itemize}
    \item \textbf{Non-Sparse MPI: }As observed in the fourth column of Figure \ref{fig:supplement_pixel_freq_analysis}, even for an ideal dToF transient pixel with frequencies up to 100MHz (orange line), the non-sparse MPI observed in the true transient (blue line), causes a significant bias in the maximum peak's location. This bias is largely mitigated if we include frequencies up to 400MHz (green line). Despite the presence of noise in iToF2dToF's input, the model is able reconstruct transient pixels (red line) that resemble the ideal dToF case.  
    \item \textbf{Sparse MPI at Edges: }The fourth column of Figure \ref{fig:supplement_mixed_pixel_transient} shows the reconstruction of the transient pixel at a depth discontinuity (edge). As discussed in Sections \ref{sec:supplement-5_real_qualitative_depths}
    and  \ref{sec:supplement-5_depth_image_refinement} the ground truth depth for these pixels is ambiguous. Nonetheless, learning to estimate the correct transient waveform for these sparse transient pixels is still important. And as shown in Figure \ref{fig:supplement_mixed_pixel_transient}, iToF2dToF's capability to estimate the sparse peak locations is comparable to the ideal dToF case.  
    \item \textbf{Failure Case: }The third row in Figure \ref{fig:supplement_pixel_freq_analysis} shows one transient pixel reconstruction that led to elevated depth errors. Although, iToF2dToF does a reasonable approximation of the transient pixel, the peak has a small bias to the left. This particular transient pixel was challenging to reconstruct at all frequencies, as seen in the depth errors plotted in the last column. This failure could be caused by extremely low SNR levels for those pixels, or insufficient generalization by iToF2dToF. The latter can be easily solved by increasing the dataset size or using more complex architectures and training procedures like the ones proposed in previous works.
\end{itemize}

\medskip 

\noindent \textbf{Summary:} Overall, there is a clear relationship between the depth accuracy of a ToF system and its bandwidth (maximum frequency). Our results show that, data-driven iToF-based transient imaging methods, like iToF2dToF, can extrapolate to higher frequencies and effectively increase the bandwidth of the system. 
Interestingly, the depth errors, in the ideal dToF case, plateau around 400MHz (red line in last column of Figure \ref{fig:supplement_pixel_freq_analysis}). 
This empirical finding is consistent with previous analysis on the performance of iToF systems as a function of frequency \cite{gupta2015phasor, kadambi2016macroscopic}.
Furthermore, this suggests that future work should not solely focus on extrapolating to higher frequencies, but also focus on improving the quality of the transient reconstructions. One simple approach to achieve this could be to input 3-4 frequency measurements, which continues to be practical. It is important to note that the maximum frequency required to resolve MPI will depend on the scale of the scene \cite{gupta2015phasor}, and our analysis has largely focused on indoor scenes where iToF cameras are most often used.


\clearpage

\subsection{Incorporating Amplitude Information}
\label{sec:supplement-5_synthetic_data_results_with_amplitude}

We find that using amplitude information helps improve the performance of all models on synthetic data. Unfortunately, without the use of extensive calibration \cite{guo2018tackling} or more complex architectures and learning procedures \cite{agresti2018deep, agresti2019unsupervised}, the models trained with amplitude information did not generalize well to real data. In this section, we compare the performance on synthetic data of models trained with and without amplitude, and also show the failure case on real data. Furthermore, we show that the robustness of iToF2dToF to noise continues to hold when amplitude information is added.

Similar to our setup in the main paper, the inputs to the networks are dual-frequency iToF measurements at 20 and 100MHz. Instead of normalizing each frequency measurement by its own amplitude, we normalize all frequencies by dividing by the lowest frequency amplitude. The amplitude ratio across frequencies is known to contain useful information about MPI \cite{freedman2014sra}, and has been used as an input in previous data-driven ToF models \cite{agresti2018deep}. However, for the input frequencies we find that the ratio makes training less stable due to an increased number of divisions by 0 due to noise. To resolve this we apply light gaussian smoothing (stddev = 0.5px) on the input amplitudes, which stabilized training and improved performance.

\medskip

\begin{figure}[h]
\begin{floatrow}
\capbtabbox{%
\setlength\tabcolsep{3.25pt}
\resizebox{0.5\textwidth}{!}
{
\begin{tabular}{lcccc}
\hline
\multicolumn{5}{c}{\textbf{Synthetic Test Set Percentile MAE (mm)}}                                                                                 \\ \hline
\textbf{Model}                                             & \textbf{0-75\%} & \textbf{75-85\%} & \textbf{85-95\%} & \textbf{95-99\%} \\ \hline
Phasor (No Noise) \cite{gupta2015phasor}                                                  & 9.53           & 29.58            & 46.37            & 94.79            \\ \hline
iToF2Depth Baseline                                                & 7.49           & 21.86   & 34.99   & 88.03            \\ \hline
iToF2Depth w/ Amp.                                                & 7.35           & 21.27   & 33.66   & 81.76            \\ \hline
iToF2dToF @200MHz                                                  & 7.85   & 23.26            & 36.69            & 78.60   \\ \hline
iToF2dToF @200MHz w/ Amp                                                 & 7.03   & 21.60            & 34.79            & 75.77   \\ \hline
iToF2dToF @400MHz                                                  & 7.19   & 20.42            & 32.18            & 71.56   \\ \hline
iToF2dToF @400MHz w/ Amp                                                 & 6.24   & 18.20            & 29.68            & \textbf{69.30}   \\ \hline
iToF2dToF @500MHz                                                  & 7.22   & 20.40            & 32.17            & 72.12   \\ \hline
iToF2dToF @500MHz w/ Amp                                                 & \textbf{6.19}   & \textbf{17.92}            & \textbf{29.52}            & 69.88   \\ \hline
iToF2dToF @600MHz                                                  & 7.33   & 20.66           & 32.76            & 76.13   \\ \hline
iToF2dToF @600MHz w/ Amp                                                 & 6.26   & 18.17            & 30.02            & 70.93   \\ \hline

\end{tabular}

}}
{%
  \caption{Percentile MAE calculated over the test set simulated over a wide range of SNR levels. All models benefit from being trained with amplitude. Furthermore, we continue to see that iToF2dToF performance plateaus around 400MHz, even if we train with amplitude information.}%
  \label{tab:supplement_synthetic_test_results}
}
\ffigbox{%
\centering
\centerline{
	\includegraphics[width=0.98\linewidth]{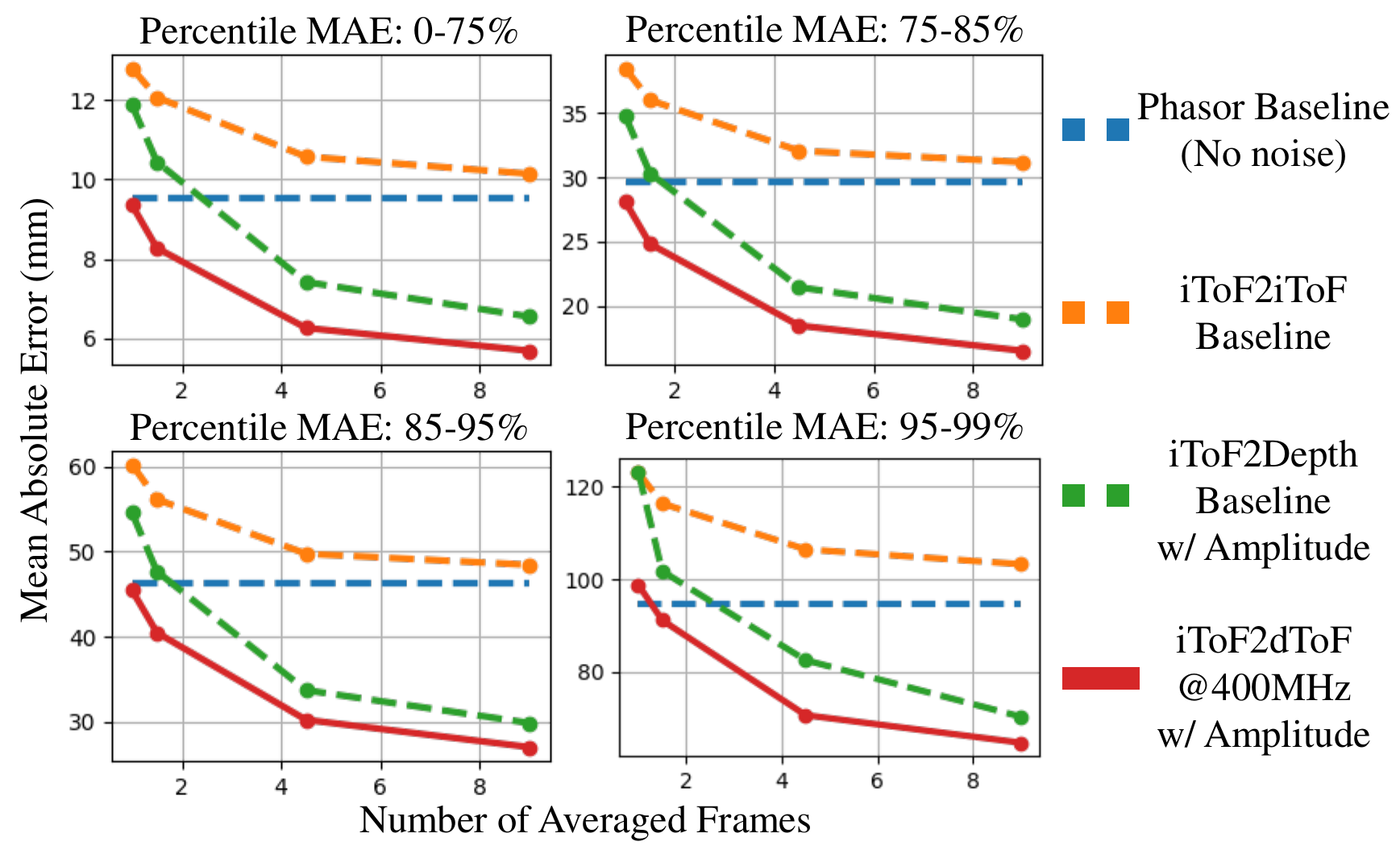}
}
}{%
\caption{\textbf{Noise vs. Error on Synthetic Test Set}. We simulate the same test set at multiple SNR levels. 
As we lower SNR, we continue to see a widening in the performance gap  between iToF2dToF and iToF2Depth when amplitude is used.}
\label{fig:supplement_simulation_noise_withamplitude}

}

\end{floatrow}
\end{figure}

\begin{figure}[h]
\centering
    \includegraphics[width=0.99\textwidth]{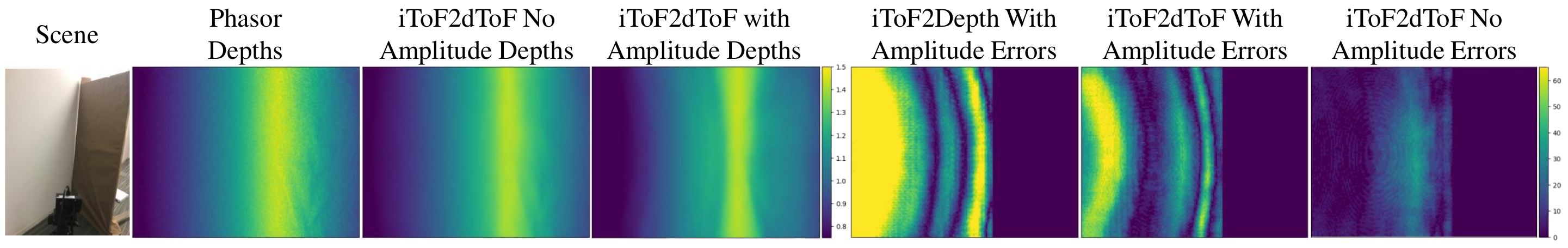}
\vspace{-0.1in}
\caption{\textbf{Real-world Result of Models Trained with Amplitude Information.} We test the data-driven models trained with amplitude information as described above on one of our real-world scenes with partial ground truth. The depth errors obtained by the data-driven models trained with amplitude information (5th and 6th columns) are significantly higher than a model not trained without amplitude information in the input (last column). 
}
\label{fig:supplement_depths_models_with_amplitude}
\end{figure}

\noindent \textbf{Main Observations:} Table \ref{tab:supplement_synthetic_test_results} shows performance of the different models evaluated in the paper with and without amplitude information. Providing amplitude as input improves the performance of all models. Furthermore, Figure \ref{fig:supplement_simulation_noise_withamplitude} shows a similar trend in which the performance gap between of iToF2dToF and iToF2Depth widens as we lower SNR. Therefore, we expect that the results we show in this paper will continue to hold if we incorporate amplitude information as input. Unfortunately, as observed in Figure \ref{fig:supplement_depths_models_with_amplitude} the data-driven models trained with amplitude information did not perform well on real-data.

\medskip 

\noindent \textbf{Summary:} Overall, providing amplitude information as part of the input to the data-driven model is beneficial for all the methods we evaluate. The challenge arises in the generalization to real data. Certain non-idealities in the hardware not captured in the synthetic data or calibration errors can lead to large depth errors in the learned models.

\vspace{-0.2in}

\clearpage

\subsection{Increasing Network Size and Training Complexity}

In this section, we compare iToF2dToF with an additional iToF2Depth model that uses a larger network. This iToF2Depth model uses the same architecture as \cite{su2018deep}, where 9 residual layers are added between the encoding and decoding stages of the U-net. This results in a model with 12.495M parameters, which is $\sim$7x more parameters than the other data-driven models in this paper. We refer to this model as \textit{iToF2Depth-E2EToF}, and train it in the exact same way as iToF2Depth. Additionally, we also train another iToF2Depth-E2EToF model using the depth gradient loss regularizer ($L_{smooth}$) used by \cite{su2018deep} with the same hyperparameter ($\lambda_s = 0.0001$). 


\medskip

\begin{figure}[h]
\centering
    \includegraphics[width=0.7\textwidth]{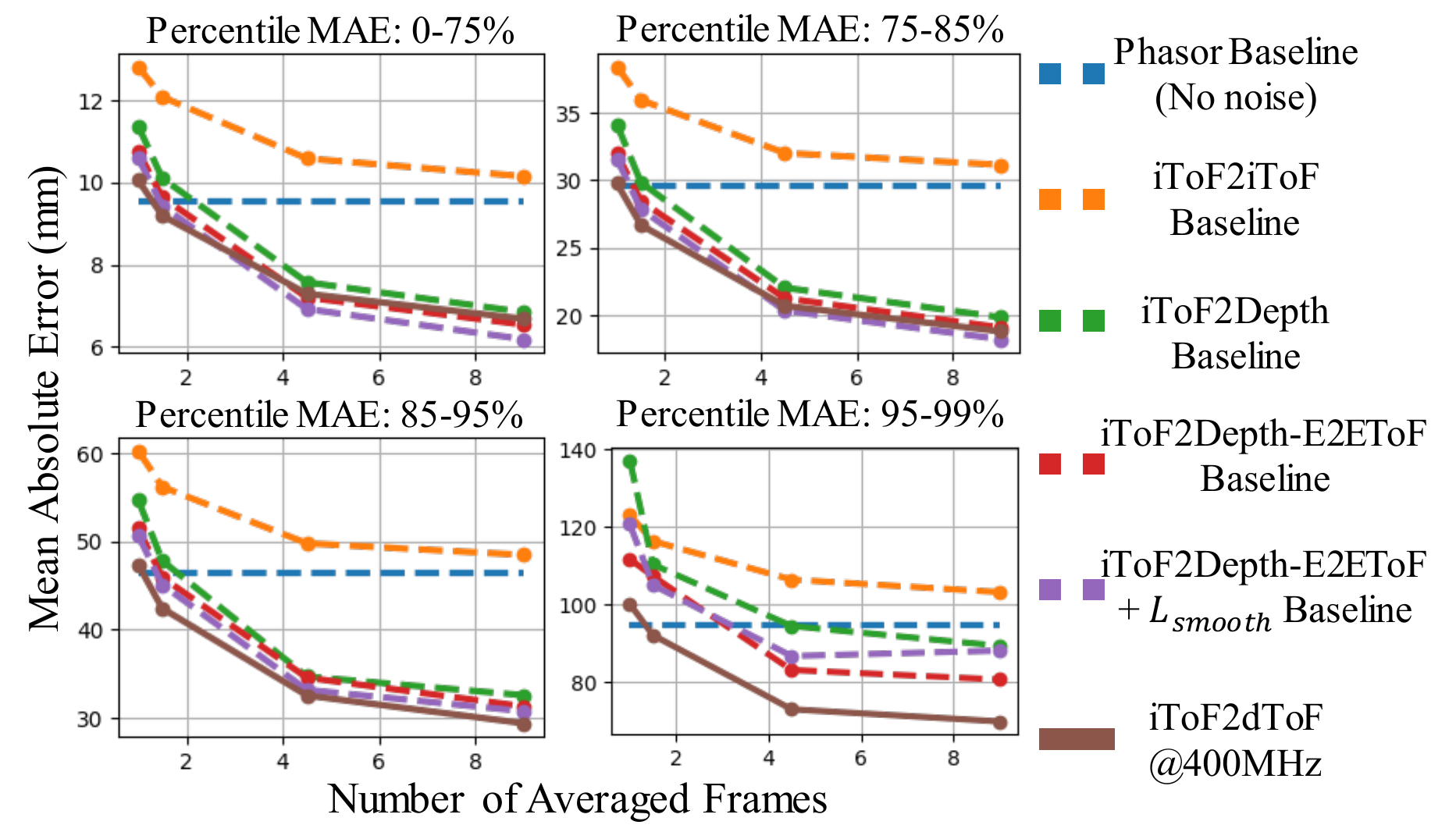}
\vspace{-0.1in}
\caption{\textbf{Noise vs. Error on Synthetic Test Set}. We simulate the same test set at multiple SNR levels as described in Section \ref{sec:supplement-2_mftof_dataset}. 
Each model was trained with images containing the full range of SNR levels. 
For all SNR levels, iToF2dToF outperforms all baselines. 
In particular, as we reduce the number of frames (reduce SNR) the performance gap between iToF2dToF and the end-to-end models widens. 
At the lowest SNR settings, iToF2Depth performs comparably or worse than a simple denoising network (iToF2iToF).
Using a larger models and regularization, like E2EToF, helps improve the robustness to noise of the end-to-end model, at the expense of using $\sim$7x more parameters and increasing training complexity. 
}
\label{fig:supplement_synthetic_noise_results}
\end{figure}

\noindent \textbf{Main Observations:} Figure \ref{fig:supplement_synthetic_noise_results} shows that using a larger network and additional regularization for iToF2Depth provides an incremental performance gains across most SNR levels. Nonetheless, despite using $\sim$7x more parameters and more complex loss functions, the performance of the iToF2Depth-E2EToF models continue to perform worse than iToF2dToF at low SNR. Lighter-weight networks that use less memory and require less computations, such as the one used for iToF2dToF and most of the baselines, are desirable in mobile applications where iToF cameras are often used.

\clearpage

\subsection{iToF2dToF with Loss on Depth Representation}

iToF2dToF models are trained by computing an L1 loss with respect to the band-limited frequency-domain transient representation. In this section we compare the performance of iToF2dToF with another iToF2dToF model that is trained with an L1 loss function with respect to the depth estimates. We refer to this new model as \textit{iToF2dToF2Depth}.

Transient pixels encode depths in their peaks. Specifically, the argmax of transient pixels without strong specular reflections or cross-talk, encodes depths. Since argmax is not a differentiable operation, we train iToF2dToF2Depth using a softargmax operator with a parameter $\beta=300$ \cite{kendall2017end}.

\begin{figure}[h]
\centering
    \includegraphics[width=0.6\textwidth]{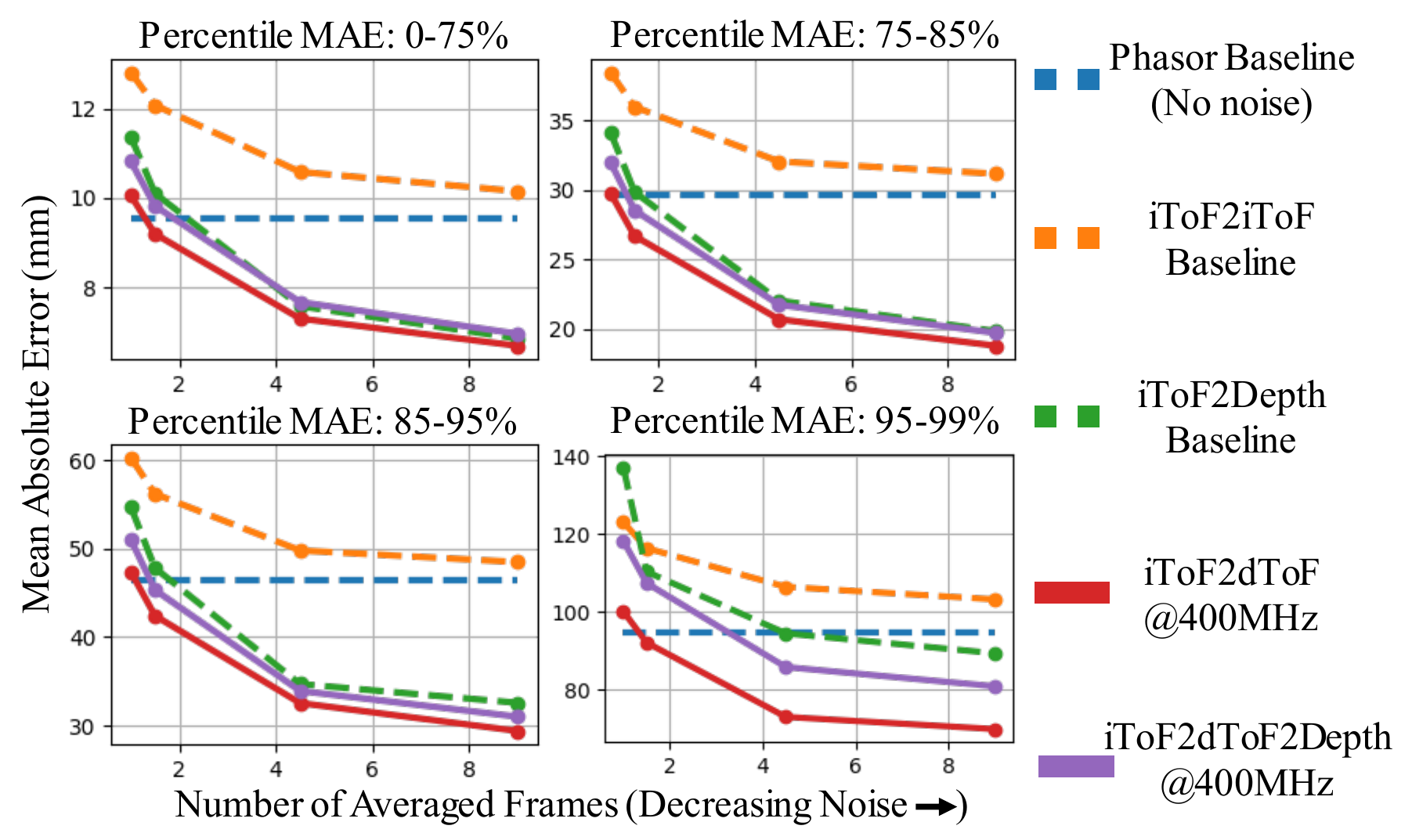}
\vspace{-0.1in}
\caption{\textbf{Noise vs. Error on Synthetic Test Set}. We simulate the same test set at multiple SNR levels as described in Section \ref{sec:supplement-2_mftof_dataset}. 
Each model was trained with images containing the full range of SNR levels. 
For all SNR levels, iToF2dToF outperforms all baselines. 
In particular, in the lowest SNR pixels (percentile 95-99\%) iToF2dToF provides significant improvements over the models trained using depths as supervision (i.e., iToF2Depth and iToF2dToF2Depth).
}
\label{fig:supplement_simulation_noise_itof2dtof2depth}
\end{figure}

\begin{figure}[h]
\begin{floatrow}
\ffigbox{%
\centering
\centerline{
	\includegraphics[width=0.98\linewidth]{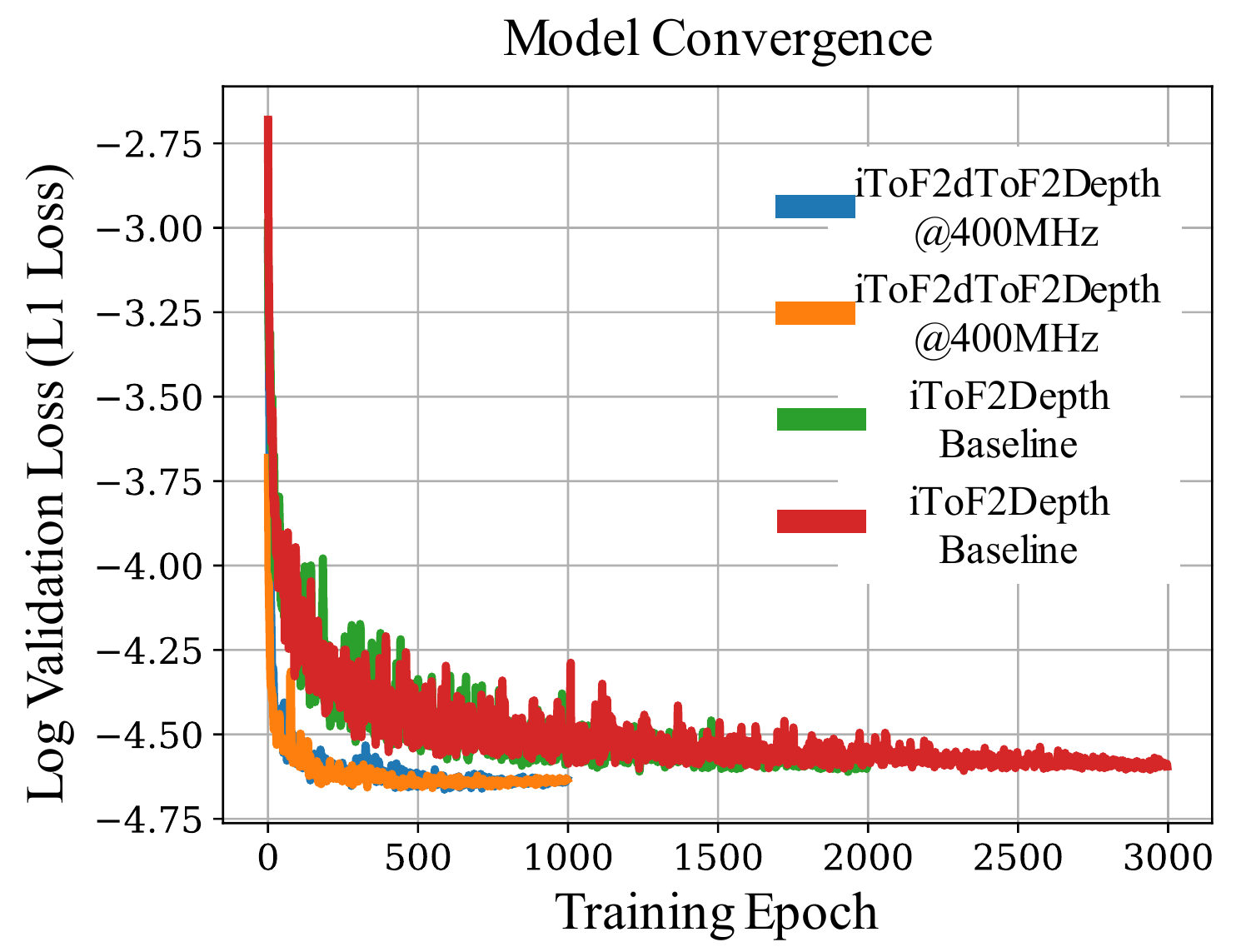}
}
}{%
\caption{L1 loss with respect to depth, computed over the validation set at each epoch during training. Both iToF2dToF2Depth training runs (blue and orange lines) converged withing 500 epochs, while both iToF2Depth training runs (green and red lines) required more than 1500 epochs to converge. For visualization purposed, a box filter of size 3 is applied to the validation curves of all models.}
\label{fig:supplement_validation_losses}
}
\ffigbox{%
\centering
\centerline{
	\includegraphics[width=0.98\linewidth]{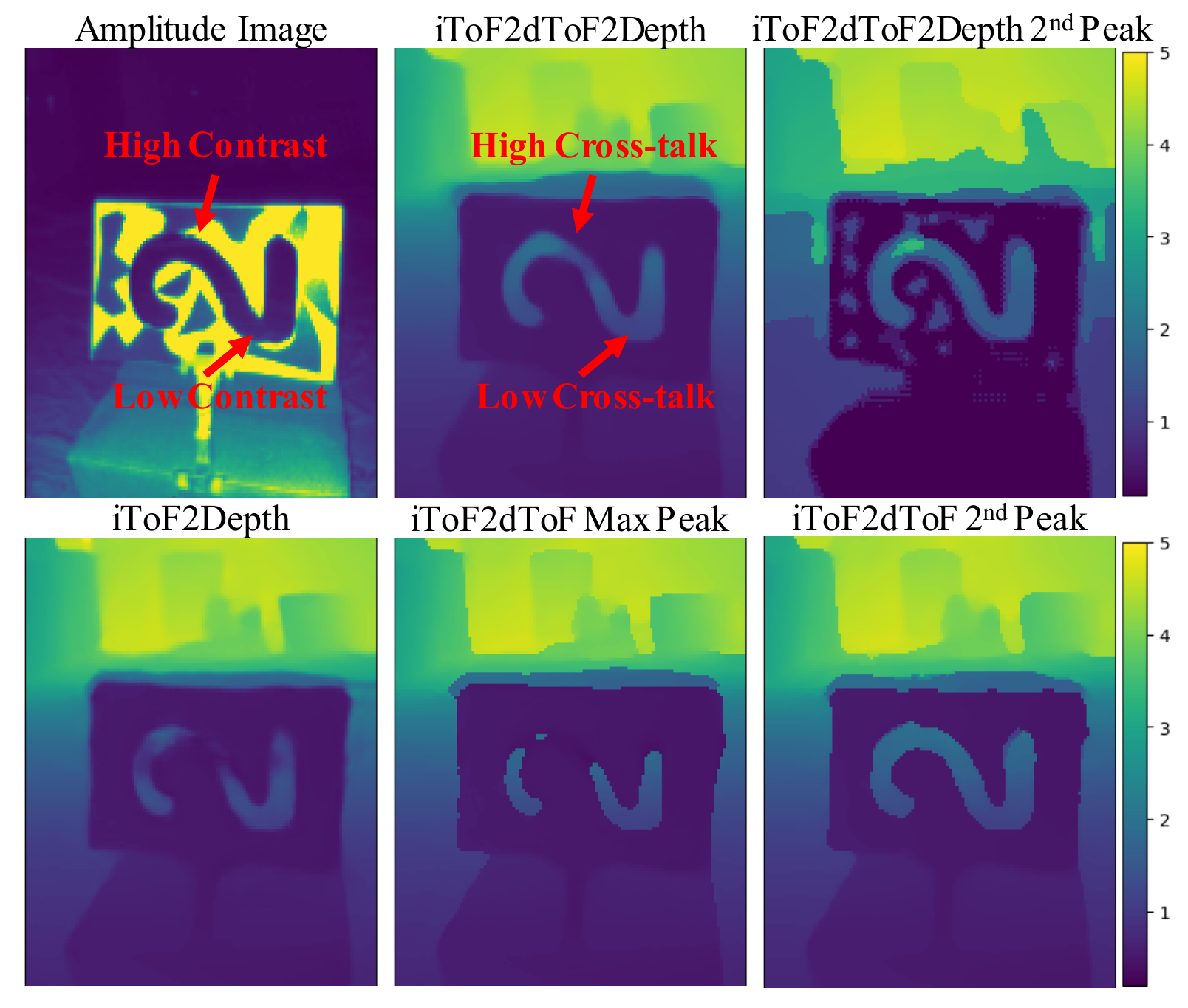}
}
}{%
\caption{\textbf{Optical Cross-Talk Correction.} As discussed in the main paper, optical cross-talk can be corrected by using the second peak of a transient pixel. Similar to iToF2Depth, but to a lesser degree, iToF2dToF2Depth blends the foreground and background depths in the pixels with significant cross-talk (row 1, col 2). Additionally, the lack of supervision on the transient representation of iToF2dToF2Depth, prevents the extraction of useful depth information from the second peaks (row 1, col 3).
}
\label{fig:supplement_xtalk_itof2dtof2depth}
}

\end{floatrow}
\end{figure}

\noindent \textbf{Main observations:} Similar to iToF2Depth and iToF2dToF, iToF2dToF2Depth, is able to mitigate MPI errors significantly in both of our real and synthetic datasets. This can be observed in Figure \ref{fig:supplement_simulation_noise_itof2dtof2depth} and Table \ref{tab:real_test_results}in the percentile MAE for the lowest 95\% errors. For the lowest SNR pixels (highest percentiles), iToF2dToF2Depth, is not able to match the performance of iToF2dToF. Furthermore, Figure \ref{fig:supplement_xtalk_itof2dtof2depth} shows that iToF2dToF2Depth also fails to fully correct for optical cross-talk, illustrating the benefits of explicit supervision on the transient representation. Nonetheless, we do observe two interesting aspects in iToF2dToF2Depth:

\begin{itemize}
    \item \textbf{Transient Reconstruction:} Similar to iToF2dToF, the max peak of iToF2dToF2Depth does not encode to correct for cross-talk (Figure \ref{fig:supplement_xtalk_itof2dtof2depth} Column 2). However, different from iToF2dToF, the second peak in the reconstructed transient of iToF2dToF2Depth does not encode the correct depths (Figure \ref{fig:supplement_xtalk_itof2dtof2depth} Column 3). Nonetheless, it is interesting that the second peak of some of the pixels in iToF2dToF2Depth do seem to encode some information even though no explicit supervision was given on the transient representation of iToF2dToF2Depth.
    \item \textbf{Training Convergence:} Figure \ref{fig:supplement_validation_losses} shows the average validation loss per epoch for 2 training runs of iToF2dToF2Depth and iToF2Depth. The intermediate transient representation makes iToF2dToF2Depth training more stable (less variance in loss), and is able to converge in fewer epochs (around 500) than iToF2Depth (more than 1500 epochs). Therefore, even if the benefits of iToF2dToF at low SNR and challenging scenarios are not needed, it is still beneficial to train data-driven iToF systems using an intermediate transient representation (i.e., iToF2dToF2Depth) to reduce training time without sacrificing accuracy.
\end{itemize}

\noindent \textbf{Summary:} Embedding an intermediate transient representation within a data-driven iToF model provides different benefits regardless if the loss is computed with respect to depth or the transient representation. However, in order to fully benefit from the intermediate dToF representation in challenging scenarios (e.g., cross-talk), explicit supervision on the intermediate representation is required (i.e., iToF2dToF). An interesting direction for future work could be to design loss functions and training strategies that involve the transient representation and the final depth reconstruction, to train iToF2dToF-like models that do not require subsequent rule-based algorithms to recover the correct depth in challenging scenarios.

\begin{table}
\centering
\begin{tabular}{lcccc}
\hline
\multicolumn{5}{c}{\textbf{Real Test Set Percentile MAE (mm)}}                                                                                 \\ \hline
\textbf{Model}                                             & \textbf{0-75\%} & \textbf{75-85\%} & \textbf{85-95\%} & \textbf{95-100\%} \\ \hline
Phasor \cite{gupta2015phasor}                                                  & 42.51           & 234.93            & 440.18            & 1150.66            \\ \hline
iToF2iToF Baseline                                                 & 10.12           & 32.58            & 43.21            & 68.31            \\ \hline
iToF2Depth Baseline                                                & 7.70           & 21.82   & 30.46   & 65.05            \\ \hline
iToF2dToF @400MHz                                                  & \textbf{6.64}   & 19.46            & 27.57            & \textbf{60.29}   \\ \hline
iToF2dToF2Depth @400MHz                                                  & 6.77   & \textbf{19.30}            & \textbf{26.98}            & 65.23   \\ \hline
\end{tabular}
\vspace{-0.1in}
\caption{Percentile MAE for real-world scenes with ground truth captured as described in the main paper in Figure 7.
}
\label{tab:real_test_results}
\end{table}

\vspace{-0.2in}

\vspace{-0.2in}

\clearpage

\subsection{Does iToF2dToF Generalize to Scattering Media?}

In this section we evaluate the generalization capabilities of the data-driven models to a scene with different levels of fog density. Depths in iToF2dToF are reconstructed using the maximum peak.

\medskip

\begin{figure}[h]
\centering
    \includegraphics[width=0.99\textwidth]{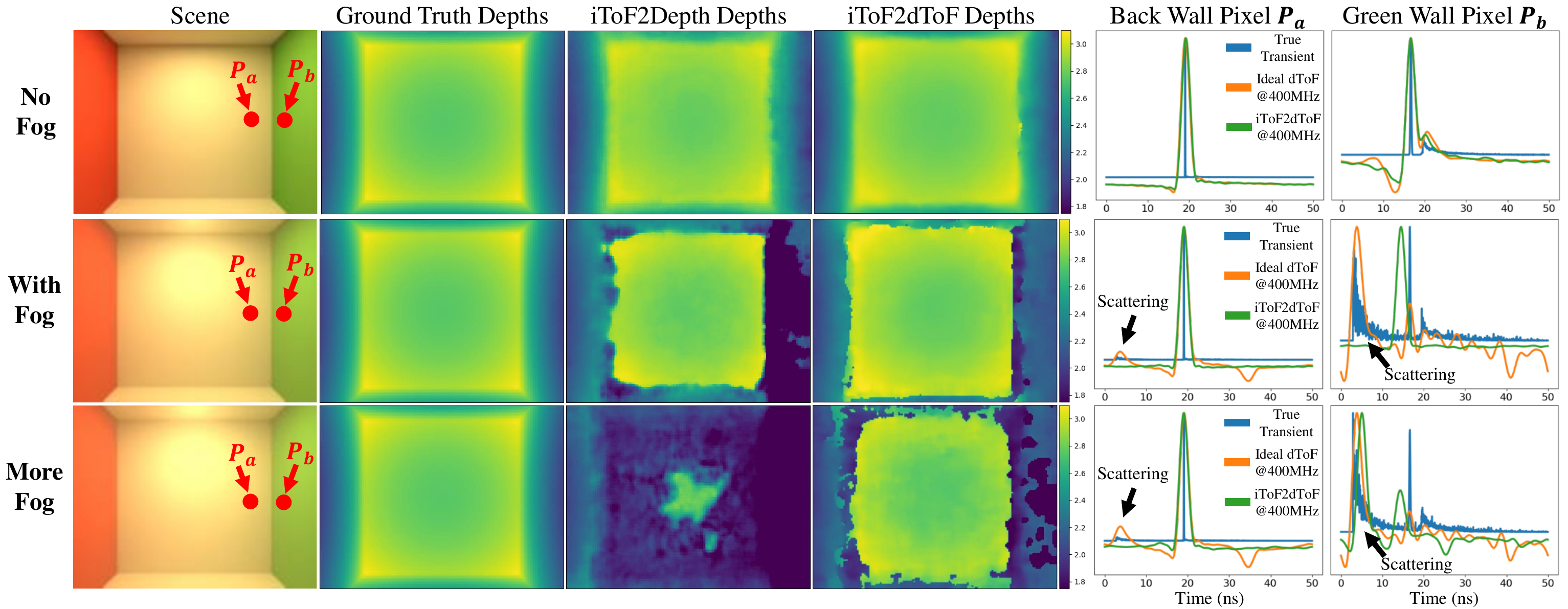}
\vspace{-0.1in}
\caption{\textbf{Fog Simulation Results}. The first row shows the reconstructed depths for the scene without fog. The second and third rows show the reconstructed depths for the same scene simulated with 2 fog density levels. The amount of scattering due to the fog can be observed in the transient pixels plotted in the last two columns. The green wall pixel $P_b$ has a very low reflectivity, so the signal due to scattering is comparable to the reflected signal from the wall. \textit{If possible, zoom in to observe the scattering signal in the transient pixels, in particular the white wall pixel $P_a$.} }

\label{fig:supplement_fog_results}
\end{figure}

\begin{figure}[h]
\centering
    \includegraphics[width=0.8\textwidth]{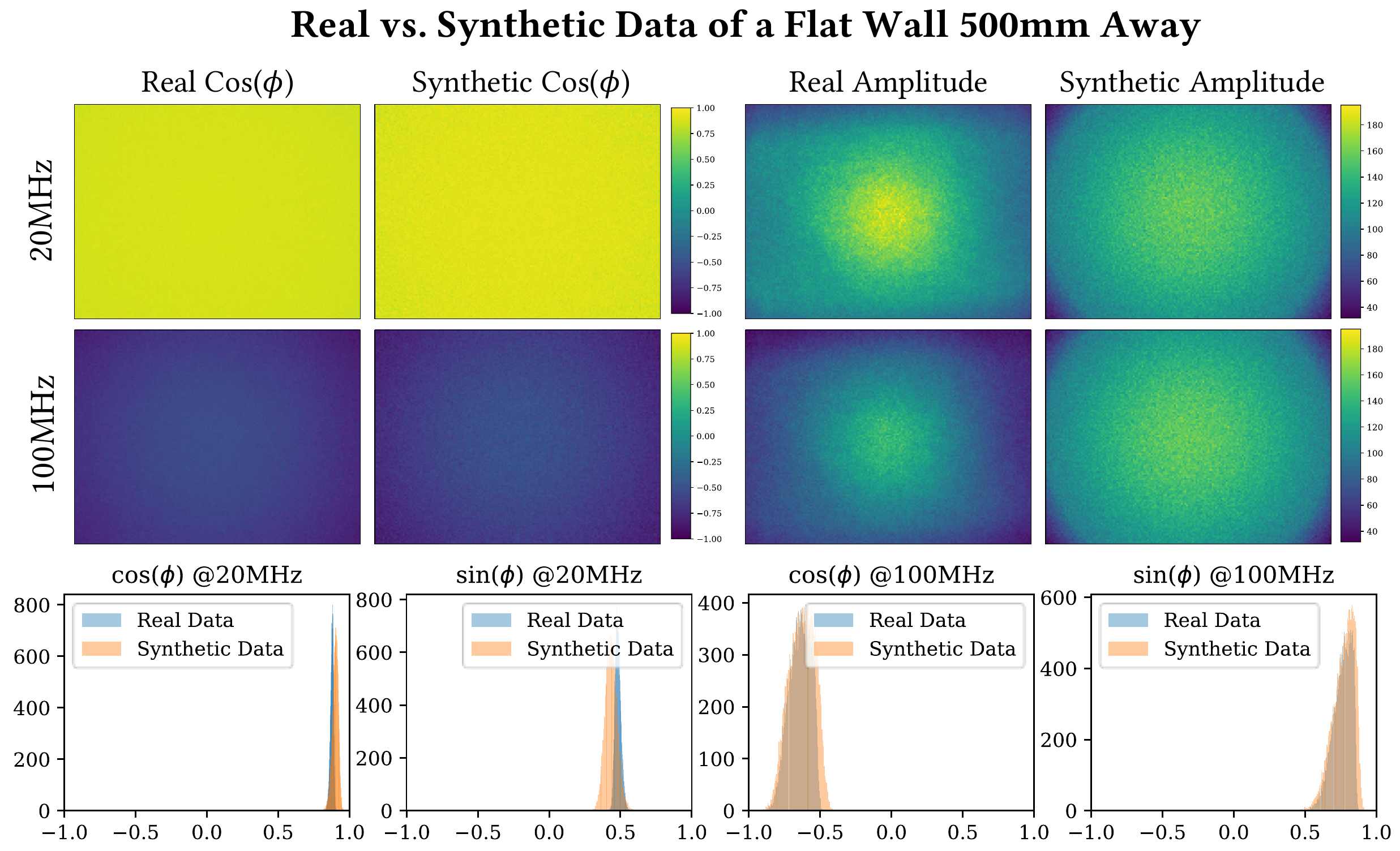}
    \includegraphics[width=0.8\textwidth]{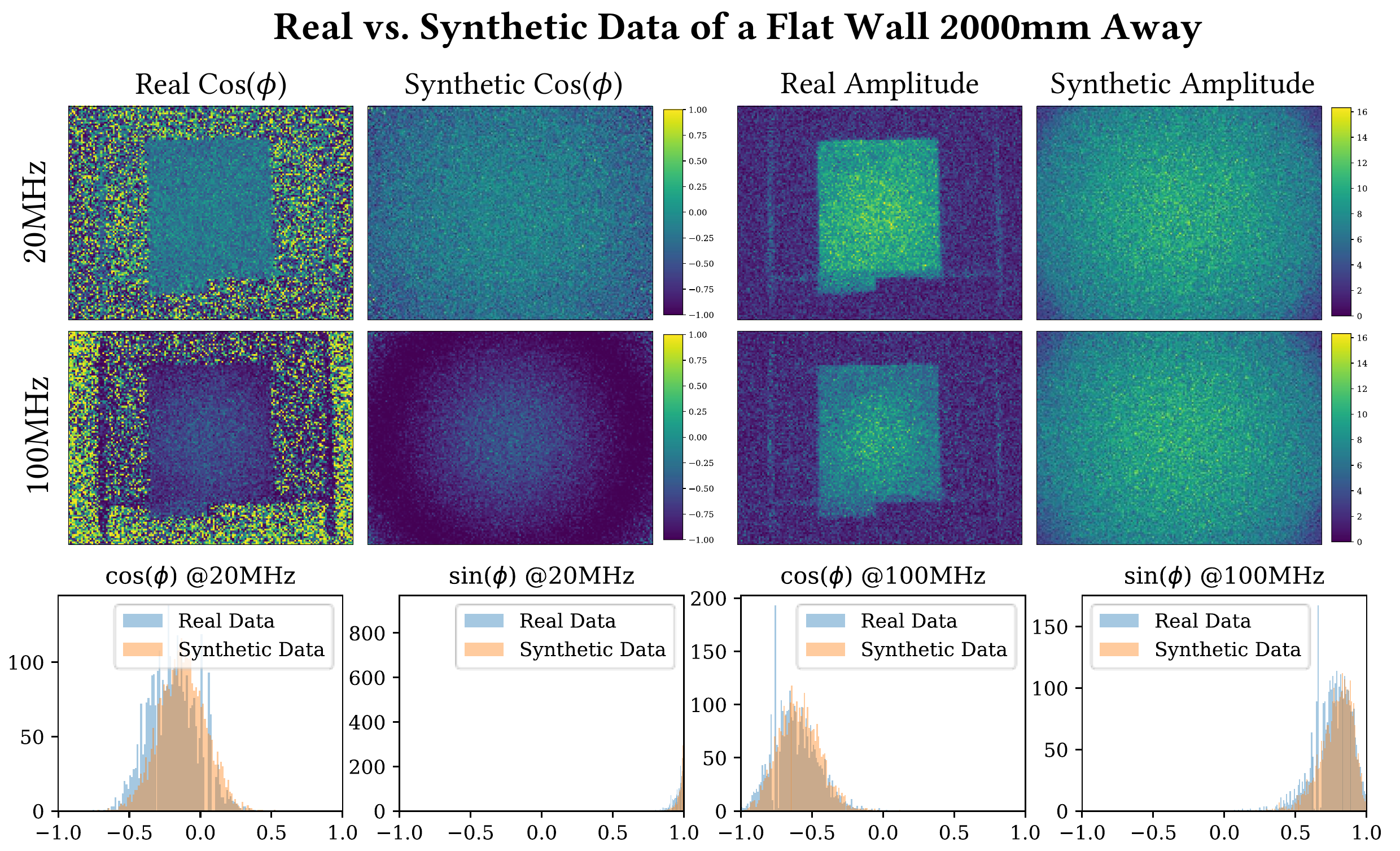}
\vspace{-0.1in}
\caption{\textbf{Synthetic vs. Real ToF Images} of a flat wall 500mm and 2000mm away from the camera. The first row shows the real and synthetic phase images (columns 1 and 2), and the real and synthetc amplitude images (columns 3 and 4), for 20MHz. The second row shows the same images, but for 100MHz. The flat plain used to capture the above real images did not cover the complete field of view of the camera, as seen in the real phase and amplitude images. Therefore, when calculating the distribution of the recovered phases (third row), we cropped the real and synthetic images such that we only included the valid pixels. }
\label{fig:supplement_synthetic_vs_real}
\end{figure}

\noindent \textbf{Main Observations: } Figure \ref{fig:supplement_fog_results} shows the depth reconstruction by iToF2Depth and iToF2dToF in the presence of fog. iToF2Depth completely breaks at the highest fog level, while iToF2dToF continues to reconstruct the back wall. However, the estimated transient pixels by iToF2dToF do not faithfully approximate the ideal dToF transient in the presence of fog (second and third rows). In the scene without fog (first row), we can see that iToF2dToF does a good job in approximating the ideal dToF transient. However, as we introduce fog, which drastically changes the true transient and the ideal dToF transient, iToF2dToF stops matching the ideal dToF. Interestingly, at the highest level of fog (3rd row) in the green wall pixel $P_b$ (last column), iToF2dToF does a more reasonable job. One possible explanation could be that, in this case, the scattering peak is higher than the direct reflection peak, and iToF2dToF interprets this scenario as sparse MPI.

\medskip

\noindent \textbf{Summary: }Complex light transport events caused by scattering media are not well represented in our synthetic dataset. 
This prevents iToF2dToF from producing an accurate estimate of the transient pixels. 
Nonetheless, iToF2dToF does display some robustness to these out-of-distribution examples when compared to iToF2Depth. 
An interesting direction for future work is to explore methods that will enable data-driven transient imaging methods, like iToF2dToF, to generalize to these more complex light transport events.


\clearpage
\section{Appendix}
\label{sec:appendix}

\subsection{Appendix: Additional Simulator Validation Figures}
\label{sec:appendix_simulator_validation}

\clearpage
\subsection{Appendix: Additional Real-world Quantitative Depth Errors}
\label{sec:appendix_quantitative_errors}

\begin{figure}[h]
\centering
    \includegraphics[width=0.9\textwidth]{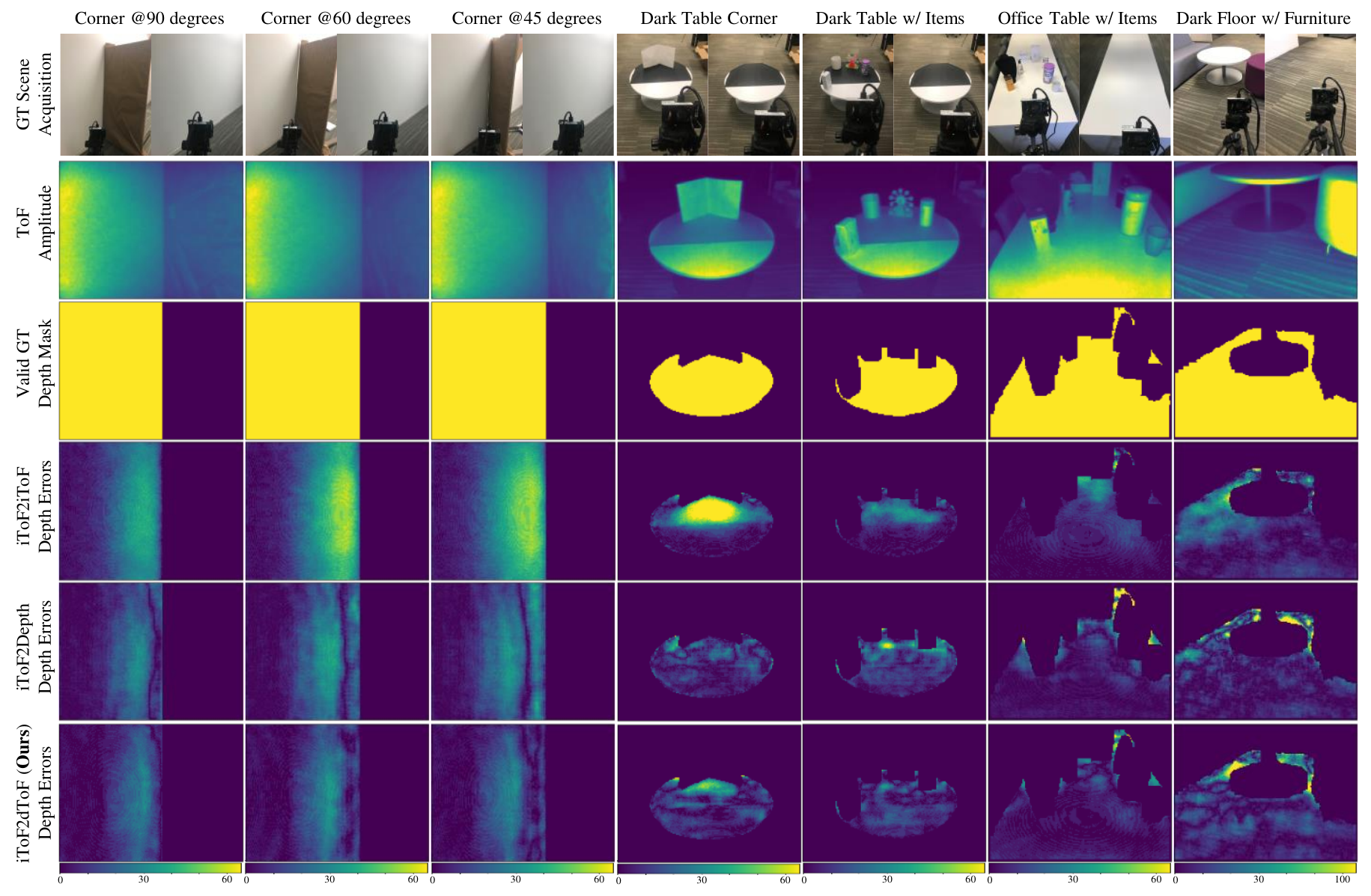}
\vspace{-0.15in}
\caption{\textbf{Real-world Depth Errors at 0.5ms Exposure Time.}}

\label{fig:supplement_quantitative_multiSNR_errors_0.5ms}
\end{figure}

\begin{figure}[h]
\centering
    \includegraphics[width=0.9\textwidth]{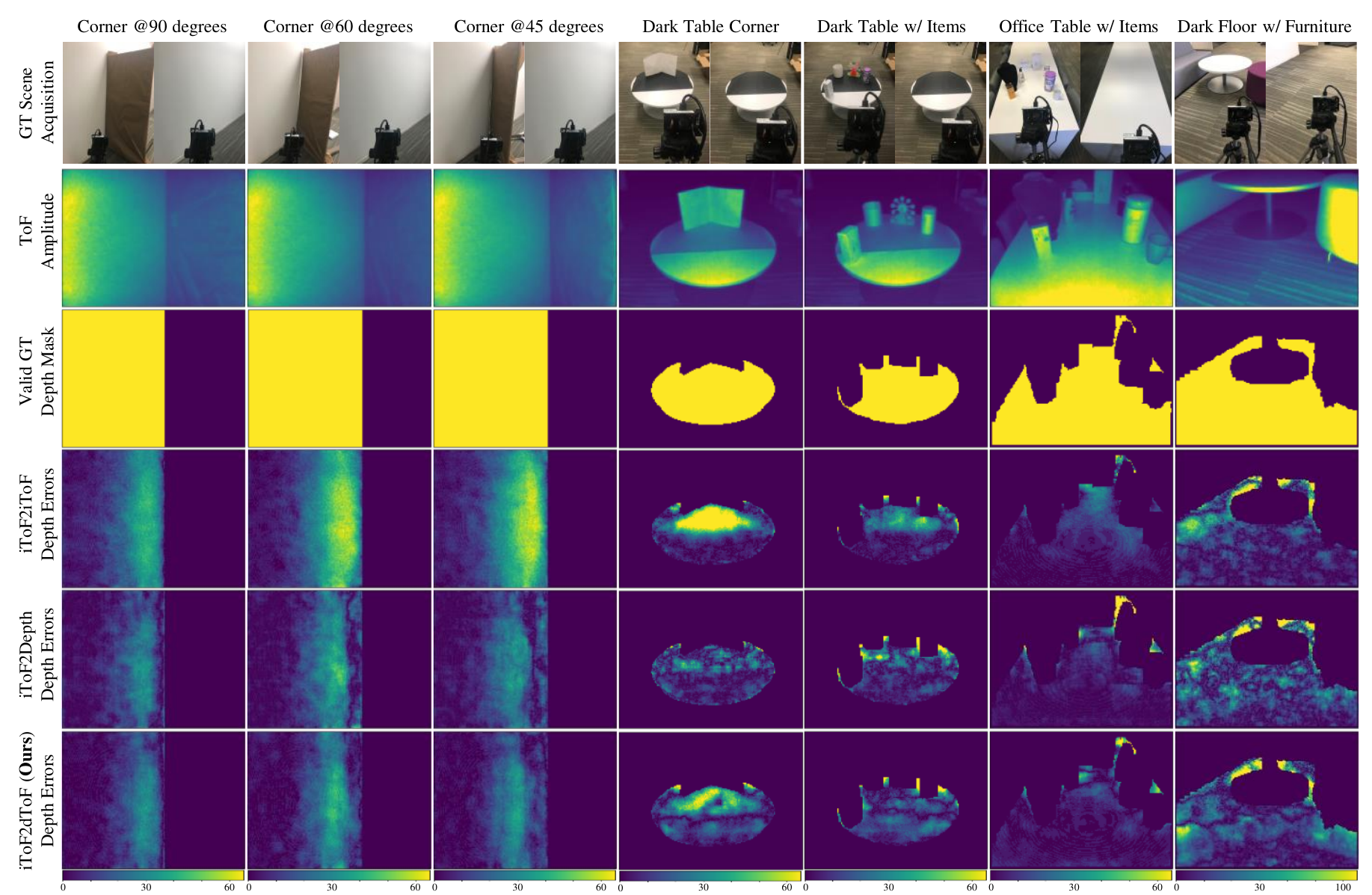}
\vspace{-0.15in}
\caption{\textbf{Real-world Depth Errors at 0.2ms Exposure Time.}}

\label{fig:supplement_quantitative_multiSNR_errors_0.2ms}
\end{figure}

\begin{figure}[h]
\centering
    \includegraphics[width=0.9\textwidth]{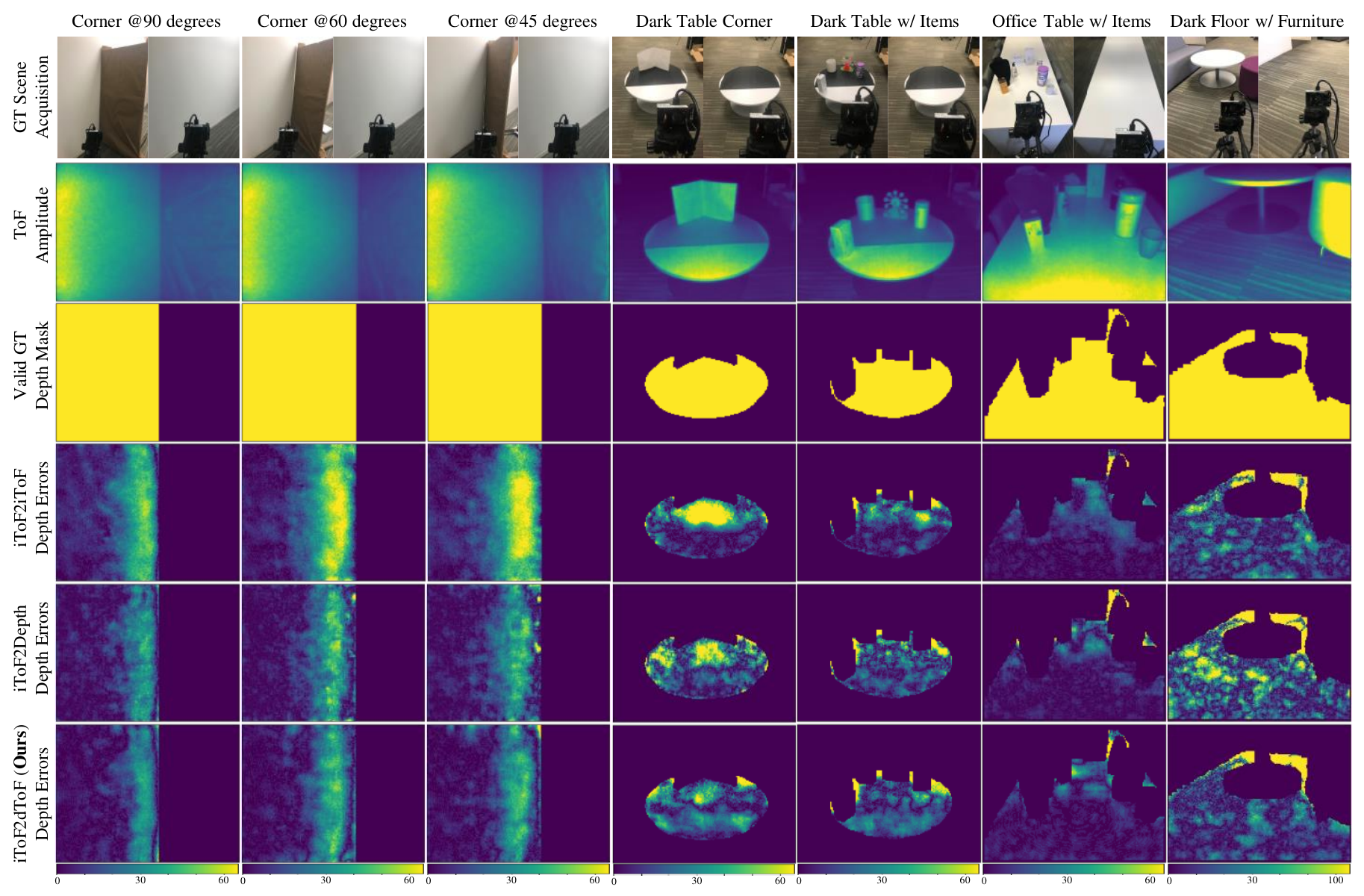}
\vspace{-0.15in}
\caption{\textbf{Real-world Depth Errors at 0.1ms Exposure Time.}}

\label{fig:supplement_quantitative_multiSNR_errors_0.1ms}
\end{figure}

\clearpage

\subsection{Appendix: Additional Exreme Low SNR Depth Reconstructions}
\label{sec:appendix_qualitative_depths}

\begin{figure}[h]
\centering
    \includegraphics[width=0.85\textwidth]{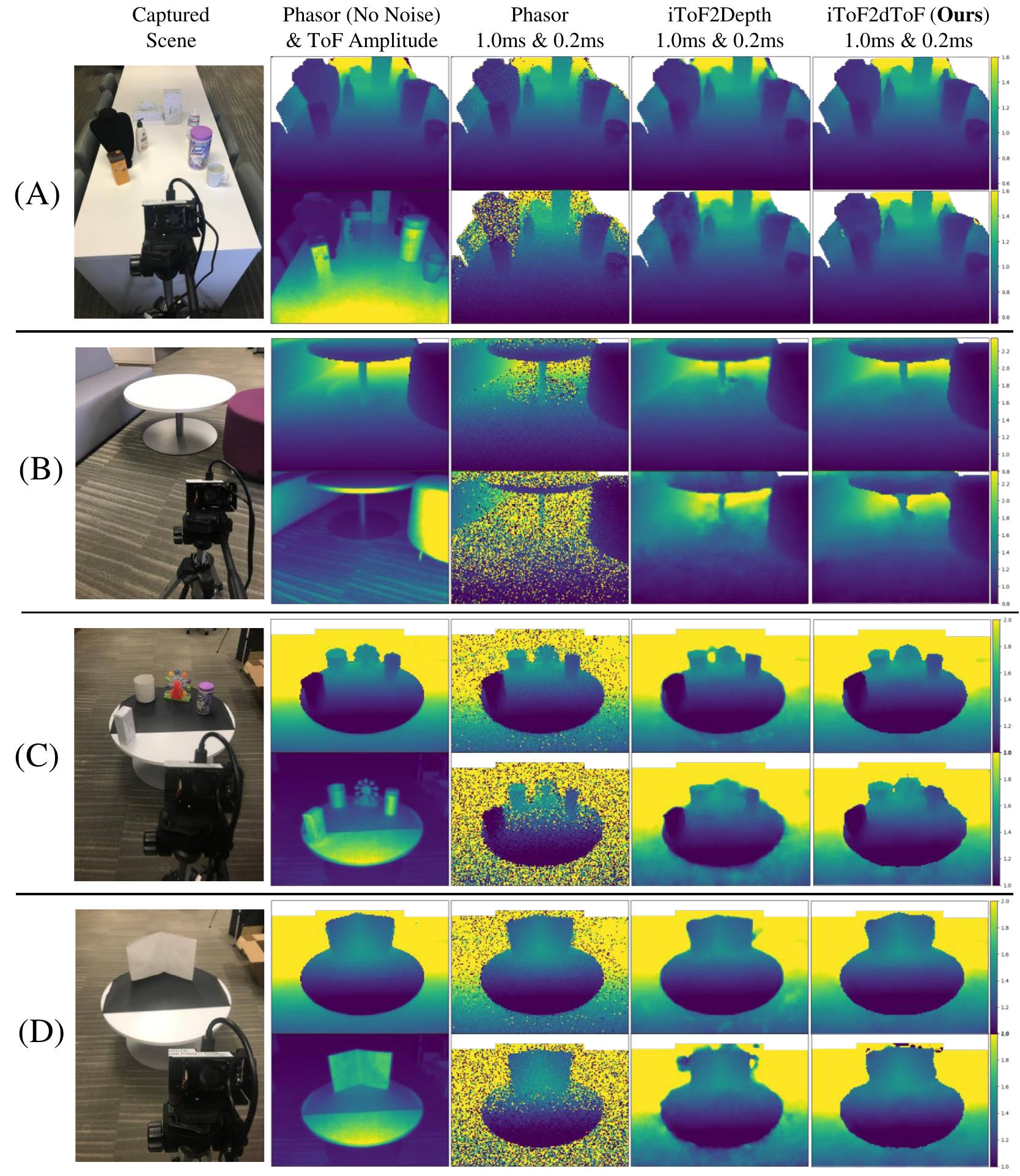}
\vspace{-0.1in}
\caption{\textbf{Extreme Low SNR Depth Reconstructions.} Recovered depths for multiple scenes at low exposure times (1.0ms and 0.2ms) . 
The Phasor (No Noise) images provide an approximate view of how the correct depth image should look like. 
The Phasor depth images (3rd column) does not use apply any denoising to the data, and it is a useful image to identify low and high SNR regions. 
For visualization purposes, we mask pixels (white regions) that exhibit phase wrapping or that were still noisy in the ``noiseless'' Phasor image. 
We find that even at extremely low exposure times iToF2dToF is still able to recover accurate depths in some regions of the image. }

\label{fig:supplement_qualitative_multiSNR_depths_1.0-0.2ms}
\end{figure}


\clearpage


{\small
\bibliographystyle{IEEEtran}
\bibliography{references}
}




\end{document}